\documentclass[runningheads]{llncs}
\pdfoutput=1
\usepackage{times}

\usepackage{graphicx}
\usepackage{amsmath,amssymb}
\usepackage{color}
\usepackage[width=122mm,left=12mm,paperwidth=146mm,height=193mm,top=12mm,paperheight=217mm]{geometry}

\usepackage{booktabs} 
\usepackage{multirow}
\usepackage{tabularx}
\newcolumntype{Y}{>{\centering\arraybackslash}X}

\usepackage{wrapfig}

\usepackage{lscape}

\usepackage{cite}
\usepackage[colorlinks=true]{hyperref}

\def\etal{\emph{et al.} }

\begin{document}

\pagestyle{headings}
\mainmatter
\def\ECCV14SubNumber{95}  

\title{Weakly- and Semi-Supervised Panoptic Segmentation} 

\titlerunning{Weakly- and Semi-Supervised Panoptic Segmentation}

\authorrunning{Li$ ^{\star}$, Arnab$ ^{\star}$, and Torr}

\author{Qizhu Li\thanks{Equal first authorship}, 
		Anurag Arnab$ ^{\star}$, and 
		Philip H.S. Torr} 
\institute{University of Oxford \\
	\email{\{liqizhu, aarnab, phst\}@robots.ox.ac.uk}
}

\maketitle

\begin{abstract}
We present a weakly supervised model that jointly performs both semantic- and instance-segmentation -- a particularly relevant problem given the substantial cost of obtaining pixel-perfect annotation for these tasks.
In contrast to many popular instance segmentation approaches based on object detectors, our method does not predict any overlapping instances.
Moreover, we are able to segment both ``thing'' and ``stuff'' classes, and thus explain all the pixels in the image.
``Thing'' classes are weakly-supervised with bounding boxes, and ``stuff'' with image-level tags.
We obtain state-of-the-art results on Pascal VOC, for both full and weak supervision (which achieves about 95\% of fully-supervised performance).
Furthermore, we present the first weakly-supervised results on Cityscapes for both semantic- and instance-segmentation.
Finally, we use our weakly supervised framework to analyse the relationship between annotation quality and predictive performance, which is of interest to dataset creators.

\keywords{weak supervision, instance segmentation, semantic segmentation, scene understanding}
\end{abstract}


\section{Introduction}

Convolutional Neural Networks (CNNs) excel at a wide array of image recognition tasks \cite{he_cvpr_2016,simonyan_iclr_2015,ren_2015}.
However, their ability to learn effective representations of images requires large amounts of labelled training data \cite{russakovsky_ijcv_2015,sun_iccv_2017}.
Annotating training data is a particular bottleneck in the case of segmentation, where labelling each pixel in the image by hand is particularly time-consuming.
This is illustrated by the Cityscapes dataset where finely annotating a single image took ``more than 1.5h on average" \cite{cordts_cvpr_2016}.
In this paper, we address the problems of semantic- and instance-segmentation using only weak annotations in the form of bounding boxes and image-level tags.
Bounding boxes take only 7 seconds to draw using the labelling method of \cite{papadopoulos_iccv_2017}, and image-level tags an average of 1 second per class \cite{papadopoulos_eccv_2014}.
Using only these weak annotations would correspond to a reduction factor of 30 in labelling a Cityscapes image which emphasises the importance of cost-effective, weak annotation strategies.

Our work differs from prior art on weakly-supervised segmentation \cite{kolesnikov_eccv_2016,wei_cvpr_2017,papandreou_2015,dai_2015,bearman_arxiv_2015} in two primary ways:
Firstly, our model jointly produces semantic- and instance-segmentations of the image, whereas the aforementioned works only output instance-agnostic semantic segmentations.
Secondly, we consider the segmentation of both ``thing'' and ``stuff'' classes \cite{forsyth_1996,adelson_2001}, in contrast to most existing work in both semantic- and instance-segmentation which only consider ``things''.

We define the problem of instance segmentation as labelling every pixel in an image with both its object class and an instance identifier \cite{arnab_cvpr_2017,arnab_bmvc_2016,zhang_iccv_2015}.
It is thus an extension of semantic segmentation, which only assigns each pixel an object class label.
``Thing'' classes (such as ``person'' and ``car'') are countable and are also studied extensively in object detection \cite{everingham_2010,lin_2014}. This is because their finite extent makes it possible to annotate tight, well-defined bounding boxes around them.
``Stuff'' classes (such as ``sky'' and ``vegetation''), on the other hand, are amorphous regions of homogeneous or repetitive textures \cite{forsyth_1996}.
As these classes have ambiguous boundaries and no well-defined shape they are not appropriate to annotate with bounding boxes \cite{lin_di_cvpr_2016}.
Since ``stuff'' classes are not countable, we assume that all pixels of a stuff category belong to the same, single instance.
Recently, this task of jointly segmenting ``things'' and ``stuff'' at an instance-level has also been named ``Panoptic Segmentation'' by \cite{kirillov_arxiv_2018}.

\begin{figure}[t]
\centering

\includegraphics[width=0.9\textwidth]{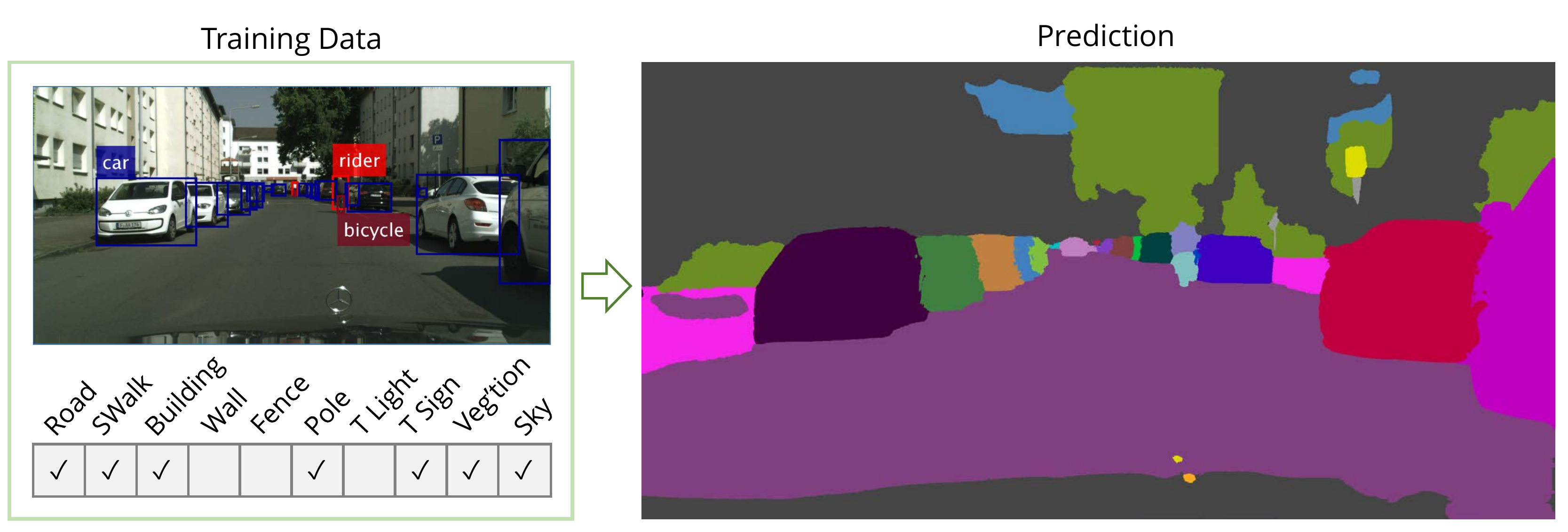}

\caption{We propose a method to train an instance segmentation network from weak annotations in the form of bounding-boxes and image-level tags. Our network can explain both ``thing'' and ``stuff'' classes in the image, and does not produce overlapping instances as common detector-based approaches \cite{he_iccv_2017,dai_cvpr_2016,li_cvpr_2017}.
}
\label{fig:teaser}
\end{figure}

Note that many popular instance segmentation algorithms which are based on object detection architectures \cite{he_iccv_2017,dai_cvpr_2016,li_cvpr_2017,liu_arxiv_2018,liu_cvpr_2016} are not suitable for this task, as also noted by \cite{kirillov_arxiv_2018}.
These methods output a ranked list of proposed instances, where the different proposals are allowed to overlap each other as each proposal is processed independently of the other.
Consequently, these architectures are not suitable where each pixel in the image has to be explained, and assigned a unique label of either a ``thing'' or ``stuff'' class as shown in Fig.~\ref{fig:teaser}.
This is in contrast to other instance segmentation methods such as \cite{arnab_cvpr_2017,bai_cvpr_2017,brabandere_cvprw_2017,kirillov_cvpr_2017,liu_iccv_2017}.

In this work, we use weak bounding box annotations for ``thing'' classes, and image-level tags for ``stuff'' classes.
Whilst there are many previous works on semantic segmentation from image-level labels, the best performing ones \cite{wei_cvpr_2017,wei_pami_2017,oh_cvpr_2017,chaudhry_bmvc_2017} used a saliency prior.
The salient parts of an image are ``thing'' classes in popular saliency datasets \cite{cheng_pami_2015,yang_cvpr_2013,shi_pami_2016} and this prior therefore does not help at all in segmenting ``stuff'' as in our case.
We also consider the ``semi-supervised'' case where we have a mixture of weak- and fully-labelled annotations.

To our knowledge, this is the first work which performs weakly-supervised, non-overlapping instance segmentation, allowing our model to explain all ``thing'' and ``stuff'' pixels in the image (Fig.~\ref{fig:teaser}).
Furthermore, our model jointly produces semantic- and instance-segmentations of the image, which to our knowledge is the first time such a model has been trained in a weakly-supervised manner.
Moreover, to our knowledge, this is the first work to perform either weakly supervised semantic- or instance-segmentation on the Cityscapes dataset.
On Pascal VOC, our method achieves about 95\% of fully-supervised accuracy on both semantic- and instance-segmentation. 
Furthermore, we surpass the state-of-the-art on fully-supervised instance segmentation as well.
Finally, we use our weakly- and semi-supervised framework to examine how model performance varies with the number of examples in the training set and the annotation quality of each example, with the aim of helping dataset creators better understand the trade-offs they face in this context.

\section{Related Work}
Instance segmentation is a popular area of scene understanding research.
Most top-performing algorithms modify object detection networks to output a ranked list of segments instead of boxes \cite{he_iccv_2017,dai_cvpr_2016,li_cvpr_2017,liu_arxiv_2018,liu_cvpr_2016,hariharan_2014}.
However, all of these methods process each instance independently and thus overlapping instances are produced -- one pixel can be assigned to multiple instances simultaneously.
Additionally, object detection based architectures are not suitable for labelling ``stuff'' classes which cannot be described well by bounding boxes \cite{lin_di_cvpr_2016}.
These limitations, common to all of these methods, have also recently been raised by Kirillov \etal\cite{kirillov_arxiv_2018}.
We observe, however, that there are other instance segmentation approaches based on initial semantic segmentation networks \cite{arnab_cvpr_2017,bai_cvpr_2017,brabandere_cvprw_2017,kirillov_cvpr_2017} which do not produce overlapping instances and can naturally handle ``stuff'' classes.
Our proposed approach extends methods of this type to work with weaker supervision.

Although prior work on weakly-supervised instance segmentation is limited, there are many previous papers on weak semantic segmentation, which is also relevant to our task.
Early work in weakly-supervised semantic segmentation considered cases where images were only partially labelled using methods based on Conditional Random Fields (CRFs) \cite{verbeek_nips_2008,he_nips_2009}.
Subsequently, many approaches have achieved high accuracy using only image-level labels \cite{kolesnikov_eccv_2016,wei_cvpr_2017,pinheiro_cvpr_2015,pathak_iccv_2015}, bounding boxes \cite{khoreva_cvpr_2017,papandreou_2015,dai_2015}, scribbles \cite{lin_di_cvpr_2016} and points \cite{bearman_arxiv_2015}.
A popular paradigm for these works is ``self-training'' \cite{scudder_1965}: a model is trained in a fully-supervised manner by generating the necessary ground truth with the model itself in an iterative, Expectation-Maximisation (EM)-like procedure \cite{papandreou_2015,dai_2015,lin_di_cvpr_2016,pathak_iccv_2015}.
Such approaches are sensitive to the initial, approximate ground truth which is used to bootstrap training of the model.
To this end, Khoreva \etal \cite{khoreva_cvpr_2017} showed how, given bounding box annotations, carefully chosen unsupervised foreground-background and segmentation-proposal algorithms could be used to generate high-quality approximate ground truth such that iterative updates to it were not required thereafter.

Our work builds on the ``self-training'' approach to perform instance segmentation.
To our knowledge, only Khoreva \etal \cite{khoreva_cvpr_2017} have published results on weakly-supervised instance segmentation.
However, the model used by \cite{khoreva_cvpr_2017} was not competitive with the existing instance segmentation literature in a fully-supervised setting.
Moreover, \cite{khoreva_cvpr_2017} only considered bounding-box supervision, whilst we consider image-level labels as well.
Recent work by \cite{hu_arxiv_2017} modifies Mask-RCNN \cite{he_iccv_2017} to train it using fully-labelled examples of some classes, and only bounding box annotations of others.
Our proposed method can also be used in a semi-supervised scenario (with a mixture of fully- and weakly-labelled training examples), but unlike \cite{hu_arxiv_2017}, our approach works with only weak supervision as well.
Furthermore, in contrast to \cite{khoreva_cvpr_2017} and \cite{hu_arxiv_2017}, our method does not produce overlapping instances, handles ``stuff'' classes and can thus explain every pixel in an image as shown in Fig.~\ref{fig:teaser}.

\section{Proposed Approach}

We first describe how we generate approximate ground truth data to train semantic- and instance-segmentation models with in Sec.~\ref{sec:gt_intro} through \ref{sec:gt_iterative}.
Thereafter, in Sec.~\ref{sec:model}, we discuss the network architecture that we use.
To demonstrate our method and ensure the reproducibility of our results, we release our approximate ground truth and the code to generate it\footnote{\scriptsize{\url{ https://github.com/qizhuli/Weakly-Supervised-Panoptic-Segmentation}}}.

\subsection{Training with weaker supervision}
\label{sec:gt_intro}
In a fully-supervised setting, semantic segmentation models are typically trained by performing multinomial logistic regression independently for each pixel in the image.
The loss function, the cross entropy between the ground-truth distribution and the prediction, can be written as 
\begin{equation}
L = -\sum_{i \in \Omega}{\log{p(l_i | \mathbf{I})}}
\end{equation}
where $l_i$ is the ground-truth label at pixel $i$, $p(l_i | \mathbf{I})$ is the probability (obtained from a softmax activation) predicted by the neural network for the correct label at pixel $i$ of image $\mathbf{I}$ and $\Omega$ is the set of pixels in the image.

In the weakly-supervised scenarios considered in this paper, we do not have reliable annotations for all pixels in $\Omega$.
Following recent work \cite{khoreva_cvpr_2017,kolesnikov_eccv_2016,bearman_arxiv_2015,pathak_iccv_2015}, we use our weak supervision and image priors to approximate the ground-truth for a subset $\Omega' \subset \Omega$ of the pixels in the image.
We then train our network using the estimated labels of this smaller subset of pixels.
Section~\ref{sec:bounding_box} describes how we estimate $\Omega'$ and the corresponding labels for images with only bounding-box annotations, and Sec.~\ref{sec:image_level} for image-level tags.

Our approach to approximating the ground truth is based on the principle of only assigning labels to pixels which we are confident about, and marking the remaining set of pixels, $\Omega \setminus \Omega'$, as ``ignore'' regions over which the loss is not computed.
This is motivated by Bansal~\etal\cite{bansal_arxiv_2017} who observed that sampling only 4\% of the pixels in the image for computing the loss during fully-supervised training yielded about the same results as sampling all pixels, as traditionally done.
This supported their hypothesis that most of the training data for a pixel-level task is statistically correlated within an image, and that randomly sampling a much smaller set of pixels is sufficient. 
Moreover, \cite{pohlen_cvpr_2017} and \cite{li_bmvc_2017} showed improved results by respectively sampling only 6\% and 12\% of the hardest pixels, instead of all of them, in fully-supervised training.

\subsection{Approximate ground truth from bounding box annotations}
\label{sec:bounding_box}
\begin{figure}[!t]
\centering

\begin{tabularx}{\linewidth}{ Y Y Y }
\includegraphics[height=3cm]{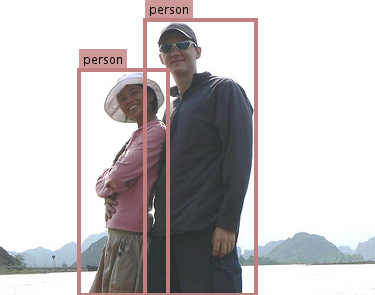} &
\includegraphics[height=3cm]{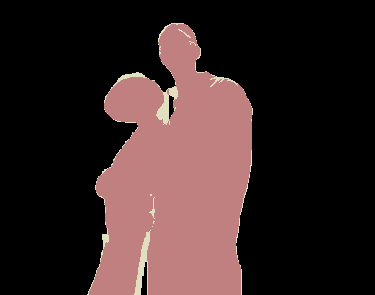} &
\includegraphics[height=3cm]{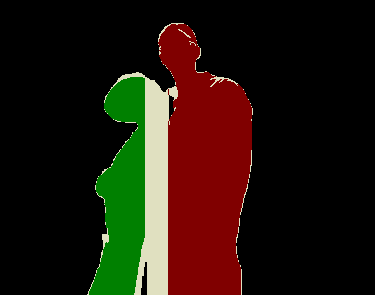}
\\
(a) Input image & (b) Semantic segmentation approximate ground truth & (c) Instance segmentation approximate ground truth
\end{tabularx}
\caption{An example of generating approximate ground truth from bounding box annotations for an image (a). A pixel is labelled the with the bounding-box label if it belongs to the foreground masks of both GrabCut \cite{rother_2004} and MCG \cite{arbelaez_2014} (b). Approximate instance segmentation ground truth is generated using the fact that each bounding box corresponds to an instance (c). Grey regions are ``ignore'' labels over which the loss is not computed due to ambiguities in label assignment.}
\label{fig:bbox_example}

\end{figure}

We use GrabCut~\cite{rother_2004} (a classic foreground segmentation technique given a bounding-box prior) and MCG \cite{arbelaez_2014} (a segment-proposal algorithm) to obtain a foreground mask from a bounding-box annotation, following \cite{khoreva_cvpr_2017}.
To achieve high precision in this approximate labelling, a pixel is only assigned to the object class represented by the bounding box if both GrabCut and MCG agree (Fig.~\ref{fig:bbox_example}).

Note that the final stage of MCG uses a random forest trained with pixel-level supervision on Pascal VOC to rank all the proposed segments.
We do not perform this ranking step, and obtain a foreground mask from MCG by selecting the proposal that has the highest Intersection over Union (IoU) with the bounding box annotation.

This approach is used to obtain labels for both semantic- and instance-segmentation as shown in Fig.~\ref{fig:bbox_example}.
As each bounding box corresponds to an instance, the foreground for each box is the annotation for that instance.
If the foreground of two bounding boxes of the same class overlap, the region is marked as ``ignore'' as we do not have enough information to attribute it to either instance.

\subsection{Approximate ground-truth from image-level annotations}
\label{sec:image_level}
\begin{figure}[t]
\centering

\begin{tabularx}{\linewidth}{YYY}
\includegraphics[width=0.98\linewidth]{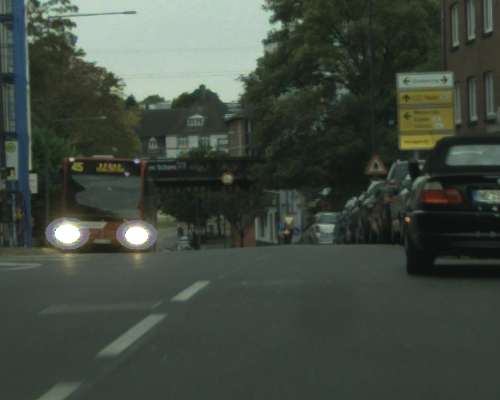} &
	\vspace{-8\baselineskip}
	\begin{tabular}{ll}
		\includegraphics[width=0.48\linewidth]{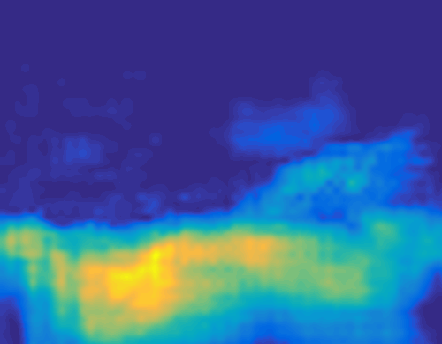} &
		\includegraphics[width=0.48\linewidth]{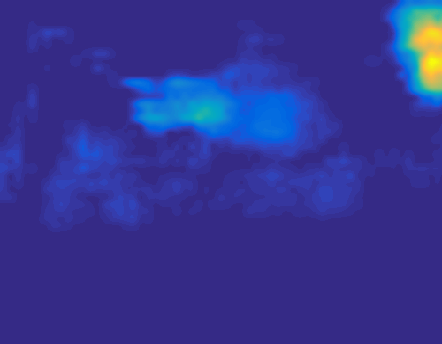}  \\
		 \includegraphics[width=0.48\linewidth]{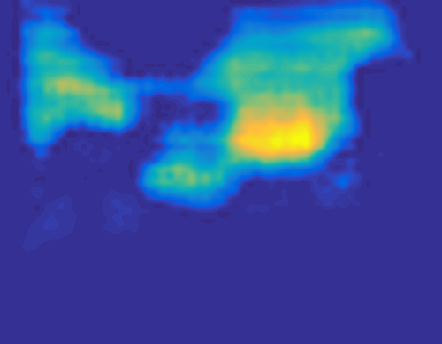}  & 
		\includegraphics[width=0.48\linewidth]{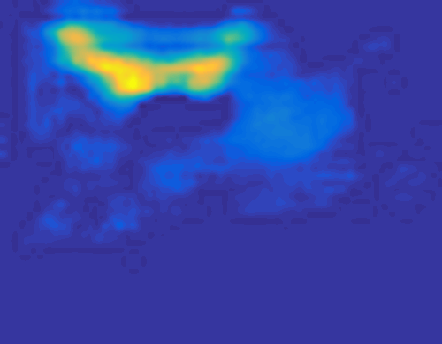} \\
	\end{tabular}
 &
\includegraphics[width=0.98\linewidth]{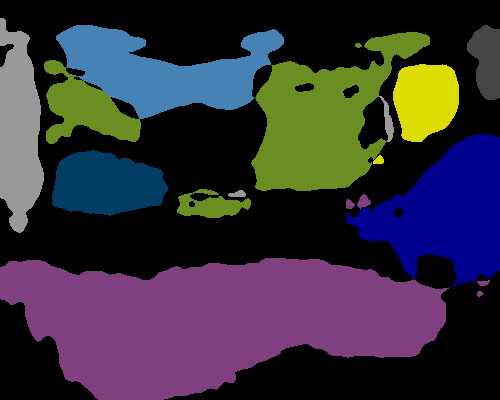}
\\
\scriptsize{Input image} & 
\scriptsize{Localisation heatmaps for road, building, vegetation and sky} & 
\scriptsize{Approximate ground truth generated from image tags}
\end{tabularx}

\caption{Approximate ground truth generated from image-level tags using weak localisation cues from a multi-label classification network.
Cluttered scenes from Cityscapes with full ``stuff'' annotations makes weak localisation more challenging than Pascal VOC and ImageNet that only have ``things'' labels.
Black regions are labelled ``ignore''. Colours follow Cityscapes convention.
}
\label{fig:cam_example}
\end{figure}

When only image-level tags are available, we leverage the fact that CNNs trained for image classification still have localisation information present in their convolutional layers \cite{zhou_cvpr_2016}.
Consequently, when presented with a dataset of only images and their tags, we first train a network to perform multi-label classification.
Thereafter, we extract weak localisation cues for all the object classes that are present in the image (according to the image-level tags).
These localisation heatmaps (as shown in Fig.~\ref{fig:cam_example}) are thresholded to obtain the approximate ground-truth for a particular class.
It is possible for localisation heatmaps for different classes to overlap. 
In this case, thresholded heatmaps occupying a smaller area are given precedence.
We found this rule, like \cite{kolesnikov_eccv_2016}, to be effective in preventing small or thin objects from being missed.

Though this approach is independent of the weak localisation method used, we used Grad-CAM \cite{selvaraju_iccv_2017}.
Grad-CAM is agnostic to the network architecture unlike CAM \cite{zhou_cvpr_2016} and also achieves better performance than Excitation BP~\cite{zhang_eccv_2016} on the ImageNet localisation task \cite{russakovsky_ijcv_2015}.

We cannot differentiate different instances of the same class from only image tags as the number of instances is unknown.
This form of weak supervision is thus appropriate for ``stuff'' classes which cannot have multiple instances.
Note that saliency priors, used by many works such as \cite{wei_cvpr_2017,wei_pami_2017,oh_cvpr_2017} on Pascal VOC, are not suitable for ``stuff'' classes as popular saliency datasets \cite{cheng_pami_2015,yang_cvpr_2013,shi_pami_2016} only consider ``things'' to be salient. 

\subsection{Iterative ground truth approximation}
\label{sec:gt_iterative}
\begin{figure}[t]
\centering

\begin{tabularx}{\linewidth}{ ccccc}

\includegraphics[width=0.19\linewidth]{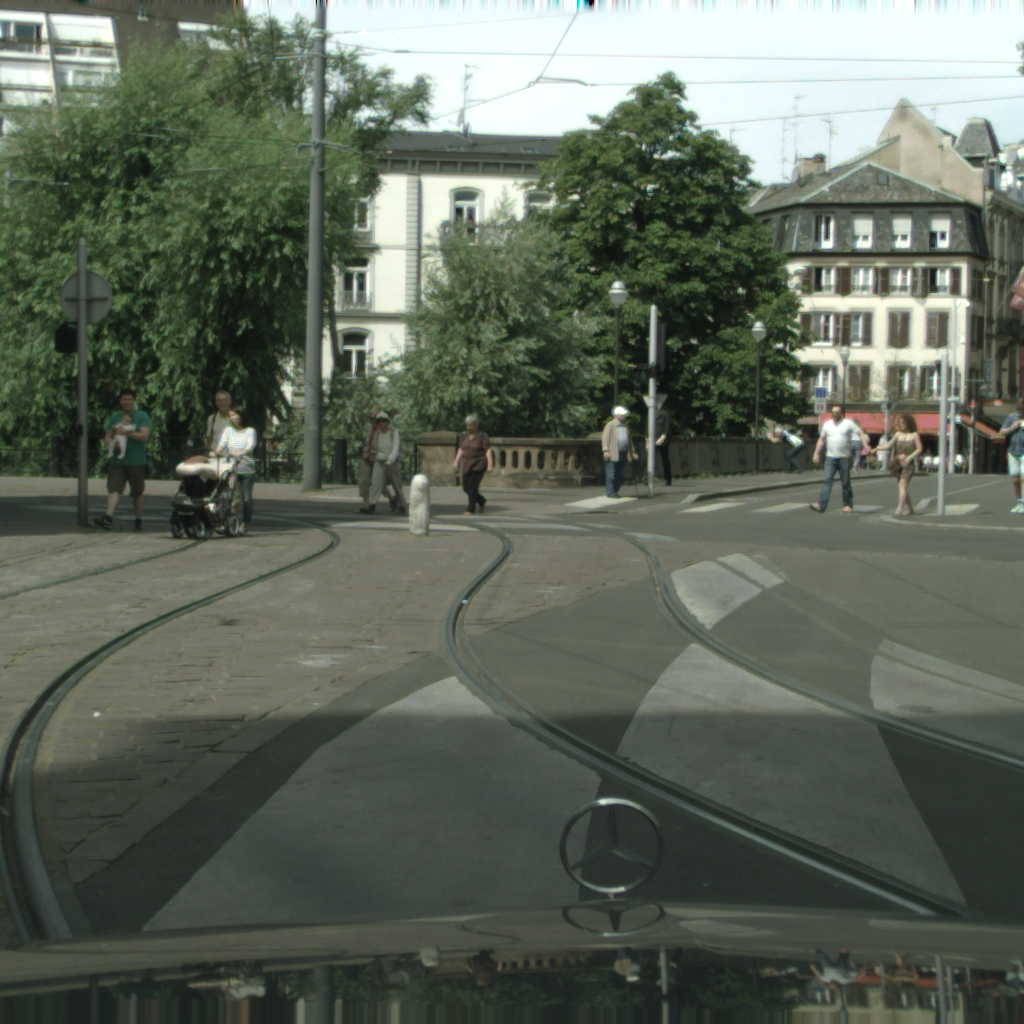}
&
\includegraphics[width=0.19\linewidth]{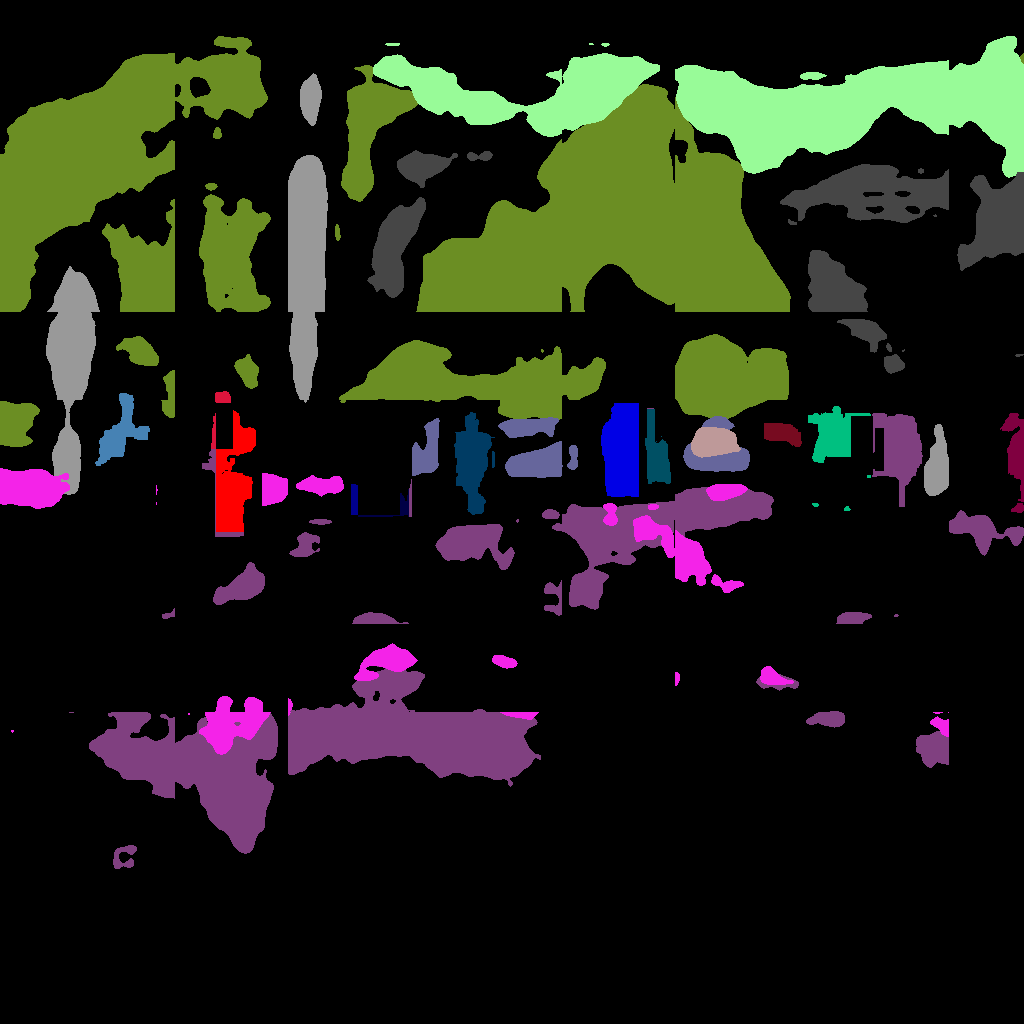}
&
\includegraphics[width=0.19\linewidth]{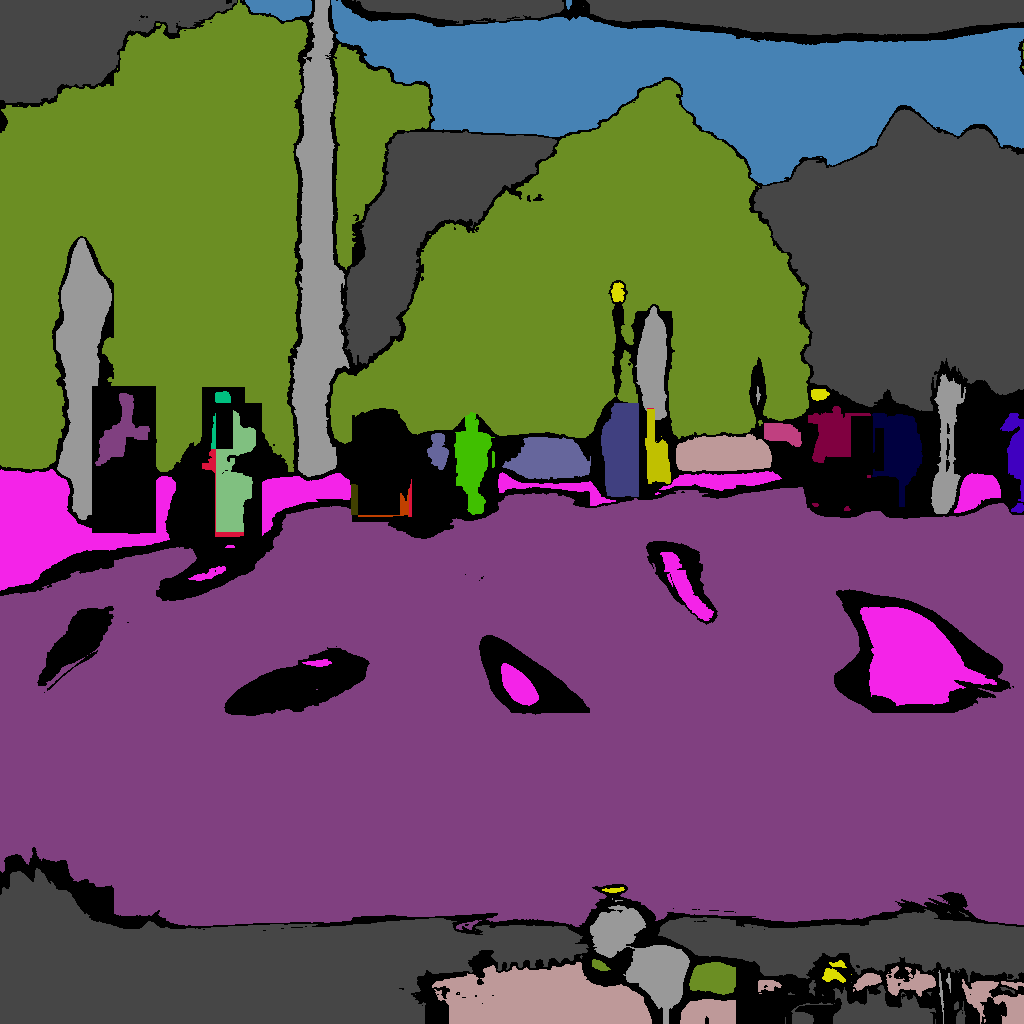}
&
\includegraphics[width=0.19\linewidth]{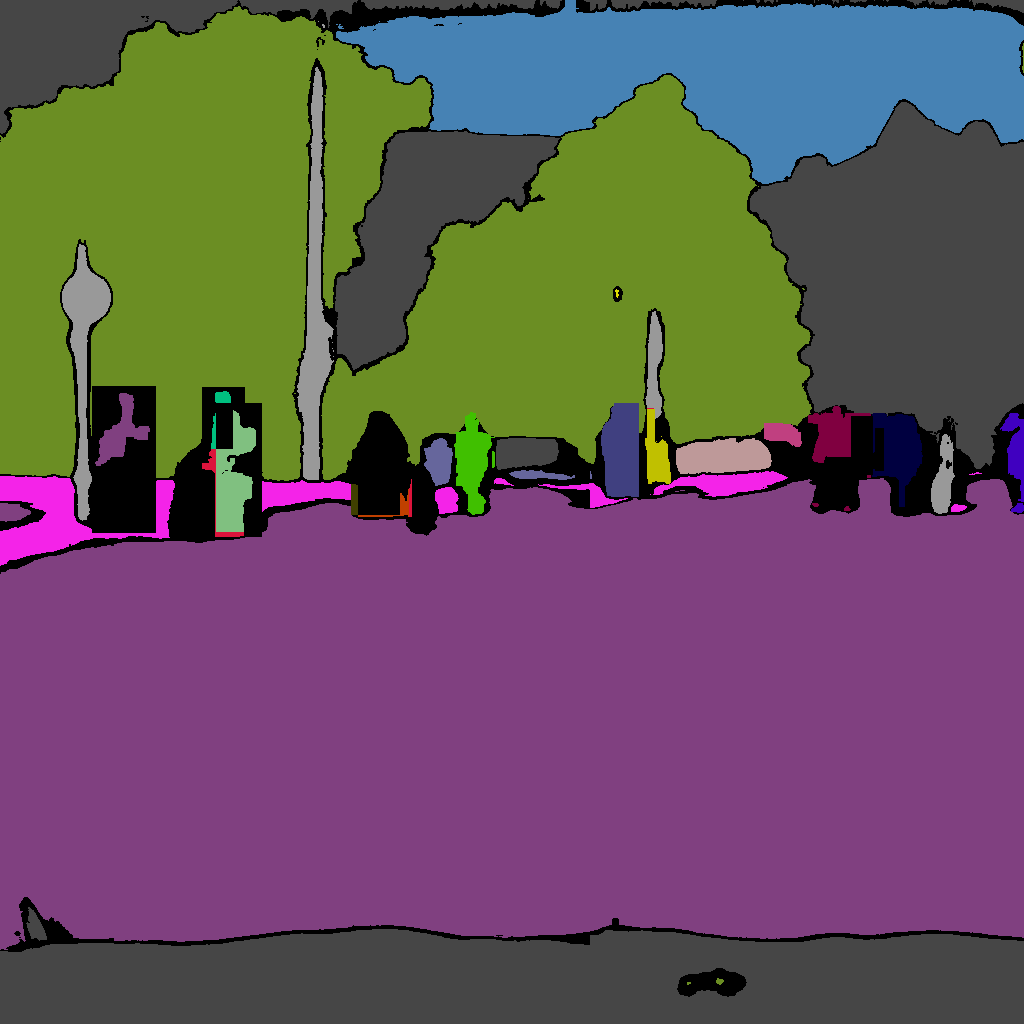}
&
\includegraphics[width=0.19\linewidth]{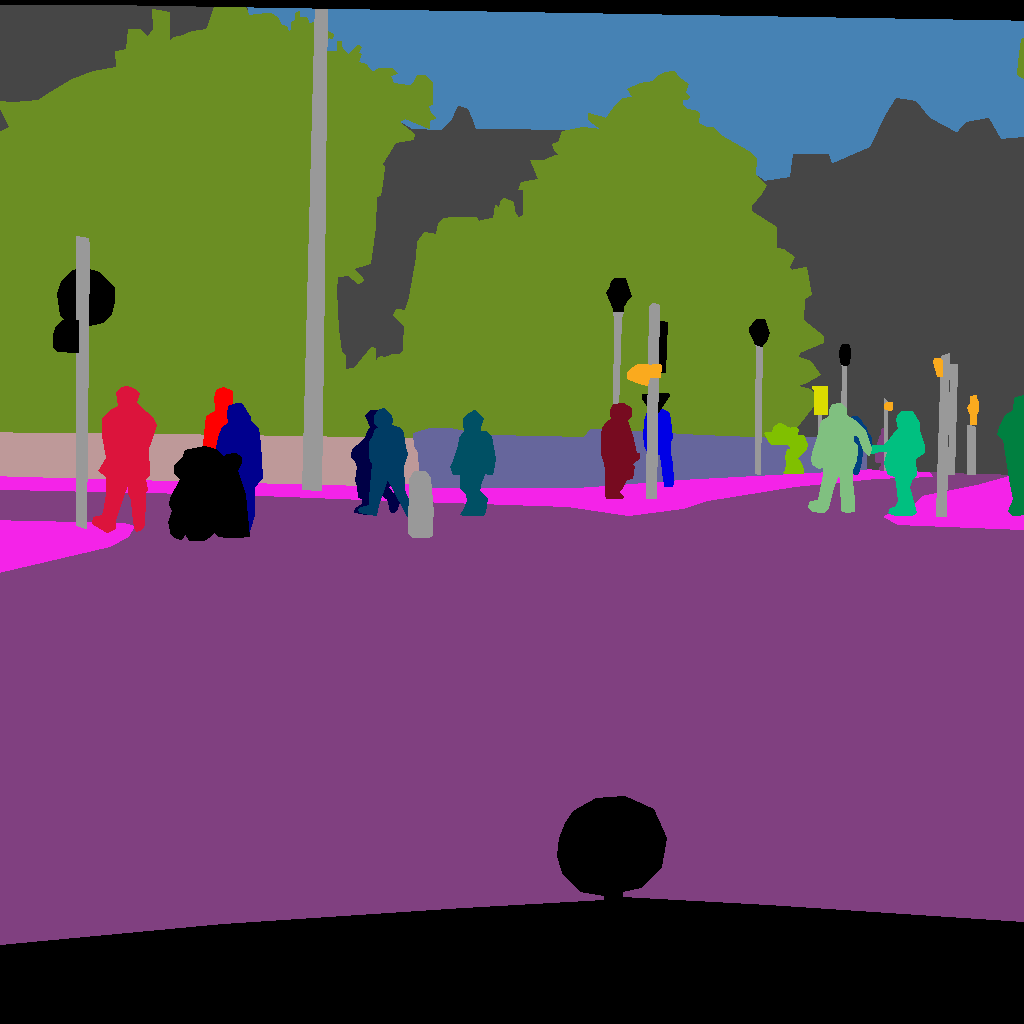}
\\
\scriptsize{Input Image} & \scriptsize{Iteration 0} & \scriptsize{Iteration 2} & \scriptsize{Iteration 5} & \scriptsize{Ground truth} \\

\end{tabularx}

\caption{
	By using the output of the trained network, the initial approximate ground truth  produced according to Sec.~\ref{sec:bounding_box} and \ref{sec:image_level} (Iteration 0) can be improved. Black regions are ``ignore'' labels over which the loss is not computed in training. Note for instance segmentation, permutations of instance labels of the same class are equivalent.}
\label{fig:iterative_gt}
\vspace{-1\baselineskip}
\end{figure}
The ground truth approximated in Sec.~\ref{sec:bounding_box} and \ref{sec:image_level} can be used to train a network from random initialisation.
However, the ground truth can subsequently be iteratively refined by using the outputs of the network on the training set as the new approximate ground truth as shown in Fig~\ref{fig:iterative_gt}.
The network's output is also post-processed with DenseCRF \cite{krahenbuhl_2011} using the parameters of Deeplab \cite{chen_2015} (as also done by \cite{kolesnikov_eccv_2016,khoreva_cvpr_2017}) to improve the predictions at boundaries.
Moreover, any pixel labelled a ``thing'' class that is outside the bounding-box of the ``thing'' class is set to ``ignore'' as we are certain that a pixel for a thing class cannot be outside its bounding box.
For a dataset such as Pascal VOC, we can set these pixels to be ``background'' rather than ``ignore''.
This is because ``background'' is the only ``stuff'' class in the dataset.

\subsection{Network Architecture}
\label{sec:model}
\begin{figure}[t]
\centering
\includegraphics[width=0.98\linewidth]{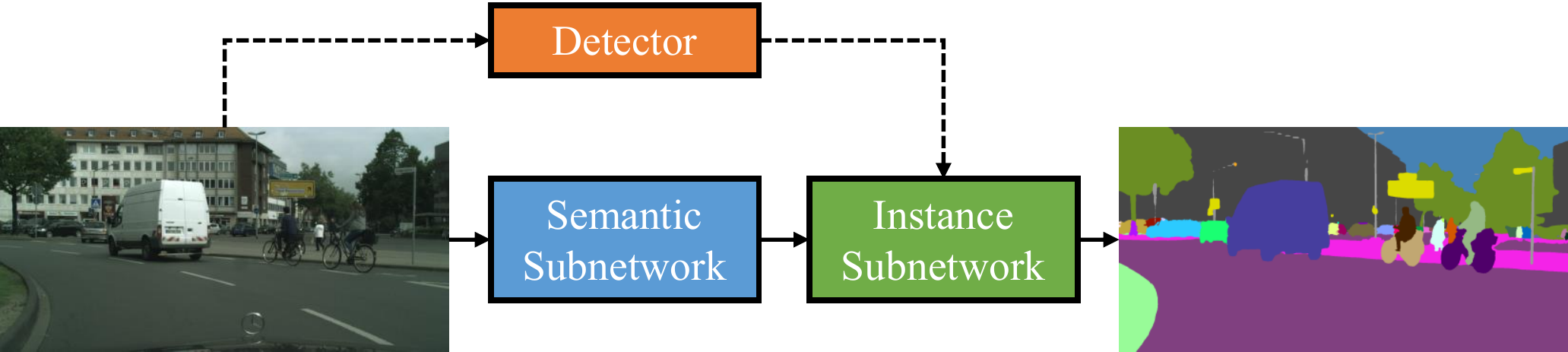}
\caption{Overview of the network architecture. An initial semantic segmentation is partitioned into an instance segmentation, using the output of an object detector as a cue. Dashed lines indicate paths which are not backpropagated through during training.}
\label{fig:model}
\end{figure}
Using the approximate ground truth generation method described in this section, we can train a variety of segmentation models.
Moreover, we can trivially combine this with full human-annotations to operate in a semi-supervised setting.
We use the architecture of Arnab~\etal\cite{arnab_cvpr_2017} as it produces both semantic- and instance-segmentations, and can be trained end-to-end, given object detections.
This network consists of a semantic segmentation subnetwork, followed by an instance subnetwork which partitions the initial semantic segmentation into an instance segmentation with the aid of object detections, as shown in Fig.~\ref{fig:model}.

We denote the output of the first module, which can be any semantic segmentation network, as $\mathbf{Q}$ where $Q_i(l)$ is the probability of pixel $i$ of being assigned semantic label $l$.
The instance subnetwork has two inputs -- $\mathbf{Q}$ and a set of object detections for the image.
There are $D$ detections, each of the form $\left(l_d, s_d, B_d \right)$ where $l_d$ is the detected class label, $s_d \in [0,1]$ the score and $B_d$ the set of pixels lying within the bounding box of the $d^{th}$ detection.
This model assumes that each object detection represents a possible instance, and it assigns every pixel in the initial semantic segmentation an instance label using a Conditional Random Field (CRF).
This is done by defining a multinomial random variable, $X_i$, at each of the $N$ pixels in the image, with $\mathbf{X} = [X_1, X_2 \ldots , X_N]^{\top}$.
This variable takes on a label from the set $\{1,\ldots,D\}$ where $D$ is the number of detections.
This formulation ensures that each pixel can only be assigned one label.
The energy of the assignment $\mathbf{x}$ to all instance variables $\mathbf{X}$ is then defined as
\begin{equation}
E(\mathbf{X} = \mathbf{x}) = -\sum_{i}^{N} \ln \left(w_1 \psi_{Box}(x_i) + w_2 \psi_{Global}(x_i) + \epsilon \right) + \sum_{i < j}^{N}\psi_{Pairwise}(x_i, x_j).
\end{equation}
The first unary term, the box term, encourages a pixel to be assigned to the instance represented by a detection if it falls within its bounding box,
\begin{equation}
\psi_{Box}(X_i = k) = 
	\begin{cases}
	s_k Q_i(l_k)  & \text{if } i \in B_k\\
	0             & \text{otherwise}.
	\end{cases}
\end{equation}
Note that this term is robust to false-positive detections \cite{arnab_cvpr_2017} since it is low if the semantic segmentation at pixel $i$, $Q_i(l_k)$ does not agree with the detected label, $l_k$. The global term,
\begin{equation}
\psi_{Global}(X_i = k) = Q_{i}(l_k),
\end{equation}
is independent of bounding boxes and can thus overcome errors in mislocalised bounding boxes not covering the whole instance.
Finally, the pairwise term is the common densely-connected Gaussian and bilateral filter \cite{krahenbuhl_2011} encouraging appearance and spatial consistency.

In contrast to \cite{arnab_cvpr_2017}, we also consider stuff classes (which object detectors are not trained for), by simply adding ``dummy'' detections covering the whole image with a score of 1 for all stuff classes in the dataset.
This allows our network to jointly segment all ``things'' and ``stuff'' classes at an instance level.
As mentioned before, the box and global unary terms are not affected by false-positive detections arising from detections for classes that do not correspond to the initial semantic segmentation $\mathbf{Q}$.
The Maximum-a-Posteriori (MAP) estimate of the CRF is the final labelling, and this is obtained by using mean-field inference, which is formulated as a differentiable, recurrent network \cite{zheng_2015,arnab_ieeespm_2018}.

We first train the semantic segmentation subnetwork using a standard cross-entropy loss with the approximate ground truth described in Sec~\ref{sec:bounding_box} and \ref{sec:image_level}.
Thereafter, we append the instance subnetwork and finetune the entire network end-to-end.
For the instance subnetwork, the loss function must take into account that different permutations of the same instance labelling are equivalent.
As a result, the ground truth is ``matched'' to the prediction before the cross-entropy loss is computed as described in \cite{arnab_cvpr_2017}.
\section{Experimental Evaluation}

\subsection{Experimental Set-up}
\paragraph{Datasets and weak supervision}
We evaluate on two standard segmentation datasets, Pascal VOC \cite{everingham_2010} and Cityscapes \cite{cordts_cvpr_2016}.
Our weakly- and fully-supervised experiments are trained with the same images, but in the former case, pixel-level ground truth is approximated as described in Sec.~\ref{sec:gt_intro} through \ref{sec:gt_iterative}.

Pascal VOC has 20 ``thing'' classes annotated, for which we use bounding box supervision. There is a single ``background'' class for all other object classes.
Following common practice on this dataset, we utilise additional images from the SBD dataset \cite{hariharan_2011} to obtain a training set of 10582 images.
In some of our experiments, we also use 54000 images from Microsoft COCO \cite{lin_2014} only for the initial pretraining of the semantic subnetwork.
We evaluate on the validation set, of 1449 images, as the evaluation server is not available for instance segmentation.

Cityscapes has 8 ``thing'' classes, for which we use bounding box annotations, and 11 ``stuff'' class labels for which we use image-level tags.
We train our initial semantic segmentation model with the images for which 19998 coarse and 2975 fine annotations are available.
Thereafter, we train our instance segmentation network using the 2975 images with fine annotations available as these have instance ground truth labelled.
Details of the multi-label classification network we trained in order to obtain weak localisation cues from image-level tags (Sec.~\ref{sec:image_level}) are described in the supplementary.
When using Grad-CAM, the original authors originally used a threshold of 15\% of the maximum value for weak localisation on ImageNet.
However, we increased the threshold to 50\% to obtain higher precision on this more cluttered dataset.

\paragraph{Network training}
Our underlying segmentation network is a reimplementation of PSPNet~\cite{zhao_cvpr_2017}.
For fair comparison to our weakly-supervised model, we train a fully-supervised model ourselves, using the same training hyperparameters (detailed in the supplementary) instead of using the authors' public, fully-supervised model.
The original PSPNet implementation \cite{zhao_cvpr_2017} used a large batch size synchronised over 16 GPUs, as larger batch sizes give better estimates of batch statistics used for batch normalisation \cite{zhao_cvpr_2017,chen_arxiv_2017}.
In contrast, our experiments are performed on a single GPU with a batch size of one $521 \times 521$ image crop.
As a small batch size gives noisy estimates of batch statistics, our batch statistics are ``frozen'' to the values from the ImageNet-pretrained model as common practice \cite{chen_arxiv_2016,huang_cvpr_2017}.
Our instance subnetwork requires object detections, and we train Faster-RCNN~\cite{ren_2015} for this task.
All our networks use a ResNet-101 \cite{he_cvpr_2016} backbone.

\paragraph{Evaluation Metrics}
We use the $AP^{r}$ metric \cite{hariharan_2014}, commonly used in evaluating instance segmentation.
It extends the $AP$, a ranking metric used in object detection \cite{everingham_2010}, to segmentation where a predicted instance is considered correct if its Intersection over Union (IoU) with the ground truth instance is more than a certain threshold.
We also report the $AP^r_{vol}$ which is the mean $AP^r$ across a range of IoU thresholds.
Following the literature, we use a range of $0.1$ to $0.9$ in increments of $0.1$ on VOC, and $0.5$ to $0.95$ in increments of $0.05$ on Cityscapes.

However, as noted by several authors \cite{yang_2012,arnab_cvpr_2017,bai_cvpr_2017,kirillov_arxiv_2018}, the $AP^r$ is a ranking metric that does not penalise methods which predict more instances than there actually are in the image as long as they are ranked correctly.
Moreover, as it considers each instance independently, it does not penalise overlapping instances.
As a result, we also report the Panoptic Quality (PQ) recently proposed by \cite{kirillov_arxiv_2018},
\newcommand{\TP}{\mathit{TP}}
\newcommand{\FN}{\mathit{FN}}
\newcommand{\FP}{\mathit{FP}}
\newcommand{\IoU}{\text{IoU}}
\begin{equation}
\small{\text{PQ}} = \underbrace{\frac{\sum_{(p, g) \in \TP} \text{IoU}(p, g)}{\vphantom{\frac{1}{2}}|\TP|}}_{\text{Segmentation Quality (SQ)}} \times \underbrace{\frac{|\TP|}{|\TP| + \frac{1}{2} |\FP| + \frac{1}{2} |\FN|}}_{\text{Detection Quality (DQ) }} \,,
\end{equation}
where $p$ and $g$ are the predicted and ground truth segments, and $\TP$, $\FP$ and $\FN$ respectively denote the set of true positives, false positives and false negatives.

\subsection{Results on Pascal VOC}
\begin{table}[t]
\centering
\caption{Comparison of semantic segmentation performance to recent methods using only weak, bounding-box supervision on Pascal VOC. Note that \cite{dai_2015} and \cite{papandreou_2015} use the less accurate VGG network, whilst we and \cite{khoreva_cvpr_2017} use ResNet-101. ``FS\%'' denotes the percentage of fully-supervised performance.}
\label{tab:voc_semantic_comparison}
\scalebox{0.94}{
\begin{tabularx}{1.06\linewidth}{lYYYYYY}
	\toprule
	\multirow{2}{*}{Method} & \multicolumn{3}{c}{Validation set} & \multicolumn{3}{c}{Test set} \\ 
	& IoU (weak)   & IoU (full)   & FS\%   & IoU (weak) & IoU (full) & FS\% \\
	\cmidrule(r){1-1} \cmidrule{2-4} \cmidrule(l){5-7}
	\multicolumn{7}{l}{\textit{Without COCO annotations}} \\
	BoxSup \cite{dai_2015}    &   62.0    &  63.8    &  \textbf{97.2} 
	                          &   64.6    &  --      &  --    \\
	Deeplab WSSL \cite{papandreou_2015}
	& 60.6        &  67.6     &   89.6   
	& 62.2        &  70.3     &   88.5 \\
	SDI \cite{khoreva_cvpr_2017} & 69.4 & 74.5 & 93.2 & -- & -- & -- \\
	Ours    &   \textbf{74.3}    &   \textbf{77.3}    &   96.1 & \textbf{75.5} & \textbf{78.6} & \textbf{96.3}\\
	\cmidrule(r){1-1} \cmidrule{2-4} \cmidrule(l){5-7}
	\multicolumn{7}{l}{\textit{With COCO annotations}} \\
	SDI \cite{khoreva_cvpr_2017} & 74.2 & 77.7 & 95.5 & -- & -- & --\\
	Ours    &   \textbf{75.7}    &   \textbf{79.0}    &   \textbf{95.8} & \textbf{76.7} & \textbf{79.4} & \textbf{96.6}\\
	 \bottomrule
\end{tabularx}
}
\\
\end{table}

\begin{table}[t]
\centering
\caption{Comparison of instance segmentation performance to recent (fully- and weakly-supervised) methods on the VOC 2012 validation set.}
\label{tab:voc_instance_comparison}
\begin{tabularx}{0.9\linewidth}{lYYYYYYY}
\toprule
\multirow{2}{*}{Method} & \multicolumn{5}{c}{$AP^r$} & \multirow{2}{*}{$AP^r_{vol}$} & \multirow{2}{*}{PQ}\\
                               & 0.5    & 0.6    & 0.7    & 0.8   & 0.9   &                                            \\
\midrule
\textit{Weakly supervised without COCO}\\
SDI \cite{khoreva_cvpr_2017} & 44.8 & -- & -- & -- & -- & -- & -- \\
Ours & \textbf{60.5} & \textbf{55.2} & \textbf{47.8} & \textbf{37.6} & \textbf{21.6} & \textbf{55.6} & \textbf{59.0} \\
\midrule
\textit{Fully supervised without COCO}\\
SDS \cite{hariharan_2014}                            & 43.8   & 34.5   & 21.3   & 8.7   & 0.9   &   --  & --                                       \\
Chen \etal \cite{chen_cvpr_2015}                      & 46.3   & 38.2   & 27.0   & 13.5  & 2.6   &   --    & --                                     \\
PFN \cite{liang_arxiv_2015}                           & 58.7   & 51.3   & 42.5   & 31.2  & 15.7  & 52.3  & --                                     \\
Ours (fully supervised) & \textbf{63.6} & \textbf{59.5} & \textbf{53.8} & \textbf{44.7} & \textbf{30.2} & \textbf{59.2} & \textbf{62.7}\\
\toprule
\textit{Weakly supervised with COCO}\\
SDI \cite{khoreva_cvpr_2017} & 46.4 & -- & -- & -- & -- & -- & --\\
Ours & \textbf{60.9} & \textbf{55.9} & \textbf{48.0} & \textbf{37.2} & \textbf{21.7} & \textbf{55.5} & \textbf{59.5} \\
\midrule          
\textit{Fully supervised with COCO} \\
Arnab \etal \cite{arnab_bmvc_2016}                     & 58.3   & 52.4   & 45.4   & 34.9  & 20.1  & 53.1 & --                                      \\
MPA \cite{liu_cvpr_2016}                   & 62.1   & 56.6   & 47.4   & 36.1  & 18.5  & 56.5  & --                                     \\
Arnab \etal \cite{arnab_cvpr_2017} & 61.7 & 55.5 & 48.6 & 39.5 & 25.1 & 57.5 & --\\
SGN \cite{liu_iccv_2017} & 61.4 & 55.9 & 49.9 & 42.1 & 26.9 & -- & --\\
Ours (fully supervised) & \textbf{63.9} & \textbf{59.3} & \textbf{54.3} & \textbf{45.4} & \textbf{30.2} & \textbf{59.5} & \textbf{63.1}\\
\bottomrule                                 
\end{tabularx}
\end{table}
Tables~\ref{tab:voc_semantic_comparison} and \ref{tab:voc_instance_comparison} show the state-of-art results of our method for semantic- and instance-segmentation respectively.
For both semantic- and instance-segmentation, our weakly supervised model obtains about 95\% of the performance of its fully-supervised counterpart, emphasising that accurate models can be learned from only bounding box annotations, which are significantly quicker and cheaper to obtain than pixelwise annotations.
Table \ref{tab:voc_instance_comparison} also shows that our weakly-supervised model outperforms some recent fully supervised instance segmentation methods such as \cite{arnab_bmvc_2016} and \cite{liang_arxiv_2015}.
Moreover, our fully-supervised instance segmentation model outperforms all previous work on this dataset.
The main difference of our model to \cite{arnab_cvpr_2017} is that our network is based on the PSPNet architecture using ResNet-101, whilst \cite{arnab_cvpr_2017} used the network of \cite{arnab_eccv_2016} based on VGG \cite{simonyan_iclr_2015}.

We can obtain semantic segmentations from the output of our semantic subnetwork, or from the final instance segmentation (as we produce non-overlapping instances) by taking the union of all instances which have the same semantic label.
We find that the IoU obtained from the final instance segmentation, and the initial pretrained semantic subnetwork to be very similar, and report the latter in Tab.\ref{tab:voc_semantic_comparison}.
Further qualitative and quantitative results, including success and failure cases, are included in the supplement.

\paragraph{End-to-end training of instance subnetwork}
Our instance subnetwork can be trained in a piecewise fashion, or the entire network including the semantic subnetwork can be trained end-to-end.
End-to-end training was shown to obtain higher performance by \cite{arnab_cvpr_2017} for full supervision.
We also observe this effect for weak supervision from bounding box annotations.
A weakly supervised model, trained with COCO annotations improves from an $AP^{r}_{vol}$ of 53.3 to 55.5.
When not using COCO for training the initial semantic subnetwork, a slightly higher increase by 3.9 from 51.7 is observed.
This emphasises that our training strategy (Sec.~\ref{sec:gt_intro}) is effective for both semantic- and instance-segmentation.

\paragraph{Iterative training} The approximate ground truth used to train our model can also be generated in an iterative manner, as discussed in Sec.~\ref{sec:gt_iterative}.
However, as the results from a single iteration (Tab.~\ref{tab:voc_semantic_comparison} and \ref{tab:voc_instance_comparison}) are already very close to fully-supervised performance, this offers negligible benefit.
Iterative training is, however, crucial for obtaining good results on Cityscapes as discussed in Sec.~\ref{sec:results_cityscapes}.

\paragraph{Semi-Supervision}
We also consider the case where we have a combination of weak and full annotations.
As shown in Tab.~\ref{tab:voc_semi_supervised}, we consider all combinations of weak- and full-supervision of the training data from Pascal VOC and COCO.
Table \ref{tab:voc_semi_supervised} shows that training with fully-supervised data from COCO and weakly-supervised data from VOC performs about the same as weak supervision from both datasets for both semantic- and instance-segmentation.
Furthermore, training with fully annotated VOC data and weakly labelled COCO data obtains similar results to full supervision from both datasets.
We have qualitatively observed that the annotations in Pascal VOC are of higher quality than those of Microsoft COCO (random samples from both datasets are shown in the supplementary).
And this intuition is evident in the fact that there is not much difference between training with weak or full annotations from COCO.
This suggests that in the case of segmentation, per-pixel labelling of additional images is not particularly useful if they are not labelled to a high standard, and that labelling fewer images at a higher quality (Pascal VOC) is more beneficial than labelling many images at a lower quality (COCO).
This is because Tab.~\ref{tab:voc_semi_supervised} demonstrates how both semantic- and instance-segmentation networks can be trained to achieve similar performance by using only bounding box labels instead of low-quality segmentation masks.
The average annotation time can be considered a proxy for segmentation quality. 
While a COCO instance took an average of 79 seconds to segment \cite{lin_2014}, this figure is not mentioned for Pascal VOC \cite{everingham_2010,everingham_ijcv_2015}.

\begin{table}[t]
	\parbox{.48\linewidth}{
		\centering
		
		\caption{Semantic- and instance-segmentation performance on Pascal VOC with varying levels of supervision from the Pascal and COCO datasets.
		The former is measured by the IoU, and latter by the $AP^r_{vol}$ and PQ.
		}
		\label{tab:voc_semi_supervised}
		\begin{tabularx}{1\linewidth}{XXYYY}
			\toprule
			\multicolumn{2}{c}{Dataset} & \multirow{2}{*}{IoU} & \multirow{2}{*}{$AP^{r}_{vol}$} & \multirow{2}{*}{PQ} \\ 
			VOC          & COCO         &                      &                     &                                                                  \\
			\cmidrule(r){1-2} \cmidrule{3-5}
			Weak         & Weak         &   75.7   & 55.5      &    59.5                                                               \\
			Weak         & Full         &  75.8   & 56.1                    &  59.8   
			\\
			Full         & Weak         &   77.5   &  58.9                   &    62.7                                                               \\
			Full         & Full         &   79.0   & 59.5                     &      63.1                     \\                                       
			\bottomrule
		\end{tabularx}

	}
	\hfill
	\parbox{.48\linewidth}{
		\centering
		\caption{
			Semantic segmentation performance on the Cityscapes validation set. 
			We use more informative, bounding-box annotations for ``thing'' classes, and this is evident from the higher IoU than on ``stuff'' classes for which we only have image-level tags.
		}
		\label{tab:cityscapes_semantic_comparison}
		\begin{tabularx}{1\linewidth}{lYYY}
			\toprule
			Method & IoU (weak)   & IoU (full)   & FS\%   \\
			\midrule
			Ours (thing classes) & 68.2 & 70.4 & 96.9 \\
			Ours (stuff classes)  & 60.2 & 72.4 & 83.1 \\
			Ours (overall)        & 63.6 & 71.6 & 88.8 \\
			\bottomrule
		\end{tabularx}
	}
\end{table}

\subsection{Results on Cityscapes}
\label{sec:results_cityscapes}
Tables \ref{tab:cityscapes_semantic_comparison} and \ref{tab:cityscapes_instance_comparison} present, what to our knowledge is, the first weakly supervised results for either semantic or instance segmentation on Cityscapes.
Table \ref{tab:cityscapes_semantic_comparison} shows that, as expected for semantic segmentation, our weakly supervised model performs better, relative to the fully-supervised model, for ``thing'' classes compared to ``stuff'' classes.
This is because we have more informative bounding box labels for ``things'', compared to only image-level tags for ``stuff''.
For semantic segmentation, we obtain about 97\% of fully-supervised performance for ``things'' (similar to our results on Pascal VOC) and 83\% for ``stuff''.
Note that we evaluate images at a single-scale, and higher absolute scores could be obtained by multi-scale ensembling \cite{zhao_cvpr_2017,chen_arxiv_2016}.

For instance-level segmentation, the fully-supervised ratios for the PQ are similar to the IoU ratio for semantic segmentation.
In Tab. \ref{tab:cityscapes_instance_comparison}, we report the $AP^r_{vol}$ and PQ for both thing and stuff classes, assuming that there is only one instance of a ``stuff'' class in the image if it is present.
\begin{table}[t]
\centering
\caption{Instance-level segmentation results on Cityscapes. On the validation set, we report results for both ``thing'' (th.) and ``stuff'' (st.) classes. The online server, which evaluates the test set, only computes the $AP^{r}$ for ``thing'' classes.
We compare to other fully-supervised methods which produce non-overlapping instances.
To our knowledge, no published work has evaluated on both ``thing'' and ``stuff'' classes.
Our fully supervised model, initialised from the public PSPNet model \cite{zhao_cvpr_2017} is equivalent to our previous work \cite{arnab_cvpr_2017}, and competitive with the state-of-art.
Note that we cannot use the public PSPNet pretrained model in a weakly-supervised setting.
}
\label{tab:cityscapes_instance_comparison}
\scalebox{0.9}{
	\begin{tabularx}{1.07\linewidth}{lYYYYYYYYYY}
		\toprule
		& \multicolumn{9}{c}{Validation}                                & Test      \\
		& \multicolumn{3}{c}{$AP^{r}_{vol}$} & \multicolumn{3}{c}{PQ} & \multicolumn{3}{c}{IoU} & $AP^{r}_{vol}$ \\
		Method & th. & st. & all & th. & st. & all & th. & st. & all & th. \\
		\cmidrule(r){1-1} \cmidrule(r){2-4} \cmidrule(r){5-7} \cmidrule(l){8-10} \cmidrule(l){11-11}
		Ours (weak, ImageNet init.) &   17.0    &    33.1      &   26.3     &   35.8        &   43.9       &  40.5      &    68.2     &    60.2     &    63.6     &    12.8      \\
		Ours (full, ImageNet init.) &   24.3    &    42.6     &   34.9      &    39.6       & 52.9         &  47.3      &   70.4     &    72.4     &    71.6     &    18.8        \\
		\cmidrule(r){1-1} \cmidrule(r){2-4} \cmidrule(r){5-7} \cmidrule(l){8-10} \cmidrule(l){11-11}
		Ours (full, PSPNet init.) \cite{arnab_cvpr_2017} & 28.6 &   52.6    &   42.5    &   42.5    &   62.1    &   53.8    &   80.1     &    79.5     &    79.8     &    23.4    \\
		\cmidrule(r){1-1} \cmidrule(r){2-4} \cmidrule(r){5-7} \cmidrule(l){8-10} \cmidrule(l){11-11}
		Pixel Encoding \cite{uhrig_gcpr_2016} &  9.9  &   --       &   --     &   --        &    --      &   --     &    --     &    --     &    --     &    8.9      \\
		RecAttend \cite{ren_cvpr_2017} & --  &   --       &   --     &   --        &    --      &   --     & --     &    --     &    --     &   9.5       \\
		InstanceCut \cite{kirillov_cvpr_2017} & --   &   --       &   --     &   --        &    --      &   --     &   --     &    --     &    --     &    13.0      \\
		DWT \cite{bai_cvpr_2017} & 21.2   &   --       &   --     &   --        &    --      &   --     &  --     &    --     &    --     &    19.4       \\
		SGN \cite{liu_iccv_2017} &  29.2  &   --       &   --     &   --        &    --      &   --     &    --     &    --     &    --     &    25.0      \\
		\bottomrule
	\end{tabularx}
}
\end{table}

Here, the $AP^{r}_{vol}$ for ``stuff'' classes is higher than that for ``things''.
This is because there can only be one instance of a ``stuff'' class, which makes instances easier to detect, particularly for classes such as ``road'' which typically occupy a large portion of the image.
The Cityscapes evaluation server, and previous work on this dataset, only report the $AP^{r}_{vol}$ for ``thing'' classes.
As a result, we report results for ``stuff'' classes only on the validation set.
Table \ref{tab:cityscapes_instance_comparison} also compares our results to existing work which produces non-overlapping instances on this dataset, and shows that both our fully- and weakly-supervised models are competitive with recently published work on this dataset.
We also include the results of our fully-supervised model, initialised from the public PSPNet model \cite{zhao_cvpr_2017} released by the authors, and show that this is competitive with the state-of-art \cite{liu_iccv_2017} among methods producing non-overlapping segmentations (note that \cite{liu_iccv_2017} also uses the same PSPNet model).
Figure 7 shows some predictions of our weakly supervised model; further 
results are in the supplementary.

\paragraph{Iterative training}
\begin{figure}[t]
\centering

\begin{tabularx}{\linewidth}{ cc}

\includegraphics[width=0.49\linewidth]{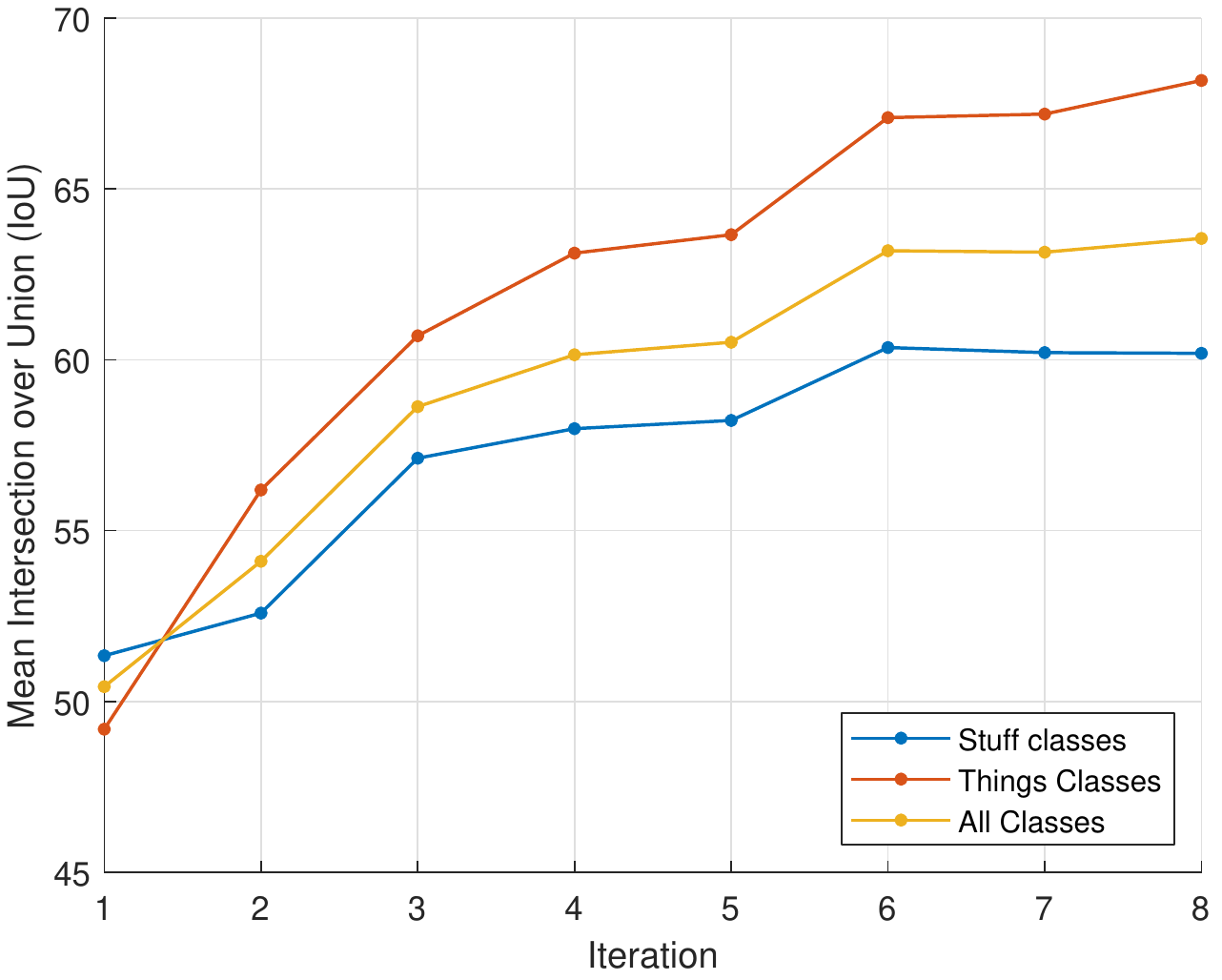}
&
\includegraphics[width=0.49\linewidth]{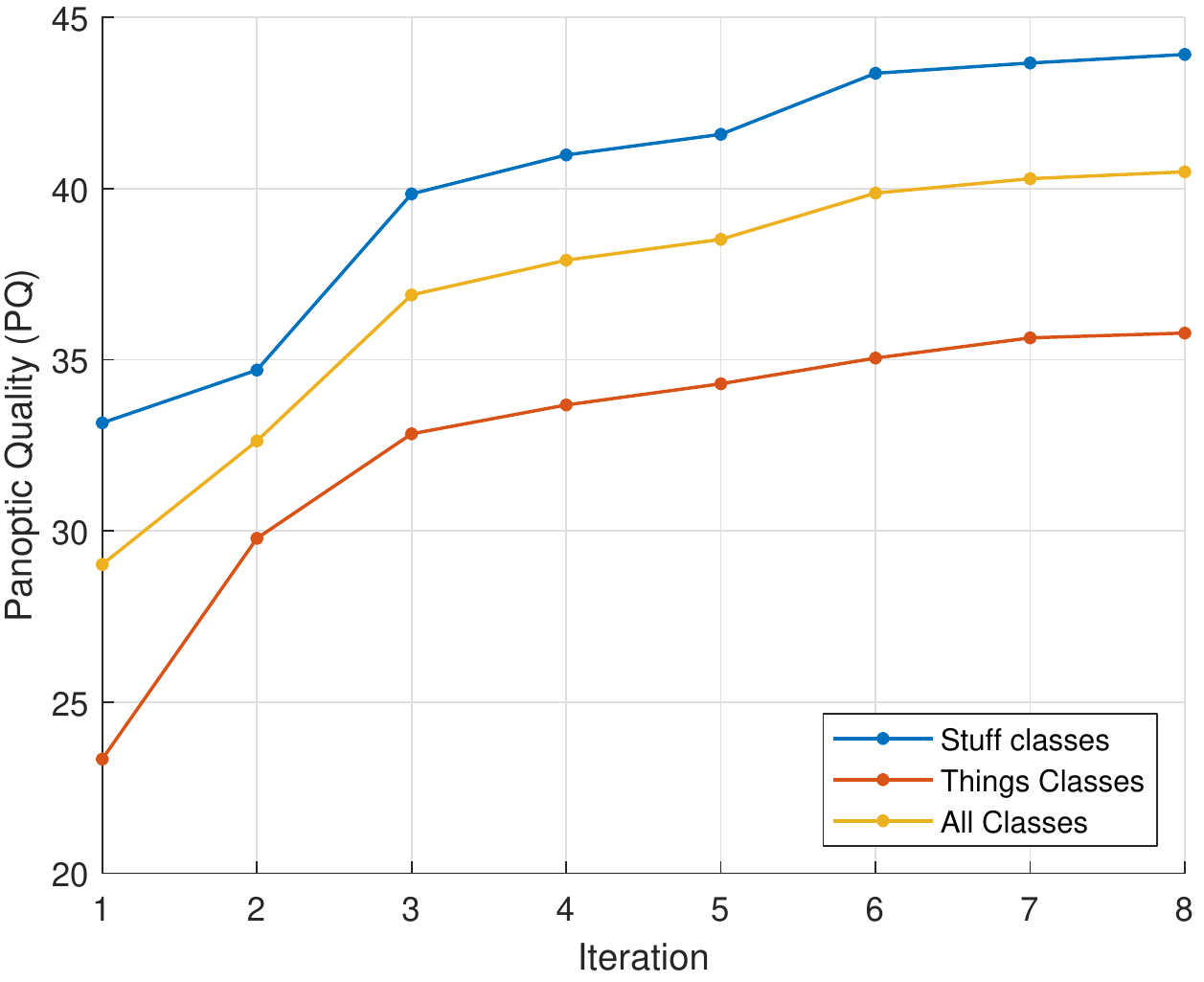}          
\\
(a) Semantic segmentation (IoU) & (b) Instance segmentation (PQ) \\

\end{tabularx}

\setlength{\tabcolsep}{6pt}

\caption{
	Iteratively refining our approximate ground truth during training improves both semantic and instance segmentation on the Cityscapes validation set.}
\label{fig:iterative_training}
\end{figure}
Iteratively refining our approximate ground truth during training, as described in Sec.~\ref{sec:gt_iterative}, greatly improves our performance on both semantic- and instance-segmentation as shown in Fig.~\ref{fig:iterative_training}.
We trained the network for 150 000 iterations before regenerating the approximate ground truth using the network's own output on the training set.
Unlike on Pascal VOC, iterative training is necessary to obtain good performance on Cityscapes as the approximate ground truth generated on the first iteration is not sufficient to obtain high accuracy.
This was expected for ``stuff'' classes, since we began from weak localisation cues derived from the image-level tags.
However, as shown in Fig.~\ref{fig:iterative_training}, ``thing'' classes also improved substantially with iterative training, unlike on Pascal VOC where there was no difference.
Compared to VOC, Cityscapes is a more cluttered dataset, and has large scale variations as the distance of an object from the car-mounted camera changes.
These dataset differences may explain why the image priors employed by the methods we used (GrabCut \cite{rother_2004} and MCG \cite{arbelaez_2014}) to obtain approximate ground truth annotations from bounding boxes are less effective.
Furthermore, in contrast to Pascal VOC, Cityscapes has frequent co-occurences of the same objects in many different images, making it more challenging for weakly supervised methods.

\paragraph{Effect of ranking methods on the $AP^{r}$}
The $AP^{r}$ metric is a ranking metric derived from object detection.
It thus requires predicted instances to be scored such that they are ranked in the correct relative order.
As our network uses object detections as an additional input and each detection represents a possible instance, we set the score of a predicted instance to be equal to the object detection score.
For the case of stuff classes, which object detectors are not trained for, we use a constant detection score of 1 as described in Sec.~\ref{sec:model}.
An alternative to using a constant score for ``stuff'' classes is to take the mean of the softmax-probability of all pixels within the segmentation mask.
Table \ref{tab:scoring} shows that this latter method improves the $AP^r$ for stuff classes.
For ``things'', ranking with the detection score performs better and comes closer to oracle performance which is the maximum $AP^r$ that could be obtained with the predicted instances.

\begin{table}[t]
	\parbox{.48\linewidth}{

	\caption{The effect of different instance ranking methods on the $AP^{r}_{vol}$ of our weakly supervised model computed on the Cityscapes validation set.}
	\label{tab:scoring}
	\begin{tabularx}{1\linewidth}{Xccc}
		\toprule
		Ranking Method          & $AP^{r}_{vol}$ th. & $AP^{r}_{vol}$ st. & PQ all \\ \midrule
		Detection score         & 17.0      & 26.7     & 40.5   \\
		Mean seg. confidence & 14.6      & 33.1     & 40.5   \\
		Oracle                  & 21.6      & 37.0     & 40.5   \\ \bottomrule
	\end{tabularx}

	}
	\hfill
	\parbox{.48\linewidth}{
			\centering
			\label{my-label}
			\begin{tabular}{cc}
				\includegraphics[width=0.69\linewidth]{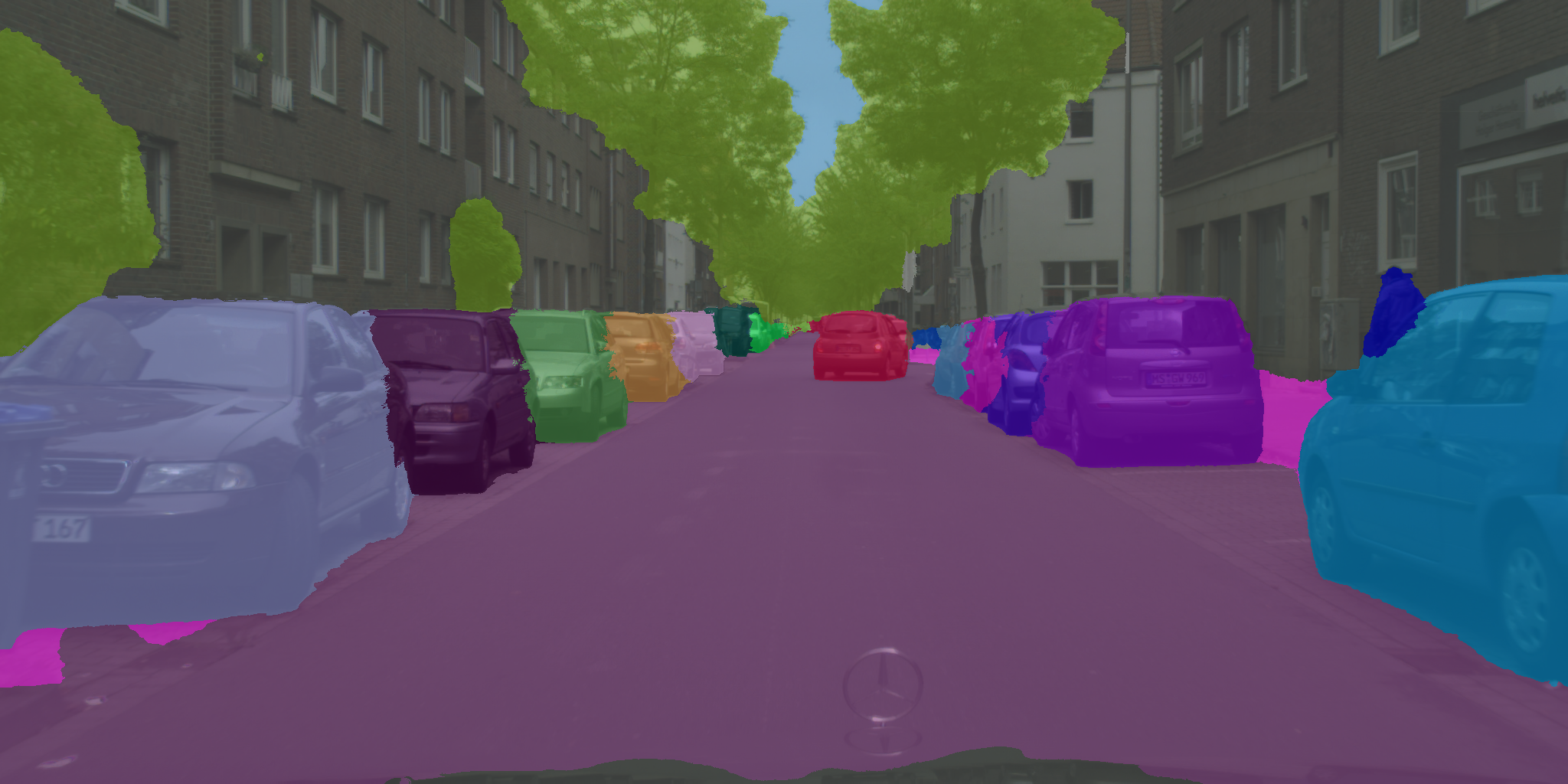} \\
				\includegraphics[width=0.69\linewidth]{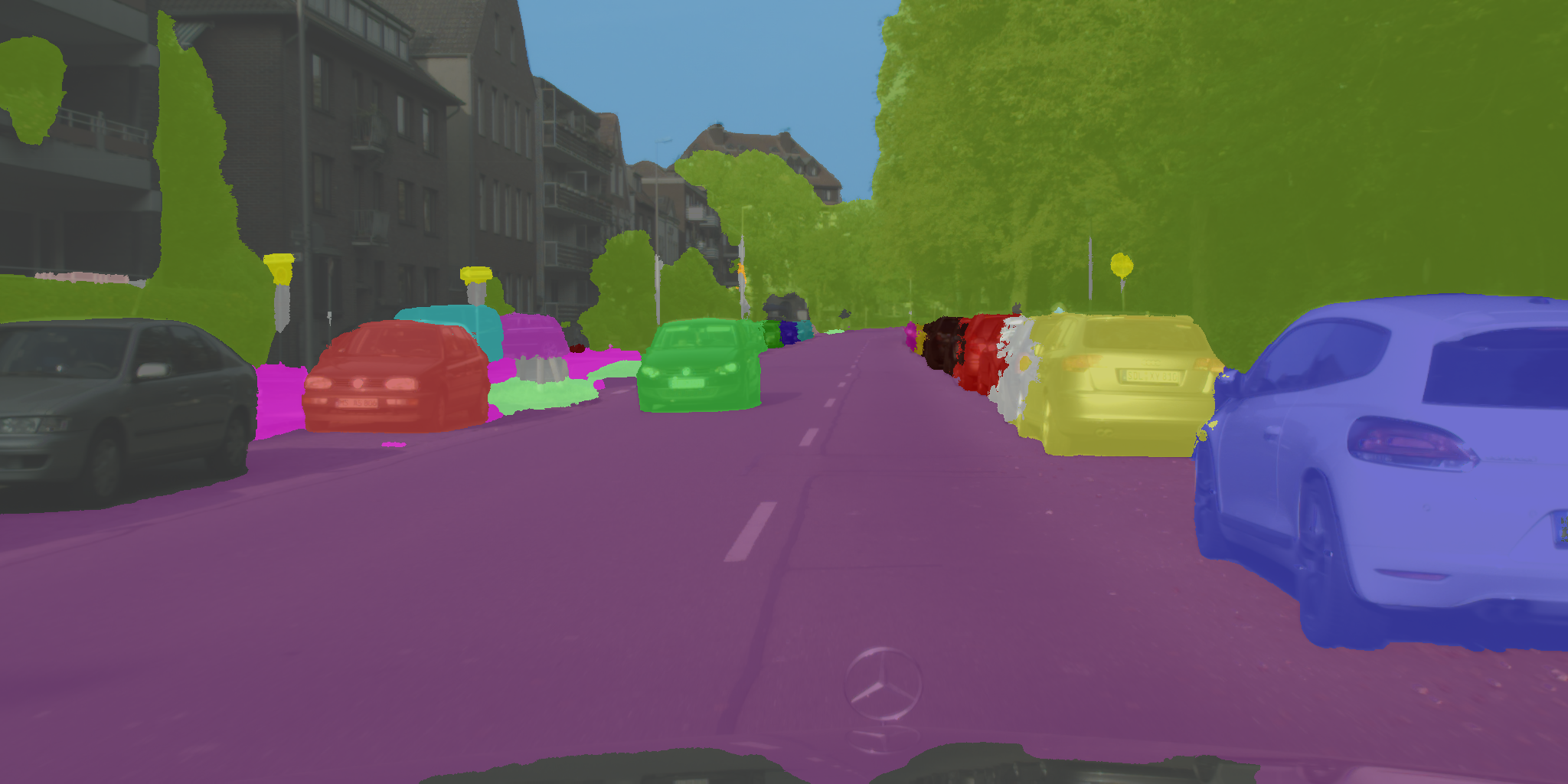}
			\end{tabular}
			\\
			\textbf{Fig. 7.} Example results on Cityscapes of our weakly supervised model.
	}
\end{table}

Changing the score of a segmented instance does not change the quality of the actual segmentation, but does impact the $AP^r$ greatly as shown in Tab.~\ref{tab:scoring}.
The PQ, which does not use scores, is unaffected by different ranking methods, and this suggests that it is a better metric for evaluating non-overlapping instance segmentation where each pixel in the image is explained.
\section{Conclusion and Future Work}

We have presented, to our knowledge, the first weakly-supervised method that jointly produces non-overlapping instance and semantic segmentation for both ``thing'' and ``stuff'' classes.
Using only bounding boxes, we are able to achieve 95\% of state-of-art fully-supervised performance on Pascal VOC.
On Cityscapes, we use image-level annotations for ``stuff'' classes and obtain 88.8\% of fully-supervised performance for semantic segmentation and 85.6\% for instance segmentation (measured with the PQ).
Crucially, the weak annotations we use incur only about 3\% of the time of full labelling.
As annotating pixel-level segmentation is time consuming, there is a dilemma between labelling few images with high quality or many images with low quality.
Our semi-supervised experiment suggests that the latter is not an effective use of annotation budgets as similar performance can be obtained from only bounding-box annotations.

Future work is to perform instance segmentation using only image-level tags and the number of instances of each object present in the image as supervision.
This will require a network architecture that does not use object detections as an additional input.

\subsubsection*{Acknowledgements}
This work was supported by Huawei Technologies Co., Ltd., the EPSRC, Clarendon Fund, ERC grant ERC-2012-AdG 321162-HELIOS, EPRSRC grant Seebibyte EP/M013774/1 and EPSRC/MURI grant EP/N019474/1.



\bibliographystyle{splncs}
\bibliography{bibliography}
\clearpage
\appendix

\section*{Appendix}
Section~\ref{sec:additional_results} presents further qualitative and quantitative results of our experiments on Cityscapes and Pascal VOC.
Section~\ref{sec:experimental_details} describes the training of the networks described in the main paper.
Section 4.2 of our paper mentioned that the annotation quality of Pascal VOC \cite{everingham_2010} is better than COCO \cite{lin_2014}.
Some randomly drawn images from these datasets are presented to illustrate this point in Sec.~\ref{sec:pascal_vs_coco_gt}.
Finally, Sec.~\ref{sec:annotation_time} shows our calculation of how much the overall annotation time is reduced by using weak annotations, in comparison to full annotations, on the Cityscapes dataset.
\section{Additional Qualitative and Quantitative Results}
\label{sec:additional_results}

\begin{figure}[t]
	\centering

	\begin{tabularx}{\linewidth}{ Y Y Y}
	Input image & Weakly-supervised model & Fully-supervised model \\		
		
	\global \def \im{frankfurt_000000_003357}
	\includegraphics[width=\linewidth]{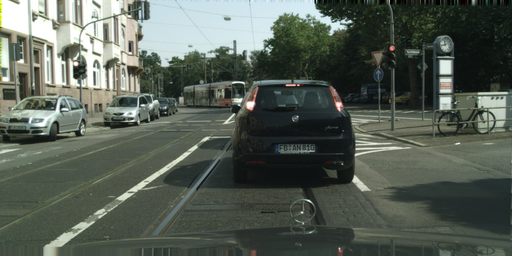} &
	\includegraphics[width=\linewidth]{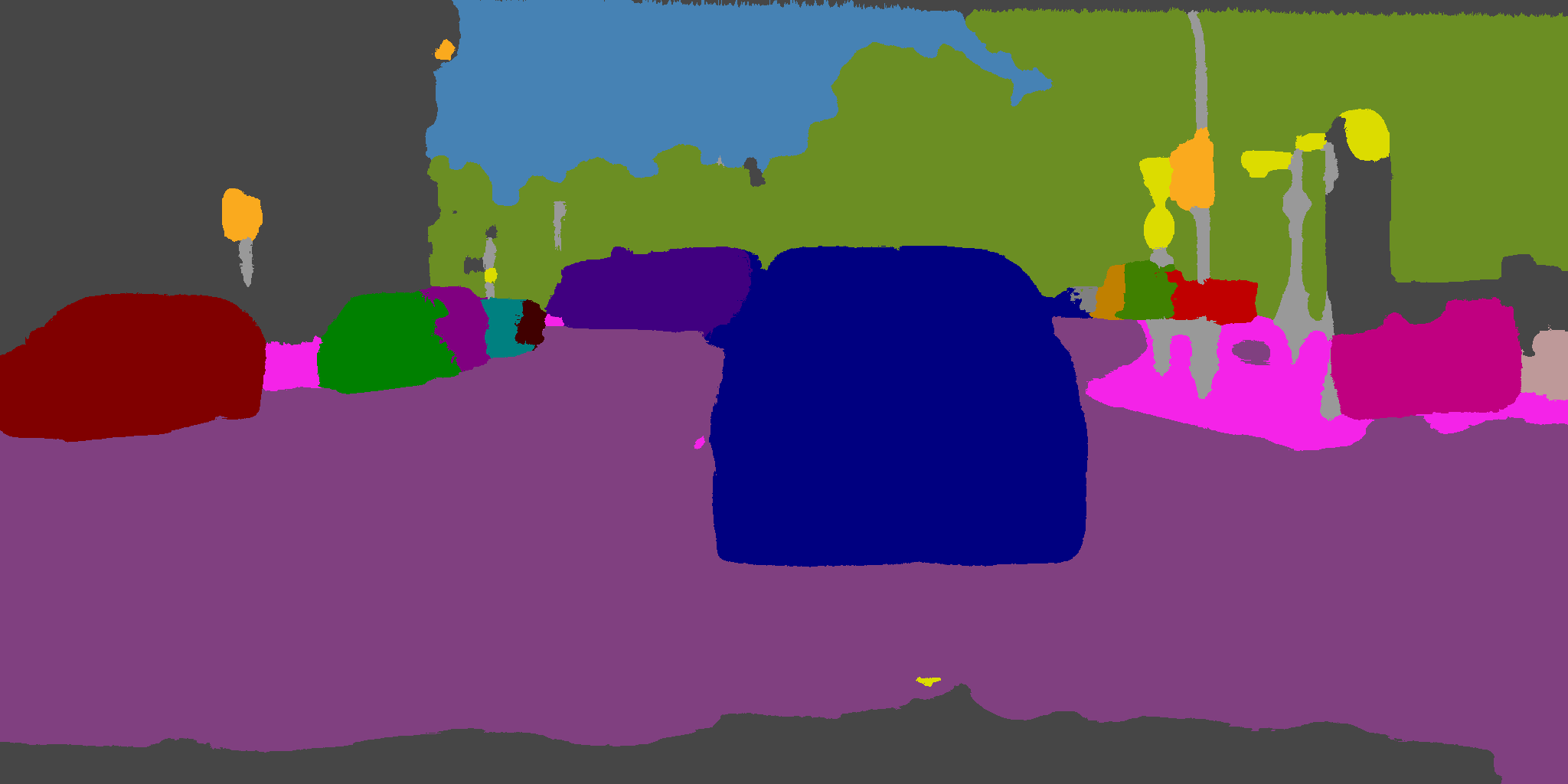} &
	\includegraphics[width=\linewidth]{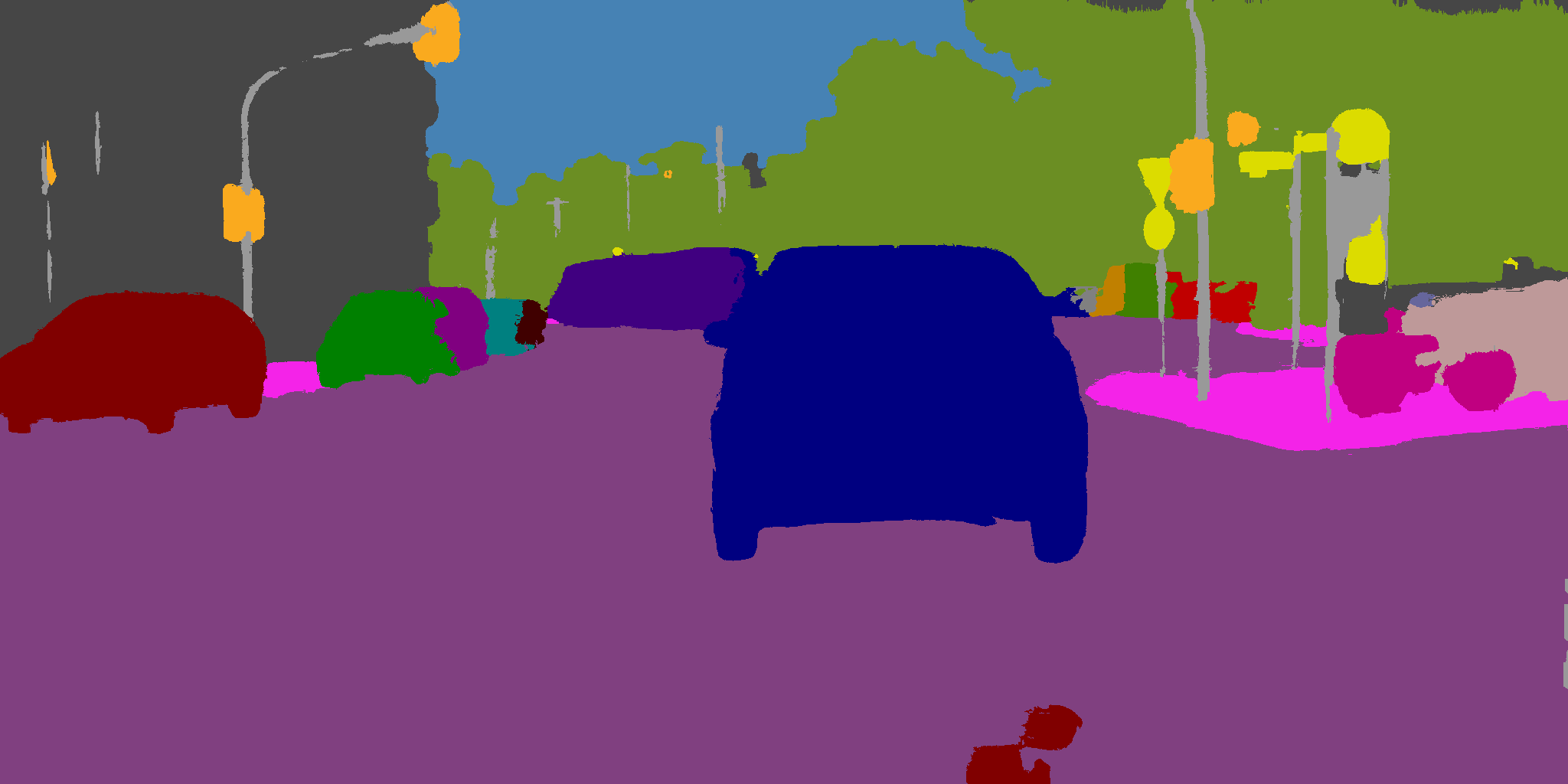} \\
	
	\global \def \im{munster_000172_000019}
	\includegraphics[width=\linewidth]{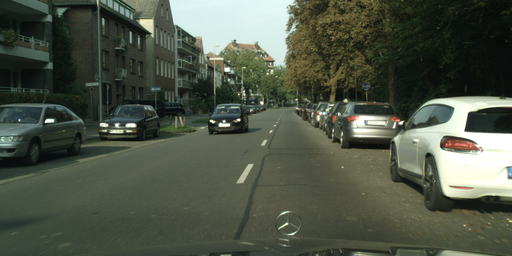} &
	\includegraphics[width=\linewidth]{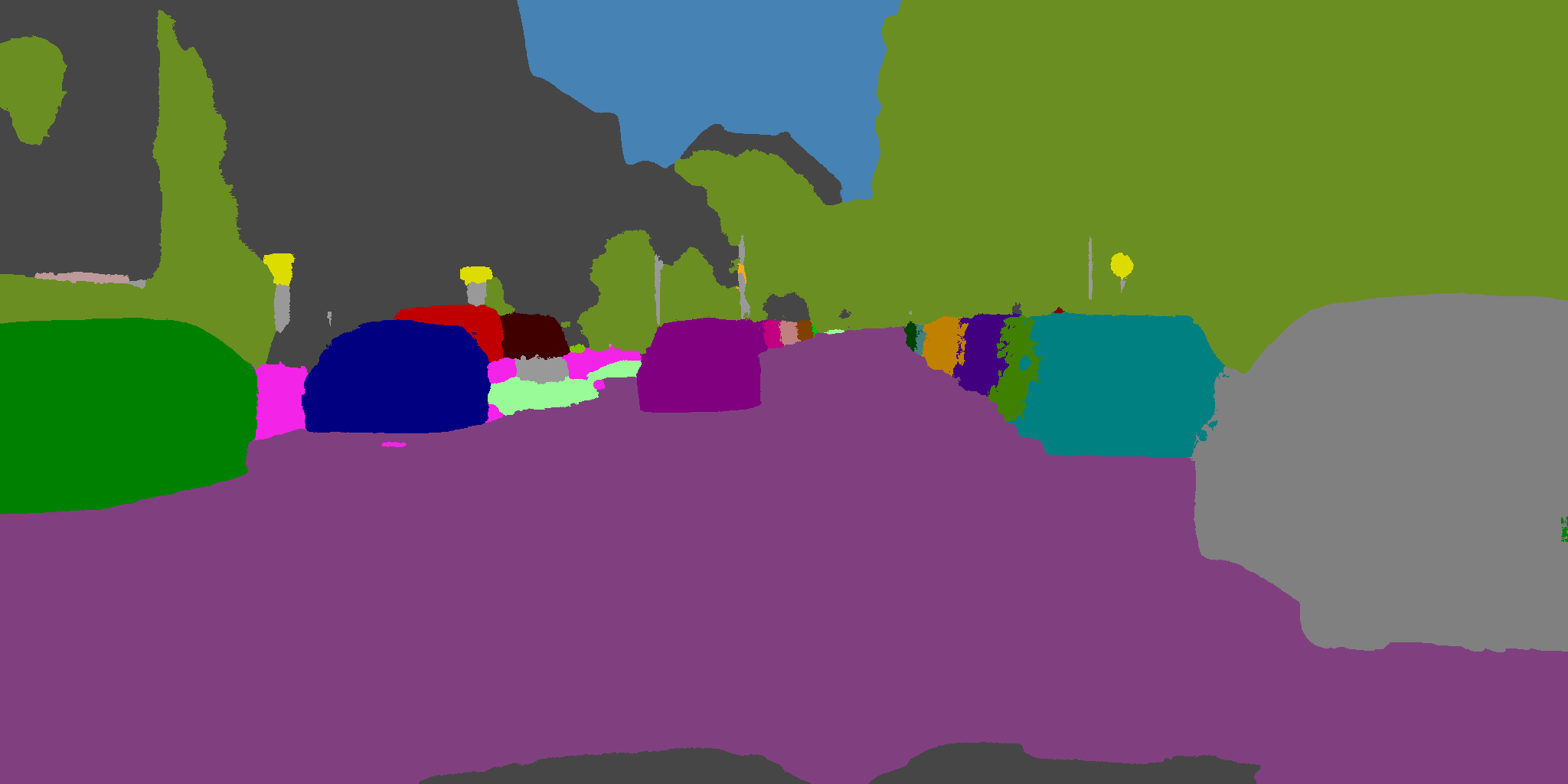} &
	\includegraphics[width=\linewidth]{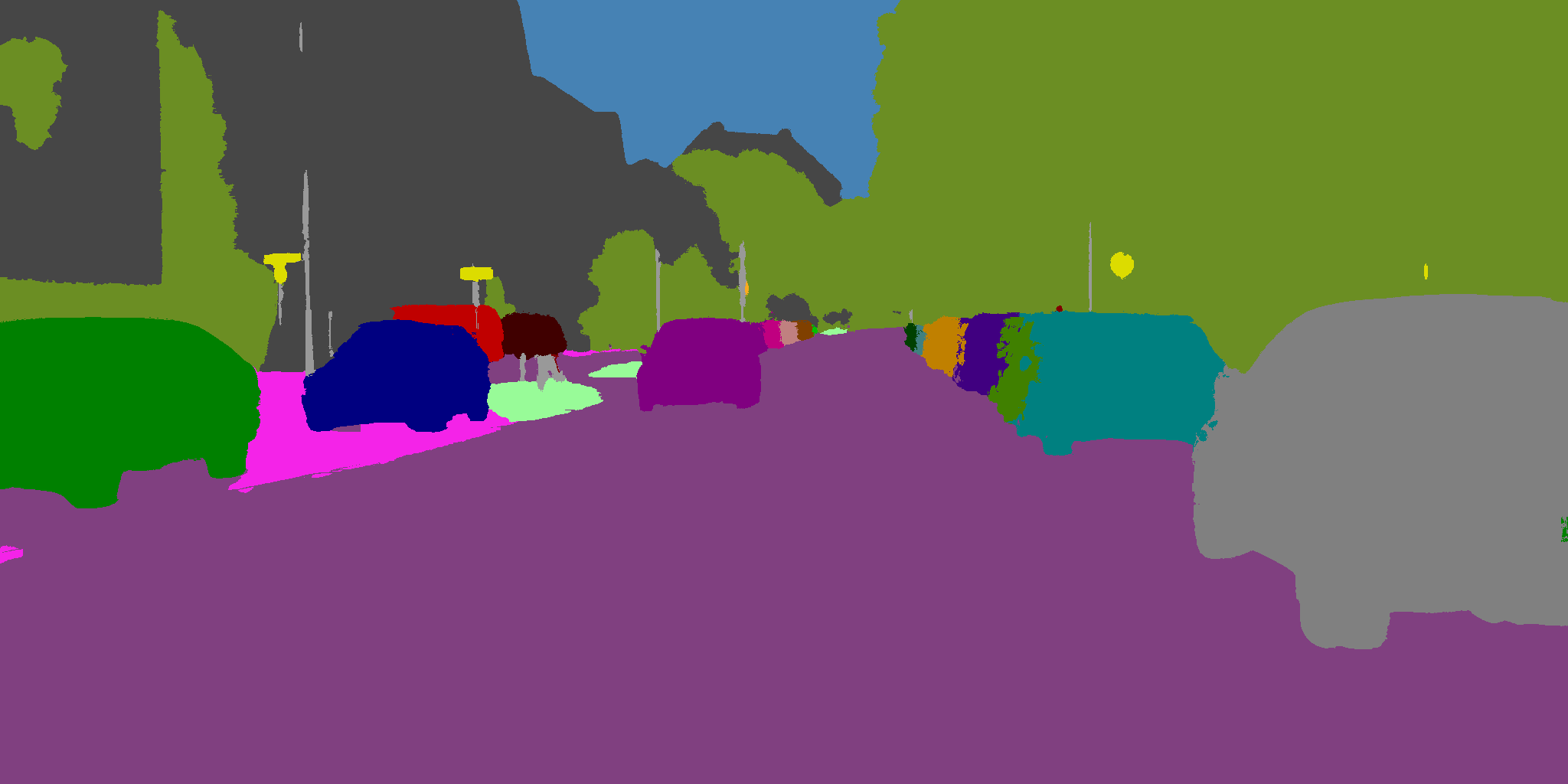} \\
	
	\global \def \im{munster_000057_000019}
	\includegraphics[width=\linewidth]{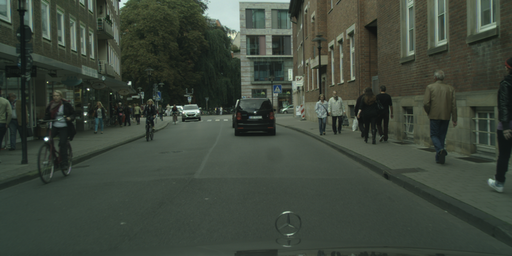} &
	\includegraphics[width=\linewidth]{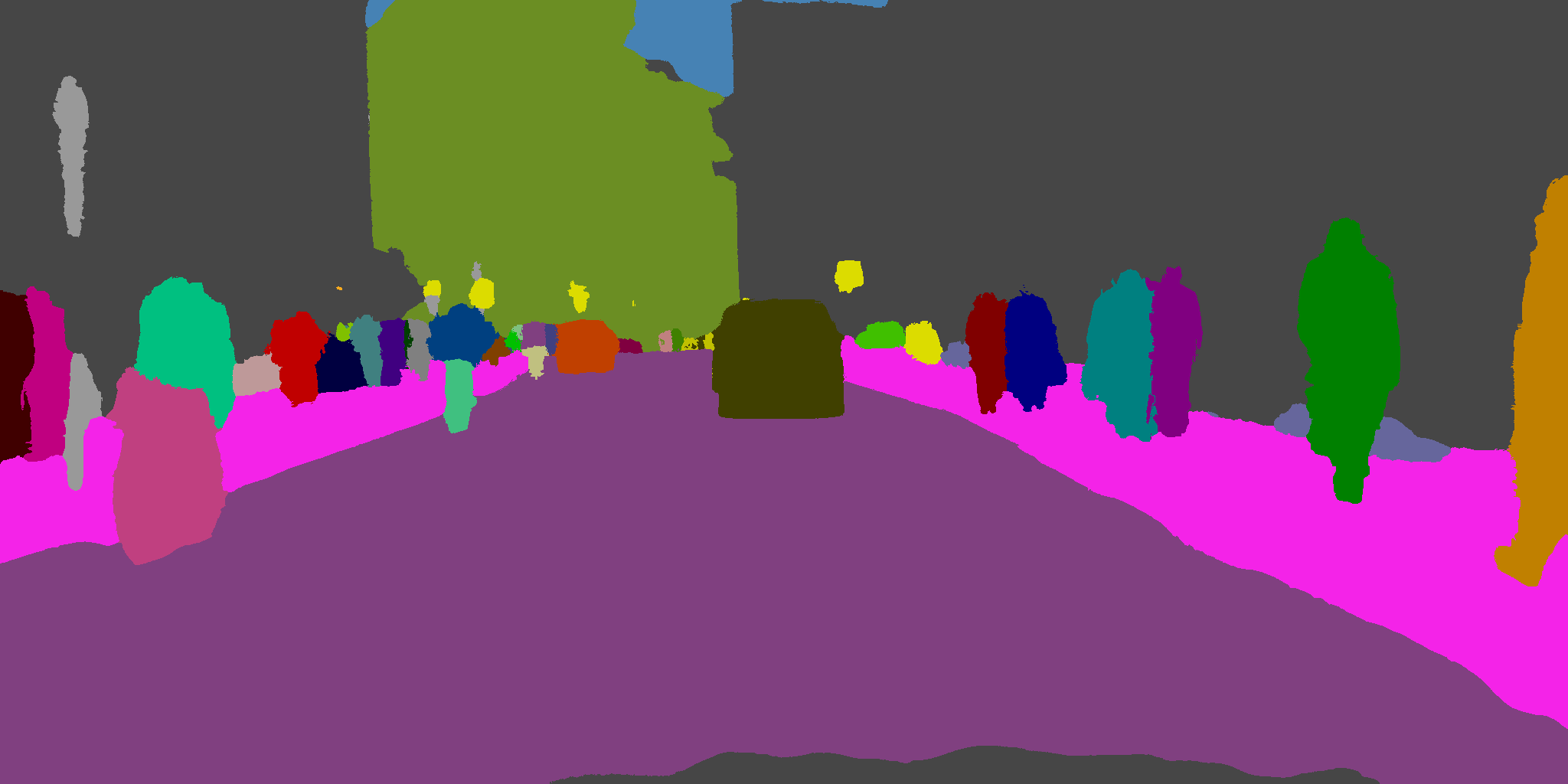} &
	\includegraphics[width=\linewidth]{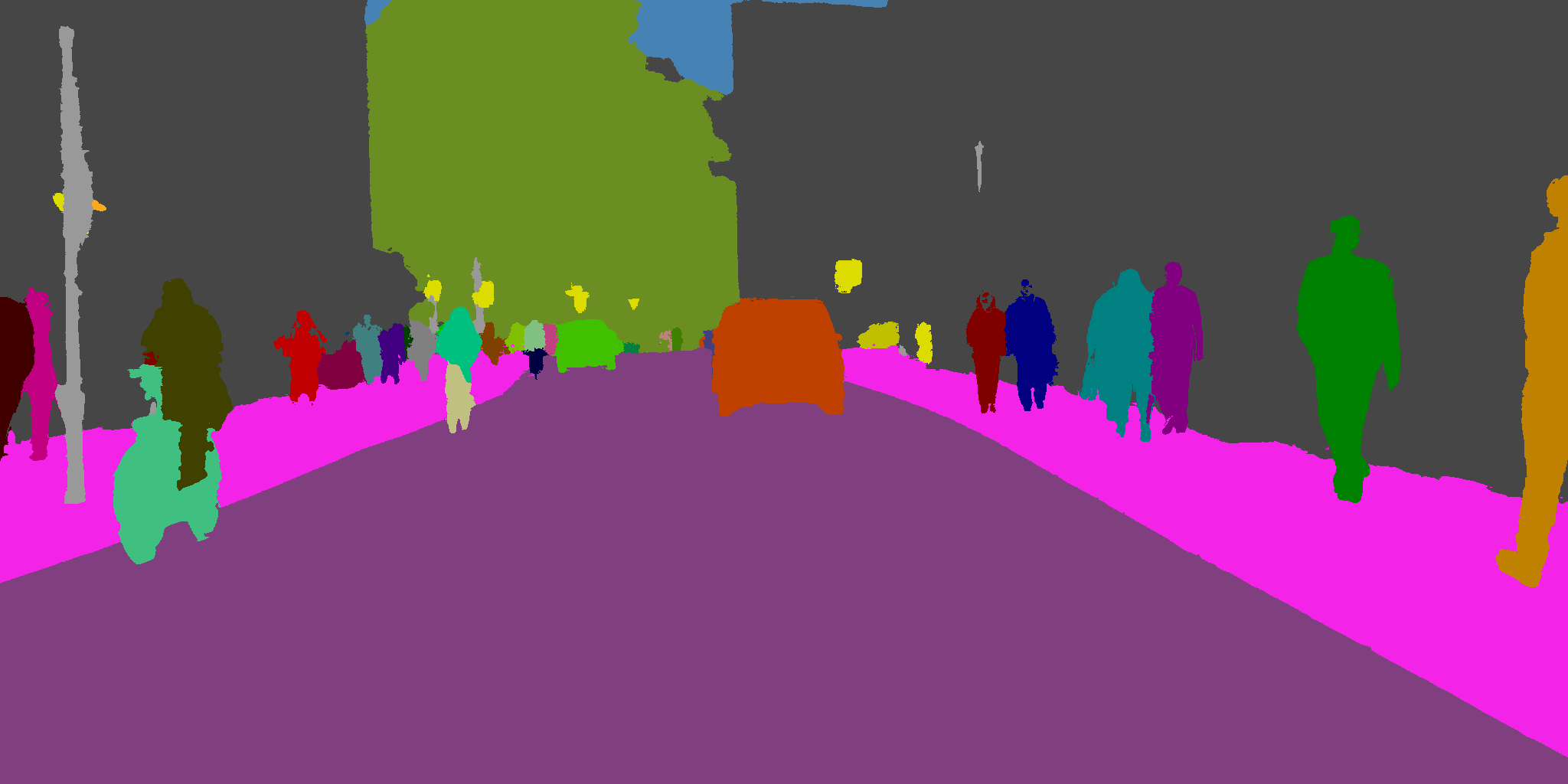} \\
	
	\global \def \im{frankfurt_000001_030310}
	\includegraphics[width=\linewidth]{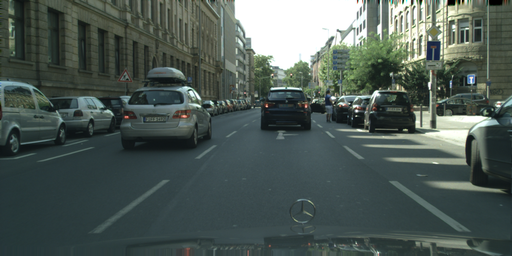} &
	\includegraphics[width=\linewidth]{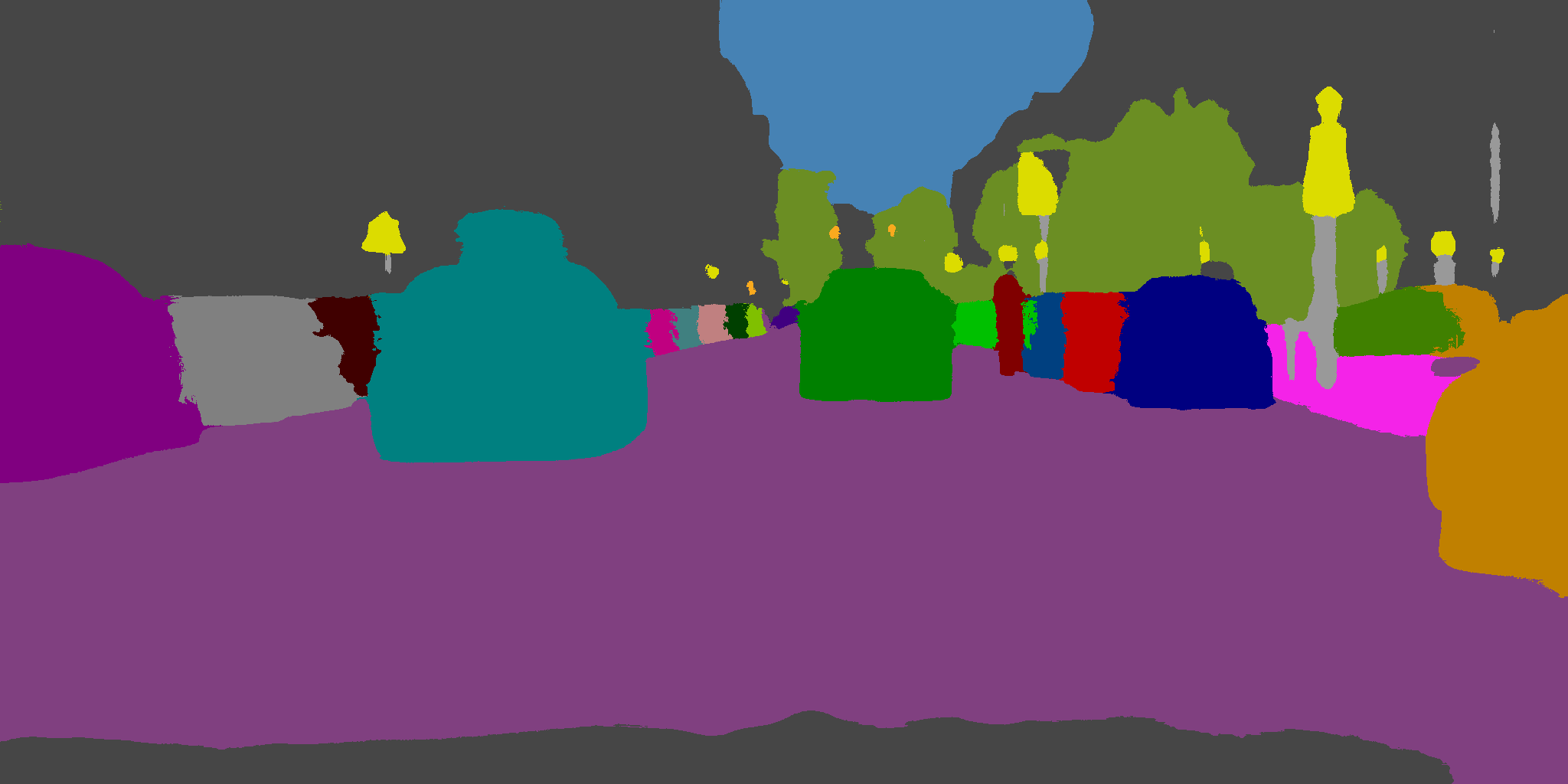} &
	\includegraphics[width=\linewidth]{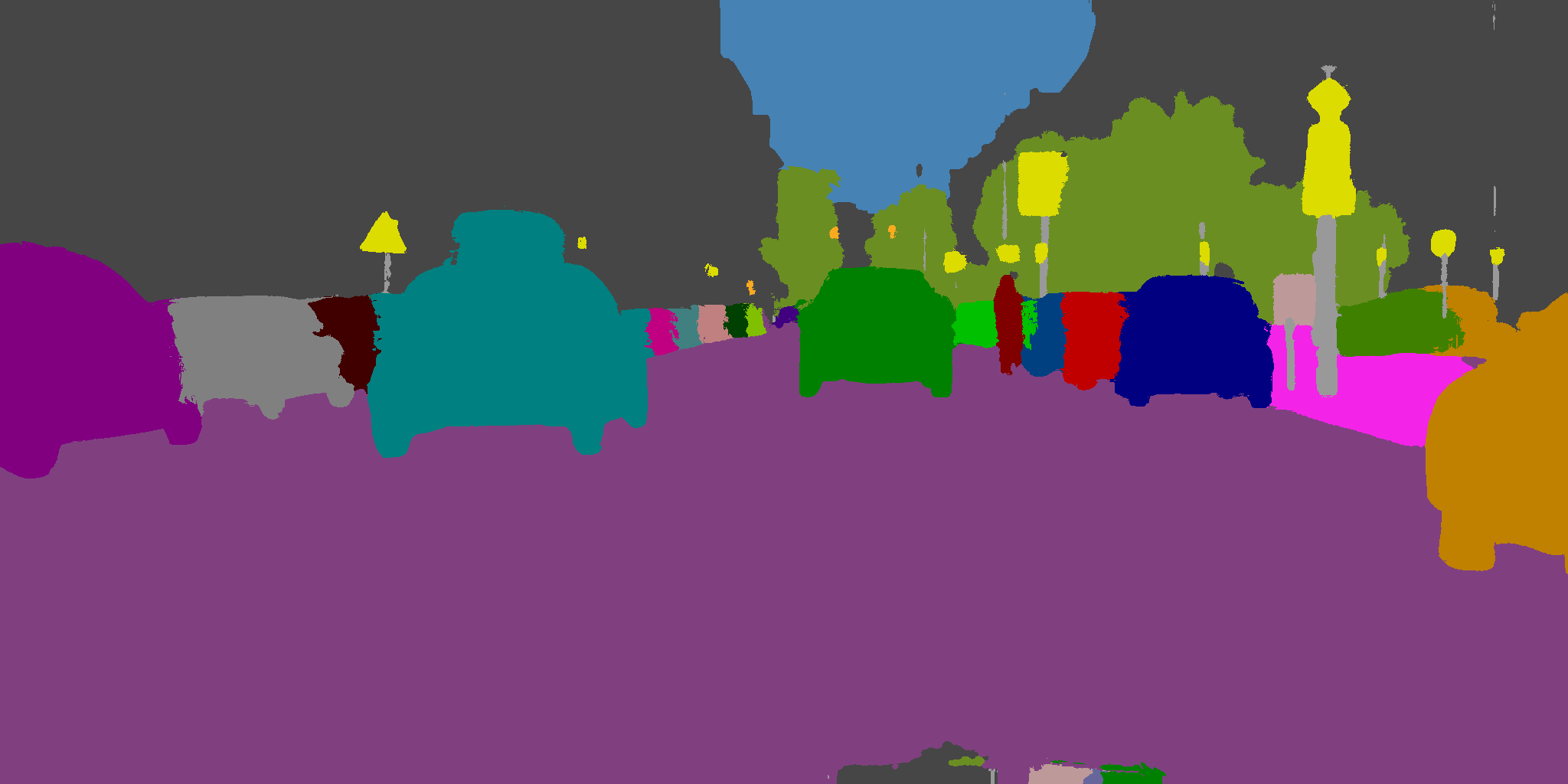} \\

	\global \def \im{frankfurt_000001_064651}
	\includegraphics[width=\linewidth]{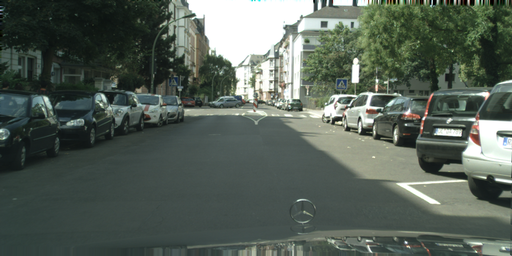} &
	\includegraphics[width=\linewidth]{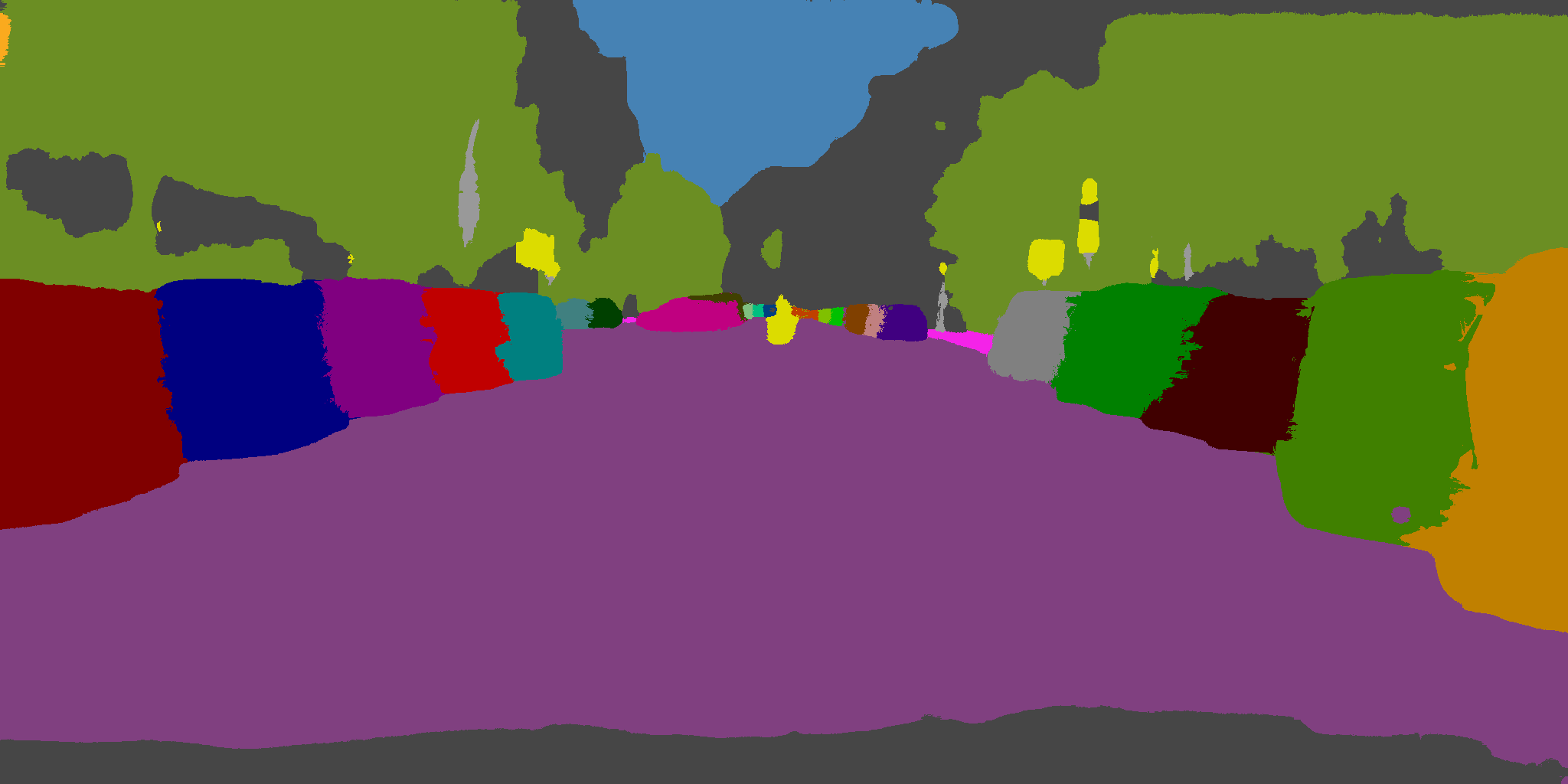} &
	\includegraphics[width=\linewidth]{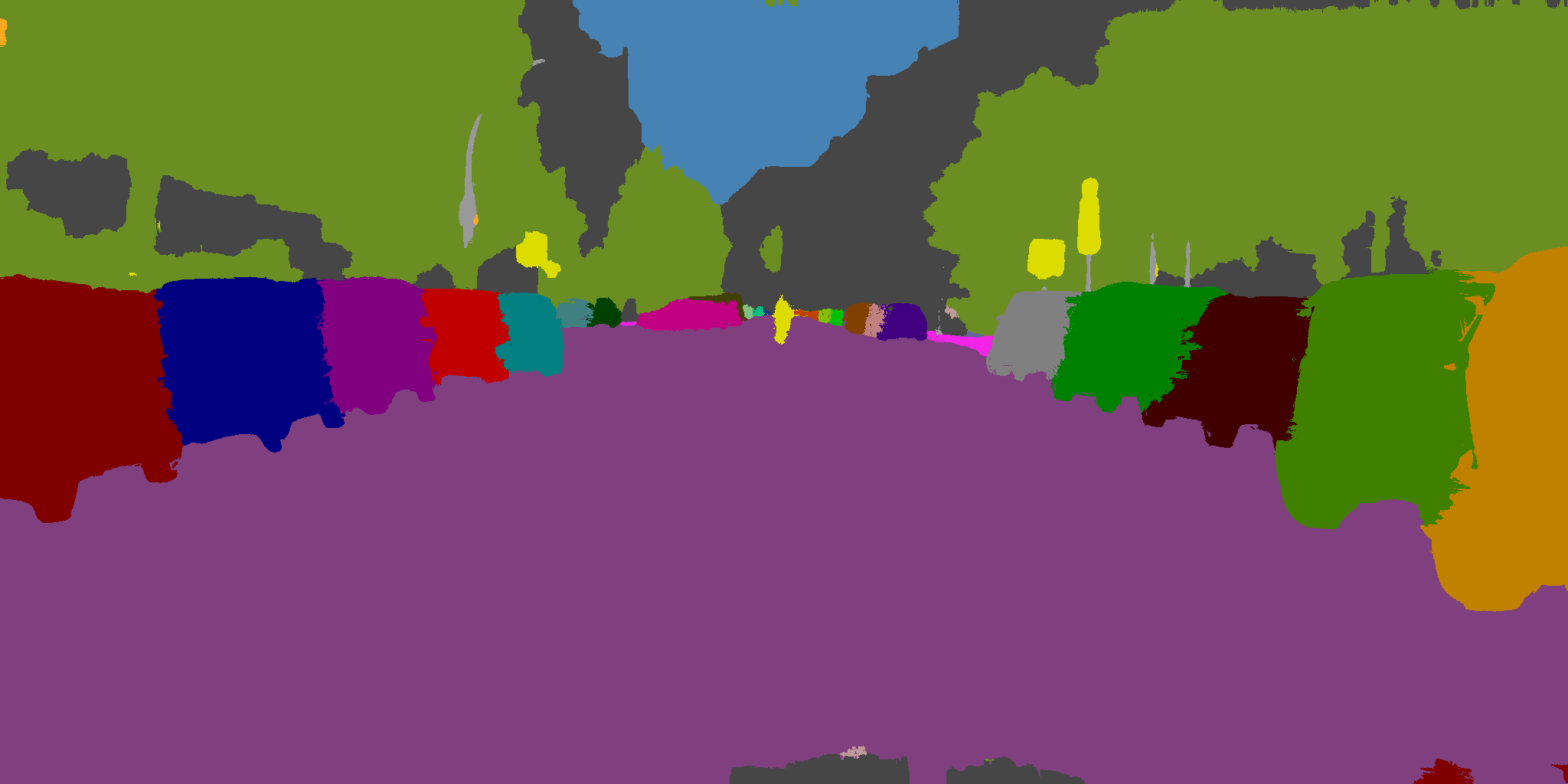} \\

	\global \def \im{frankfurt_000000_012868}
	\includegraphics[width=\linewidth]{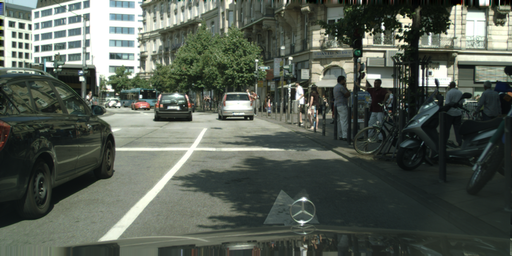} &
	\includegraphics[width=\linewidth]{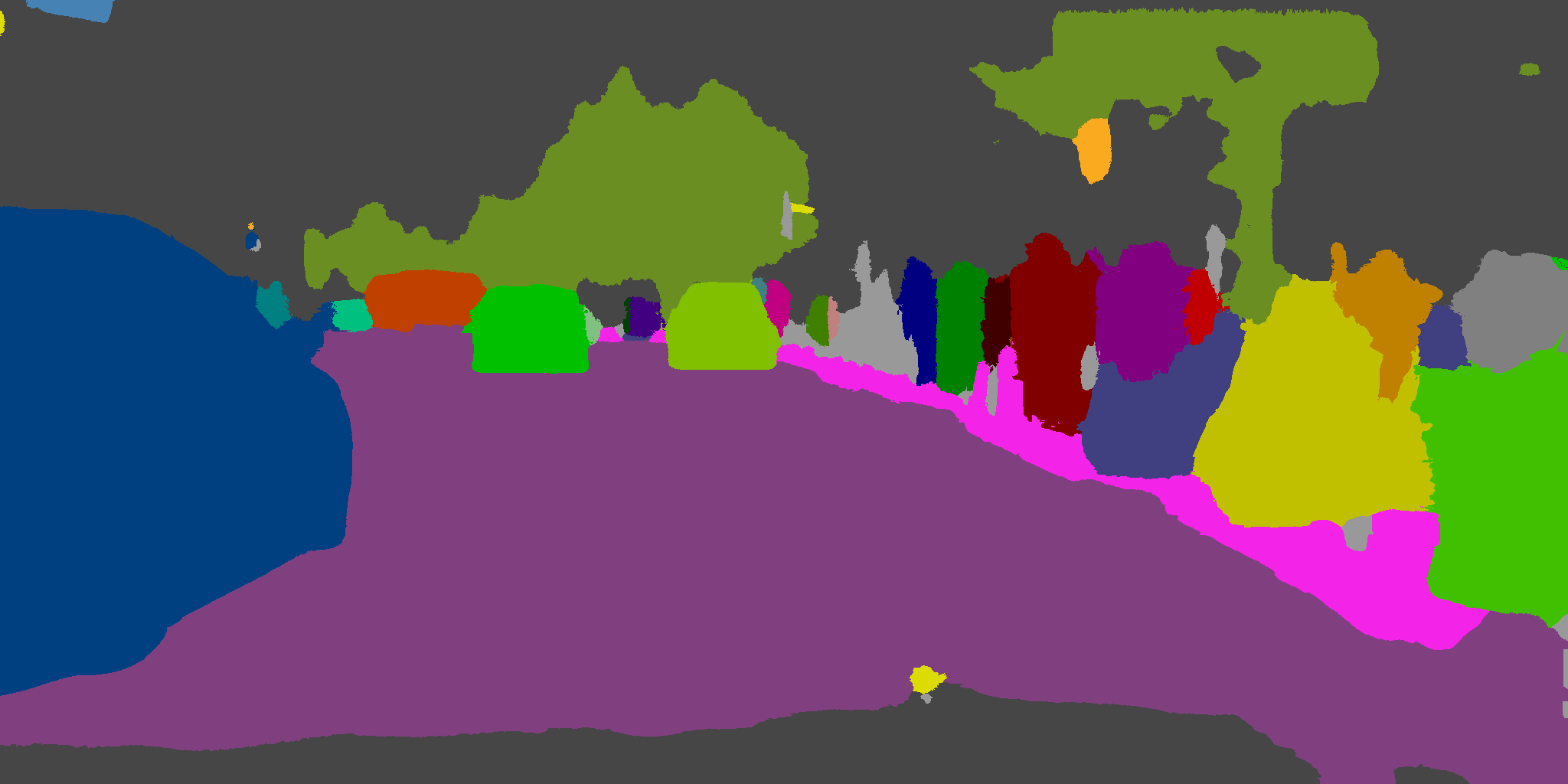} &
	\includegraphics[width=\linewidth]{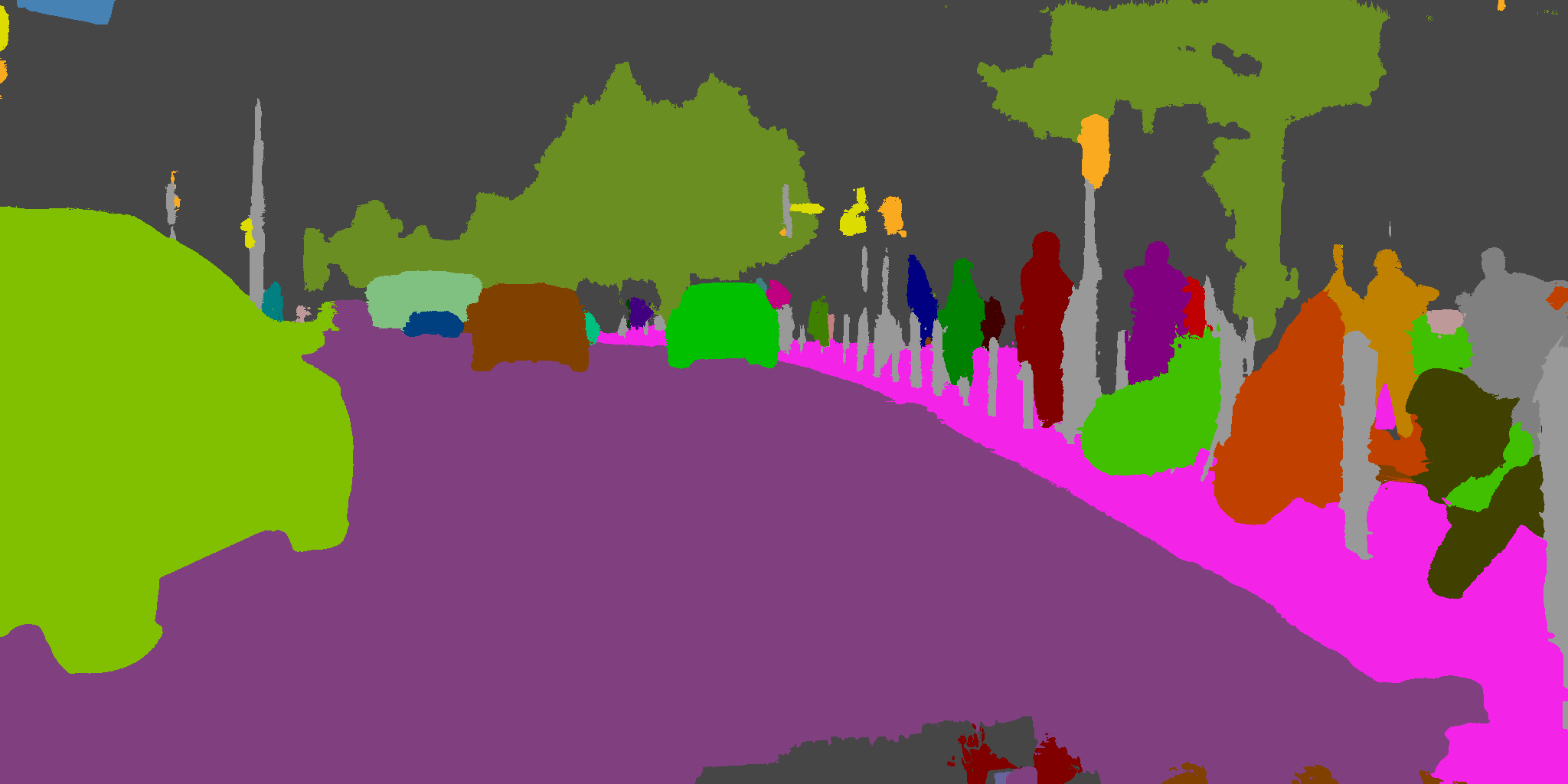} \\

	\global \def \im{lindau_000026_000019}
	\includegraphics[width=\linewidth]{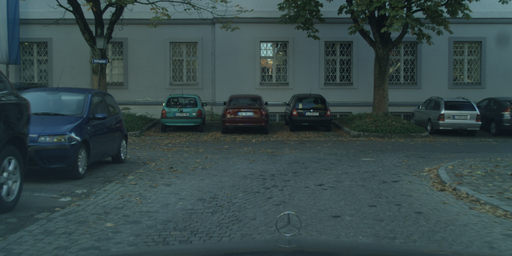} &
	\includegraphics[width=\linewidth]{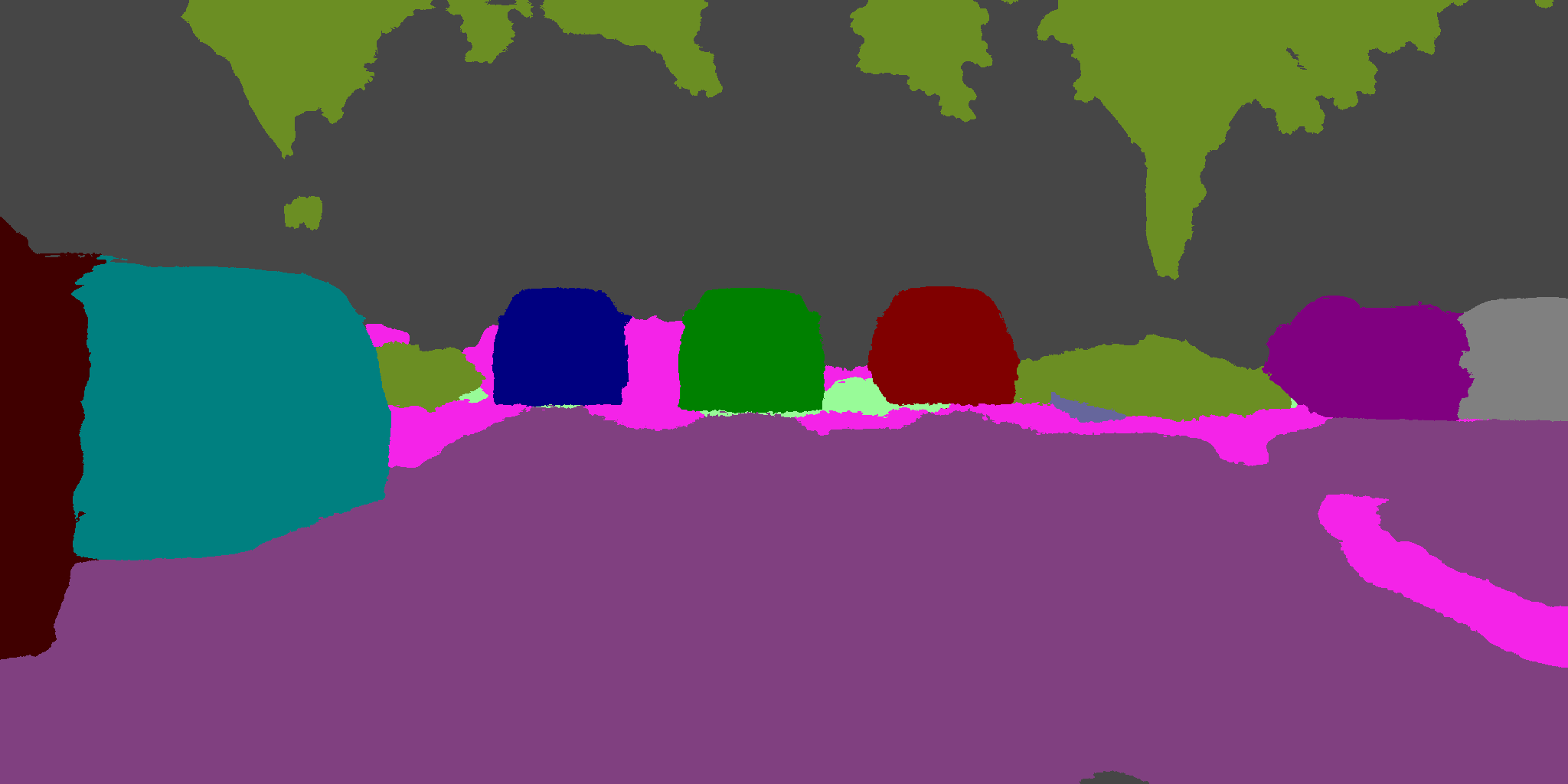} &
	\includegraphics[width=\linewidth]{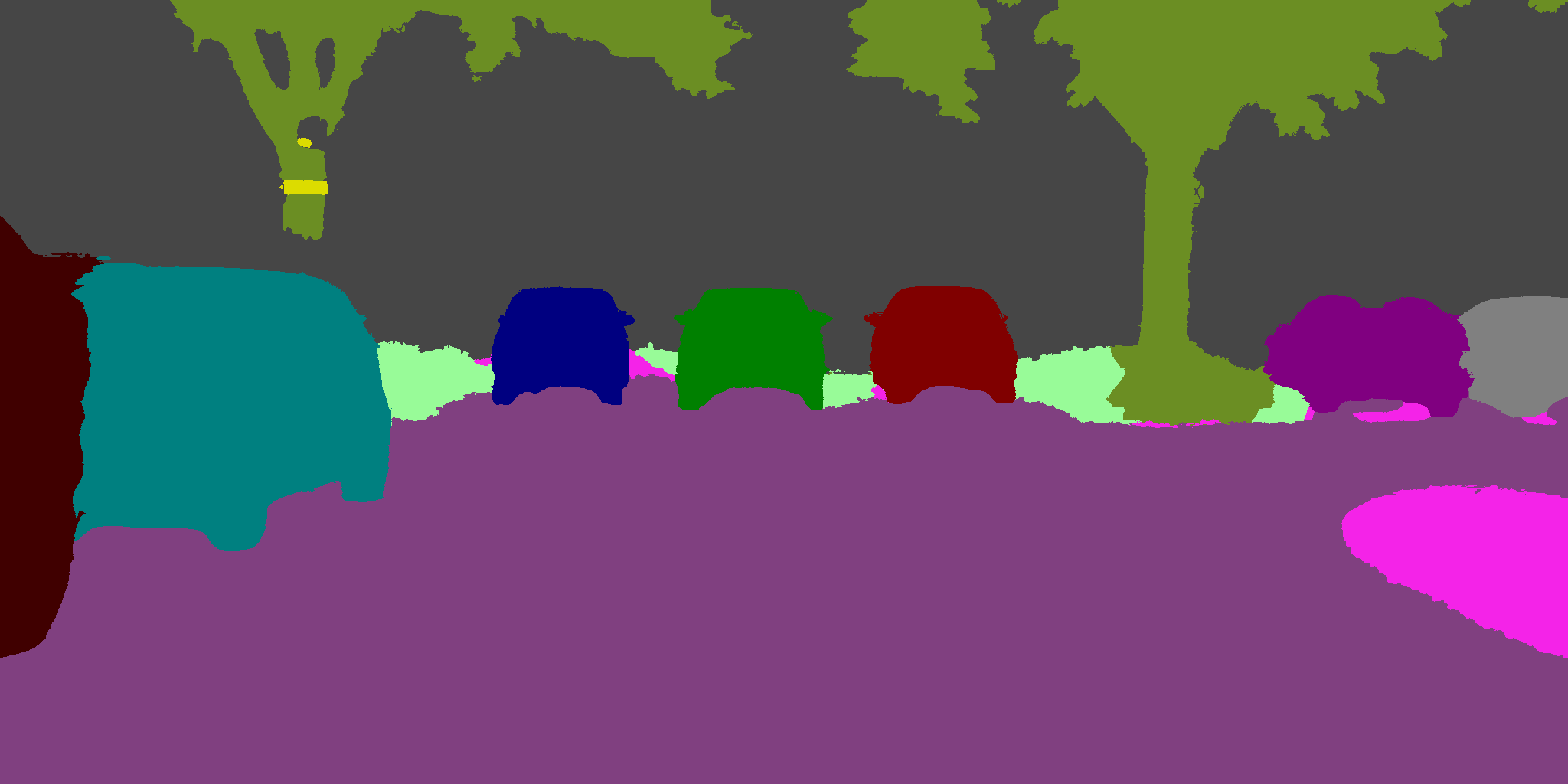} \\

	\end{tabularx}
	\caption{Comparison of our weakly- and fully-supervised instance segmentation models on the Cityscapes dataset. The fully-supervised model produces more precise segmentations, as seen by its sharper boundaries. The last row also shows how the fully-supervised model segments ``stuff'' classes such as ``vegetation'' and ``sidewalk'' more accurately.
	Both of these were expected, as the weakly-supervised model is trained only with bounding box and image tag annotations.
	Rows 3 and 6 also show some instances with different colouring.
	Each colour represents an instance ID, and a discrepancy between the two indicates that a different number of instances were segmented.}
	\label{fig:cityscapes_qualitative}
\end{figure}

Figure~\ref{fig:cityscapes_qualitative} and Tab.~\ref{tab:cityscapes_per_class} present additional qualitative and quantitative results on the Cityscapes dataset.
Similarly, Fig.~\ref{fig:voc_qualitative} and Tab.~\ref{tab:voc_per_class} show additional results on the Pascal VOC dataset.

\begin{figure}[t]
	
	\begin{tabularx}{\linewidth}{ Y Y Y}
		Input image & Weakly-supervised model & Fully-supervised model \\		
		
		\global \def \im{frankfurt_000001_031416}
		\includegraphics[width=\linewidth]{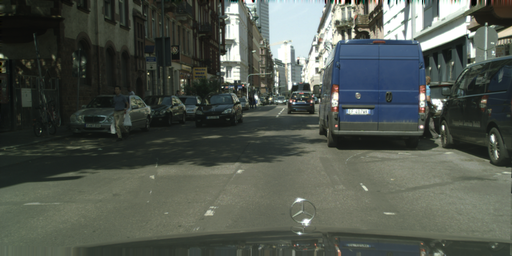} &
		\includegraphics[width=\linewidth]{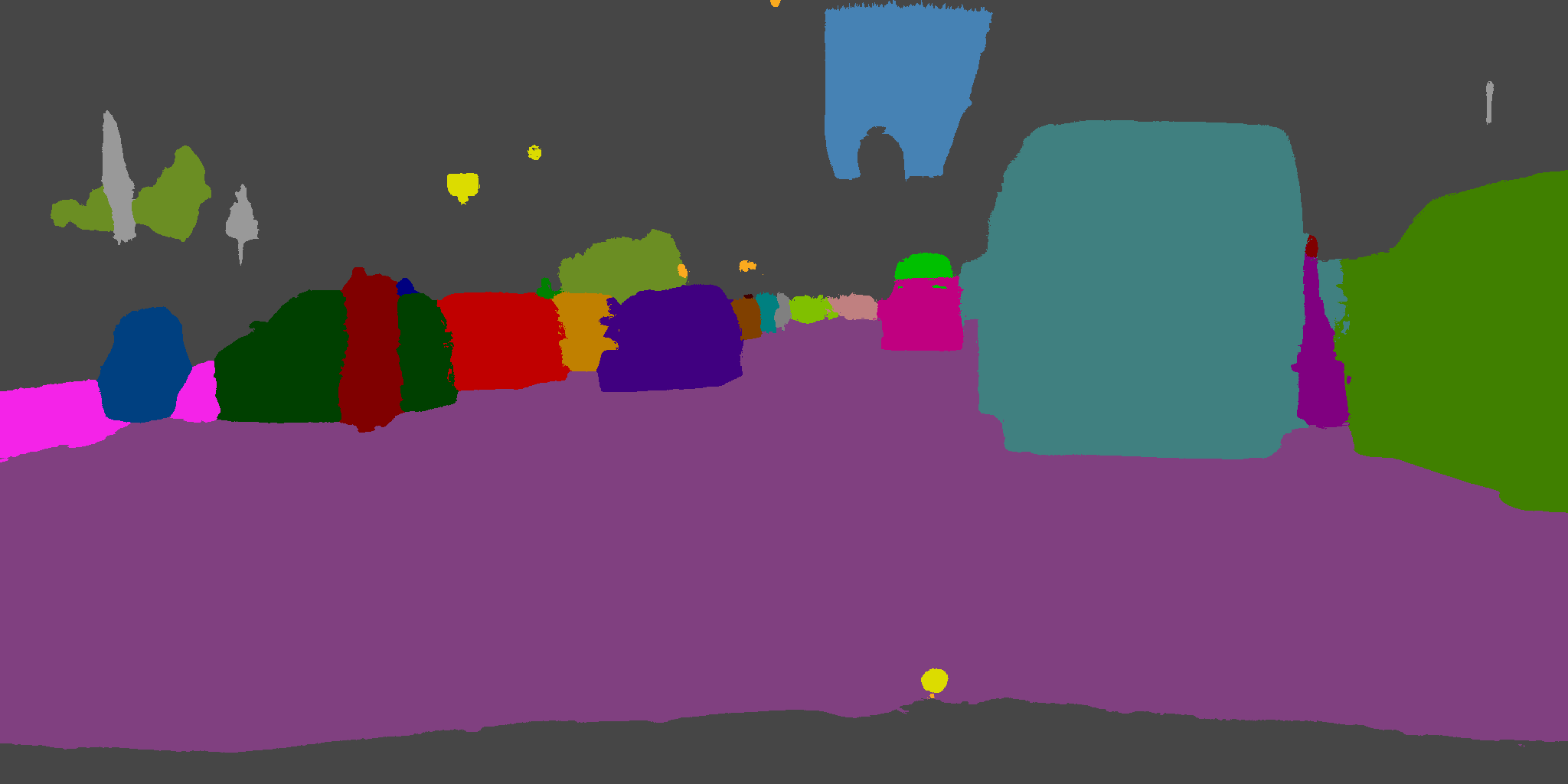} &
		\includegraphics[width=\linewidth]{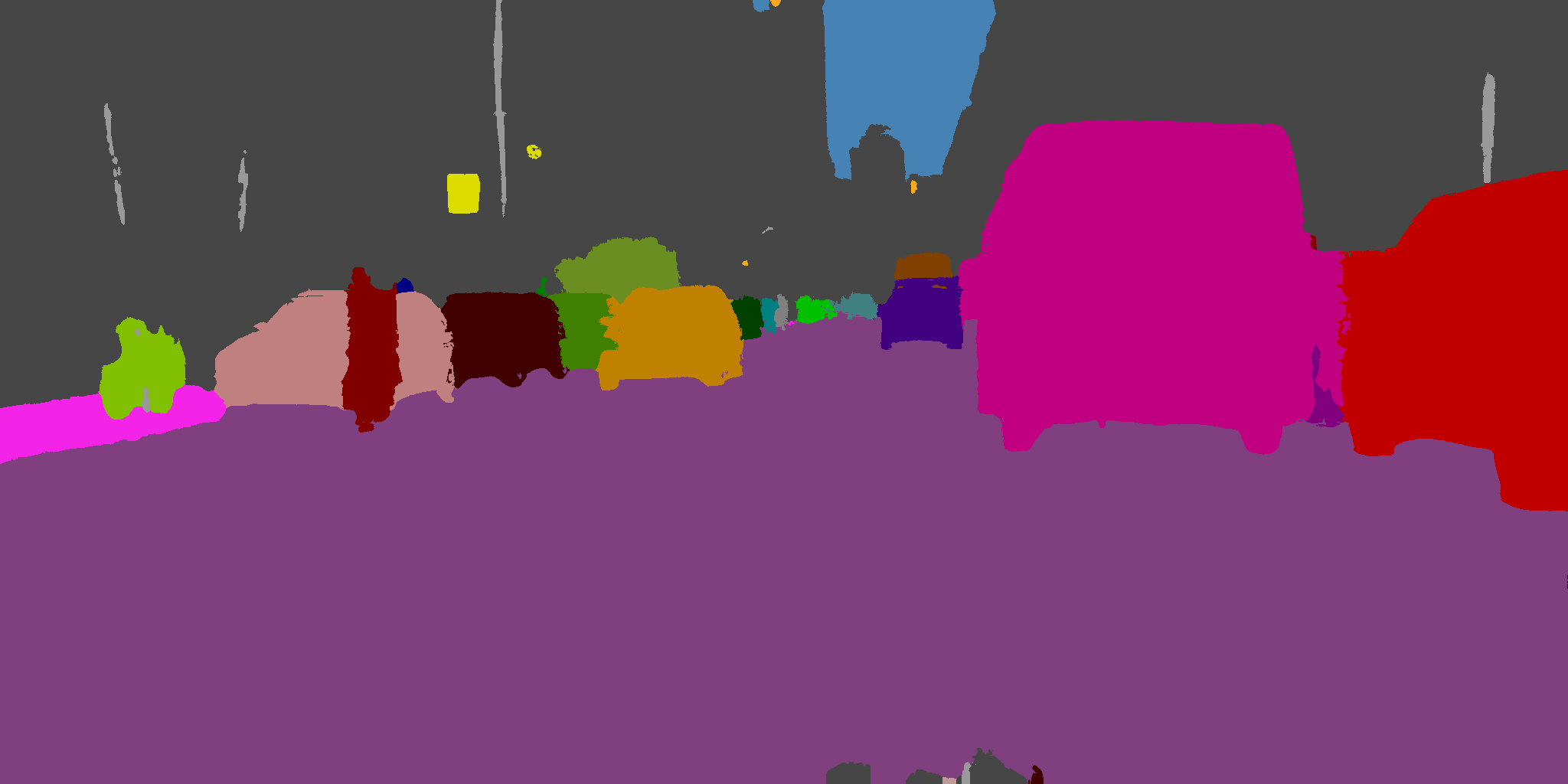} \\
		
		\global \def \im{frankfurt_000001_061763}
		\includegraphics[width=\linewidth]{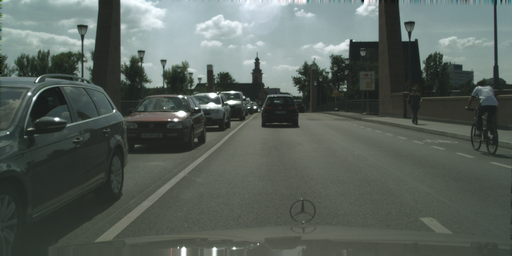} &
		\includegraphics[width=\linewidth]{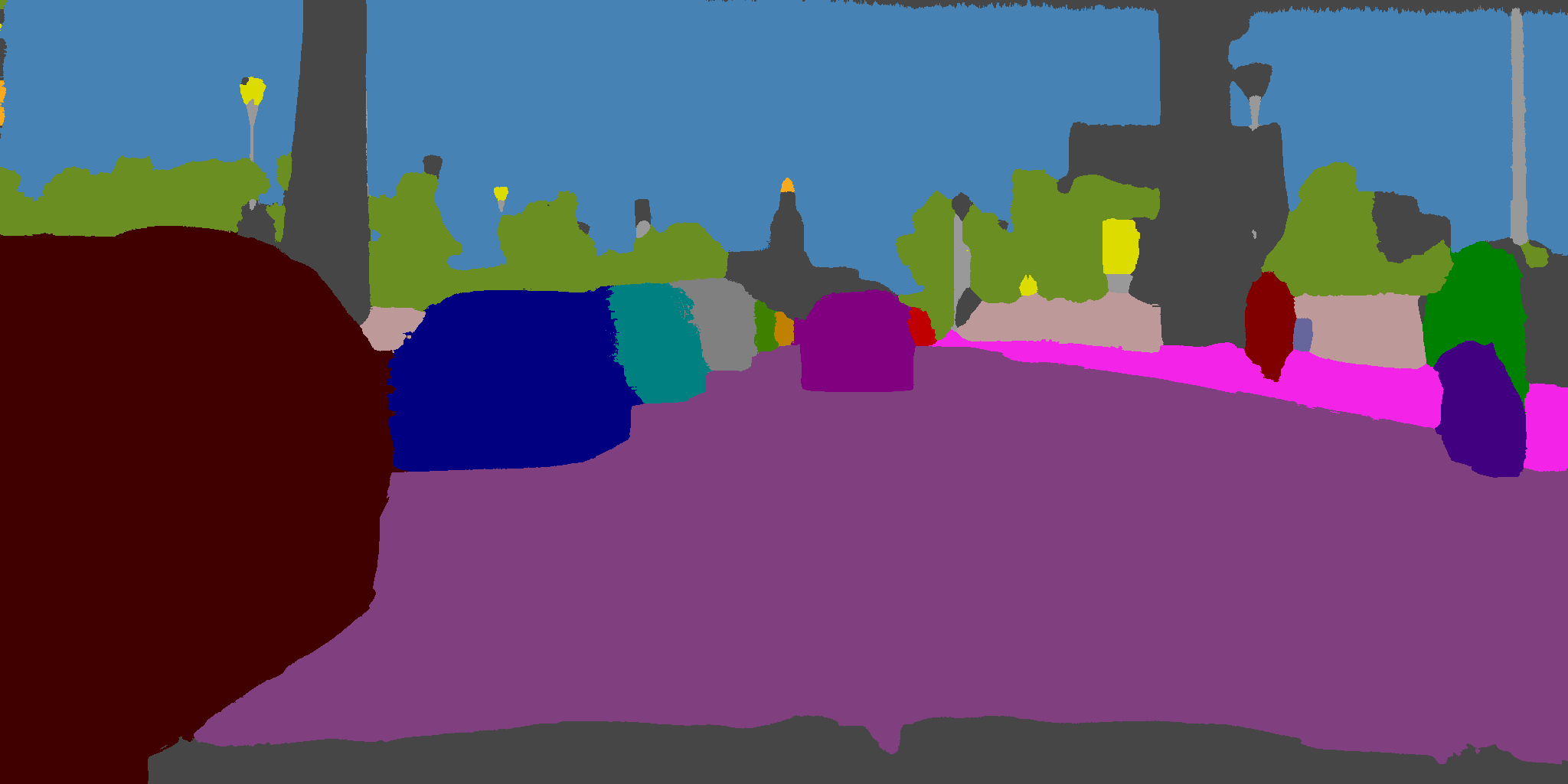} &
		\includegraphics[width=\linewidth]{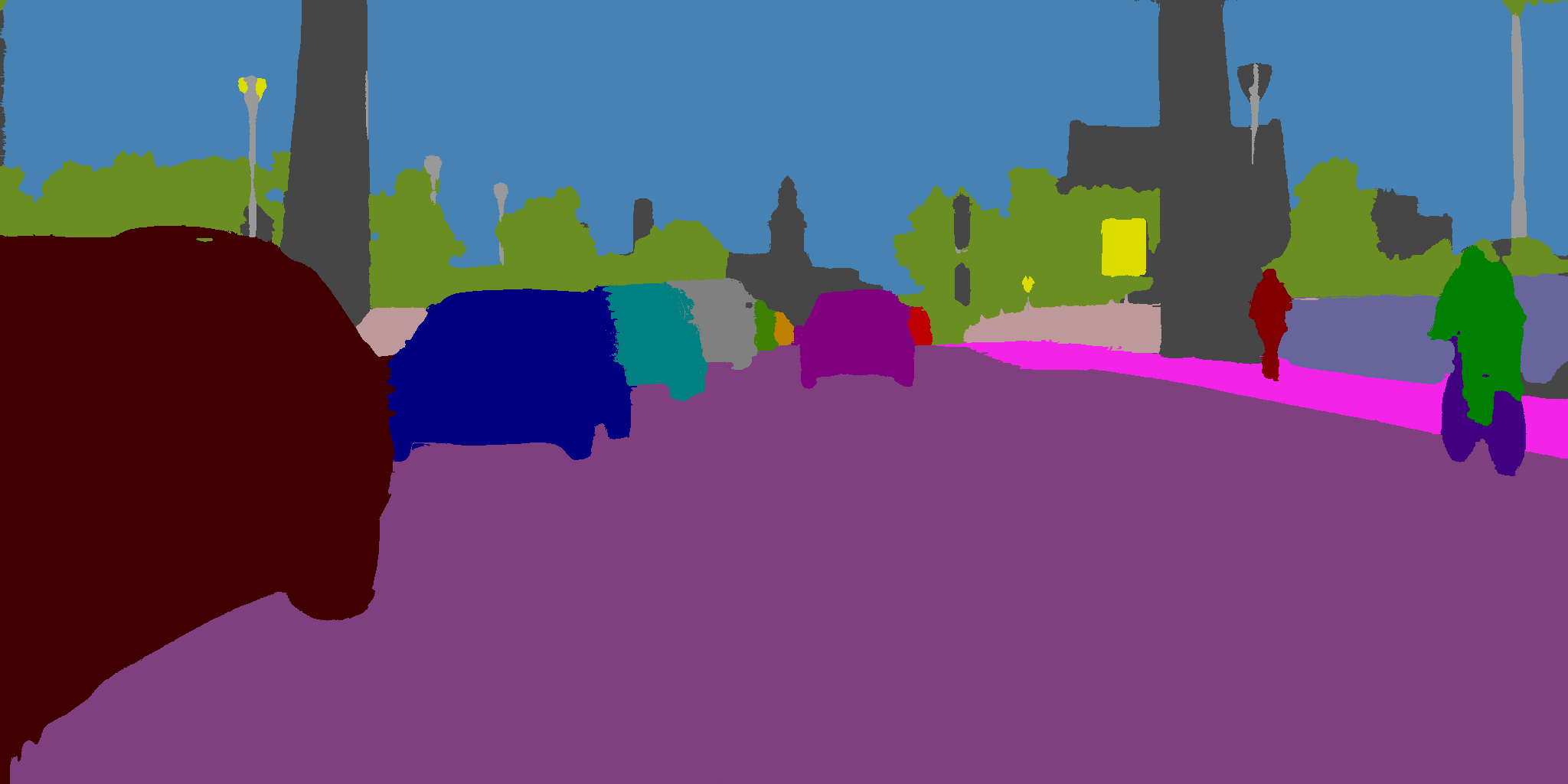} \\
		
		\global \def \im{munster_000130_000019}
		\includegraphics[width=\linewidth]{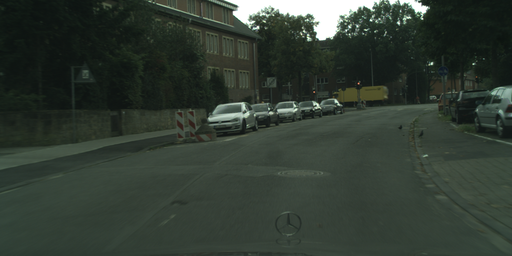} &
		\includegraphics[width=\linewidth]{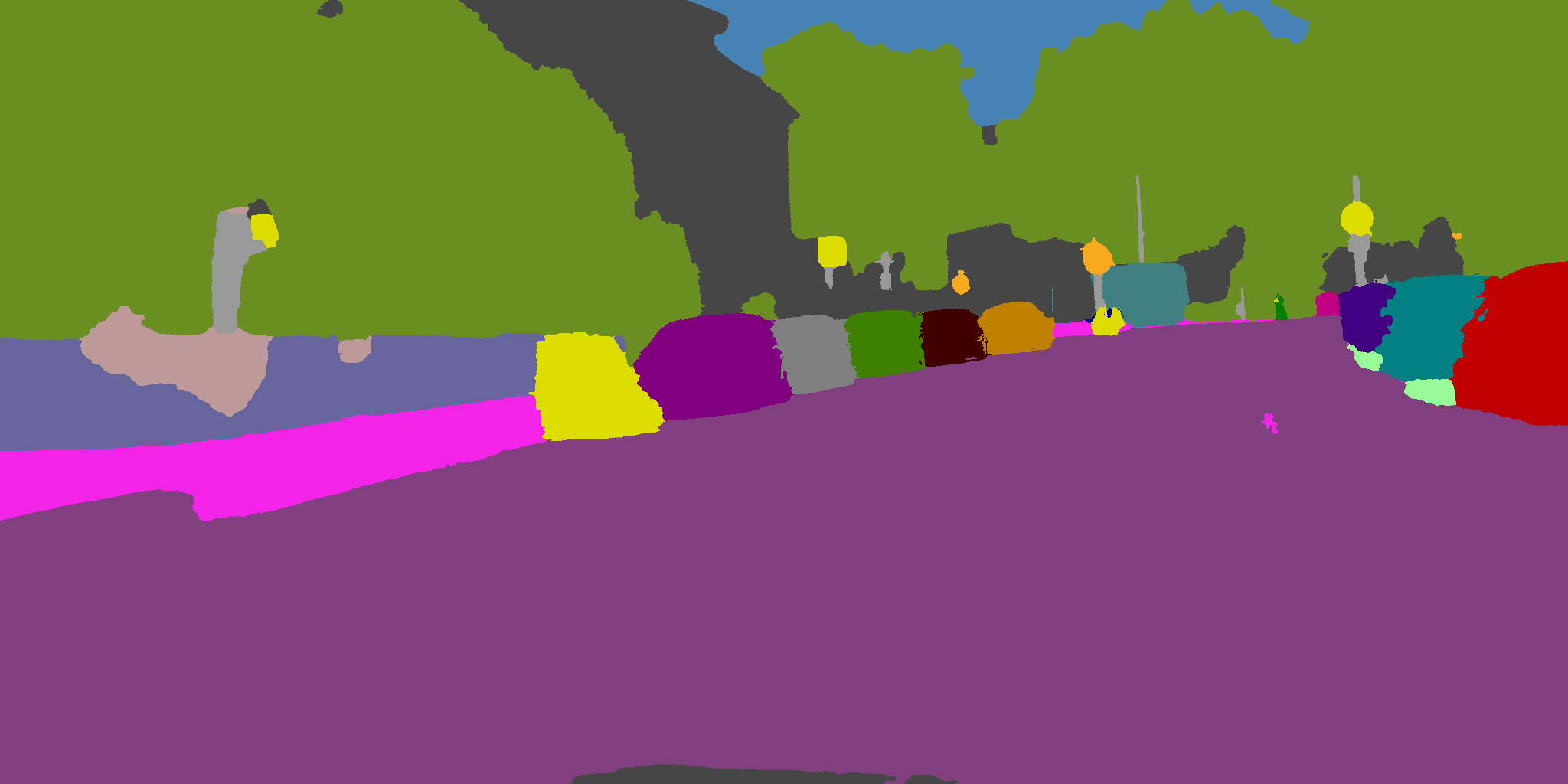} &
		\includegraphics[width=\linewidth]{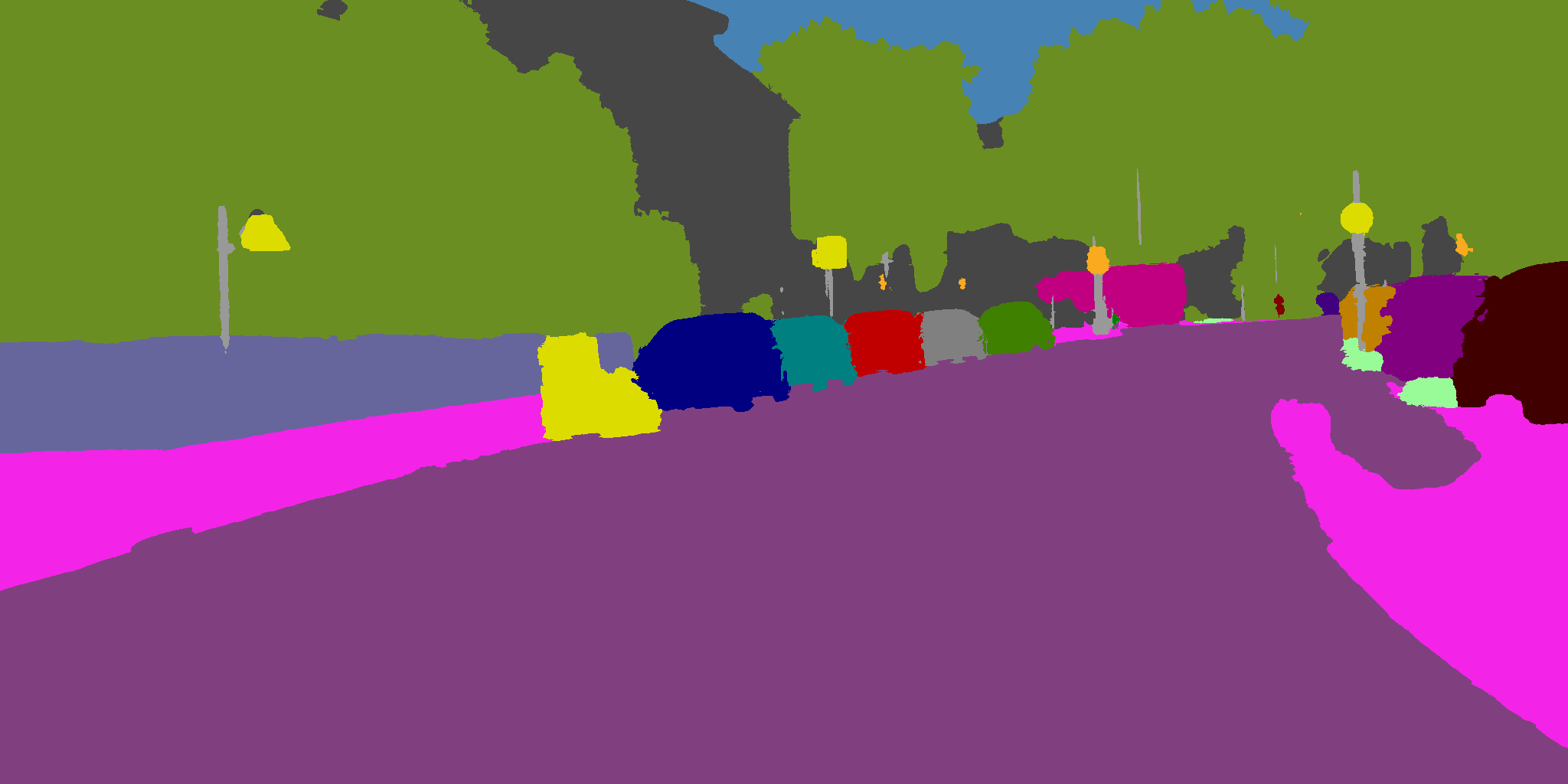} \\
		
		\global \def \im{munster_000055_000019}
		\includegraphics[width=\linewidth]{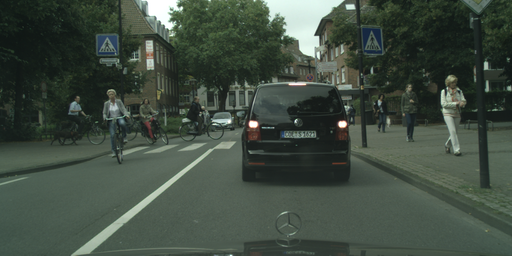} &
		\includegraphics[width=\linewidth]{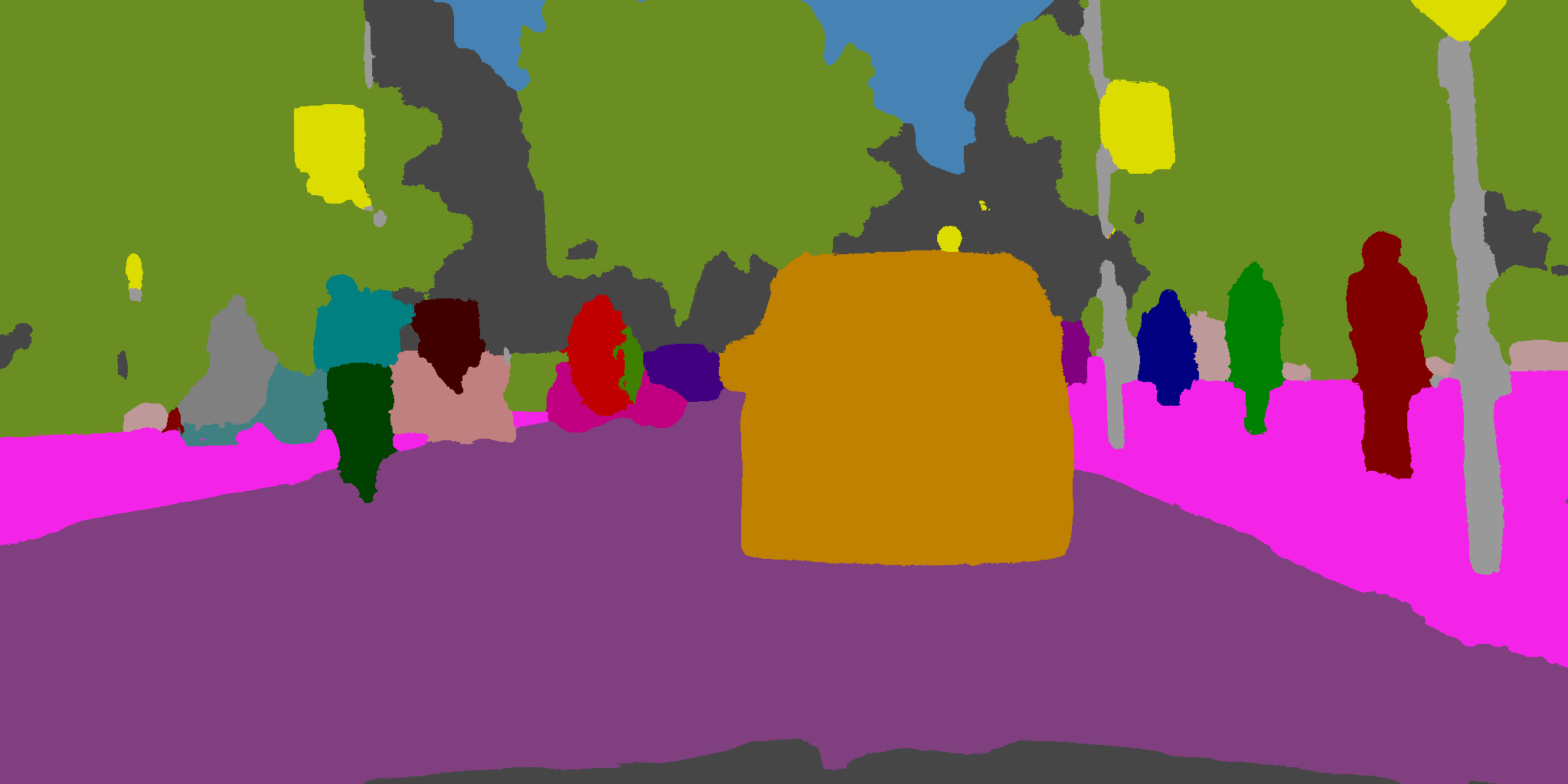} &
		\includegraphics[width=\linewidth]{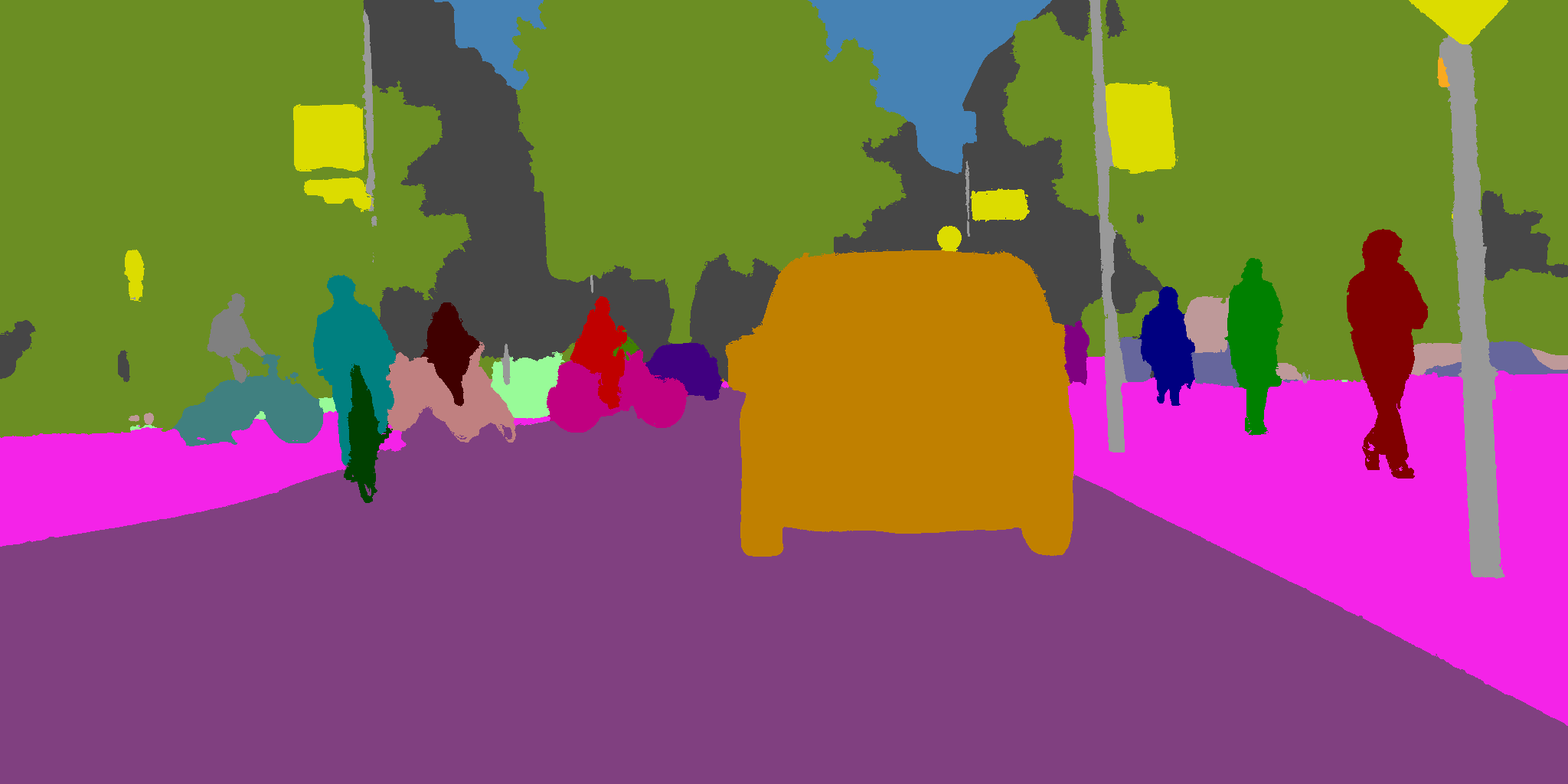} \\
		
		\global \def \im{munster_000016_000019}
		\includegraphics[width=\linewidth]{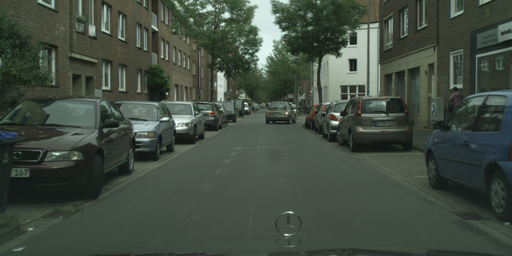} &
		\includegraphics[width=\linewidth]{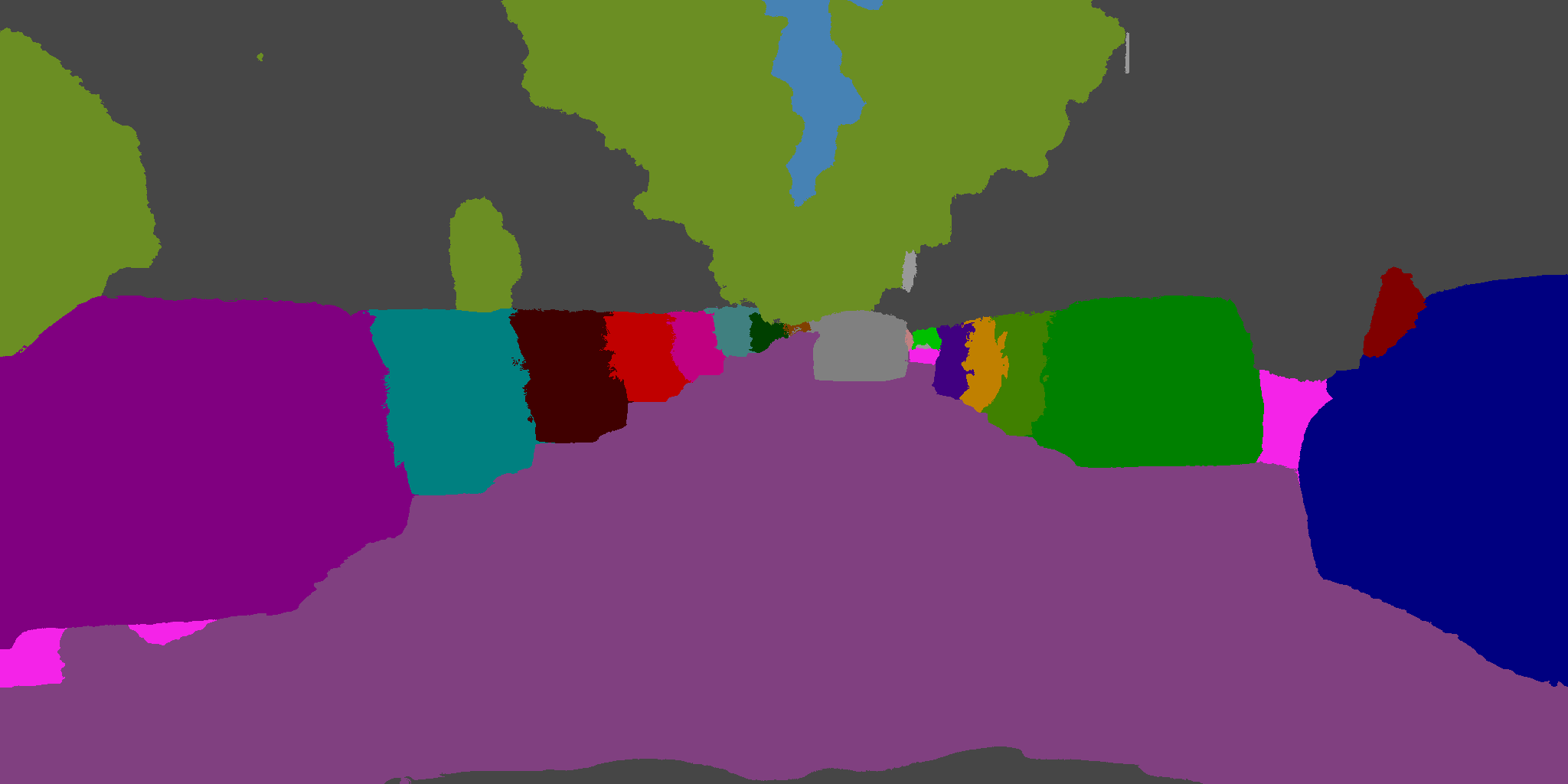} &
		\includegraphics[width=\linewidth]{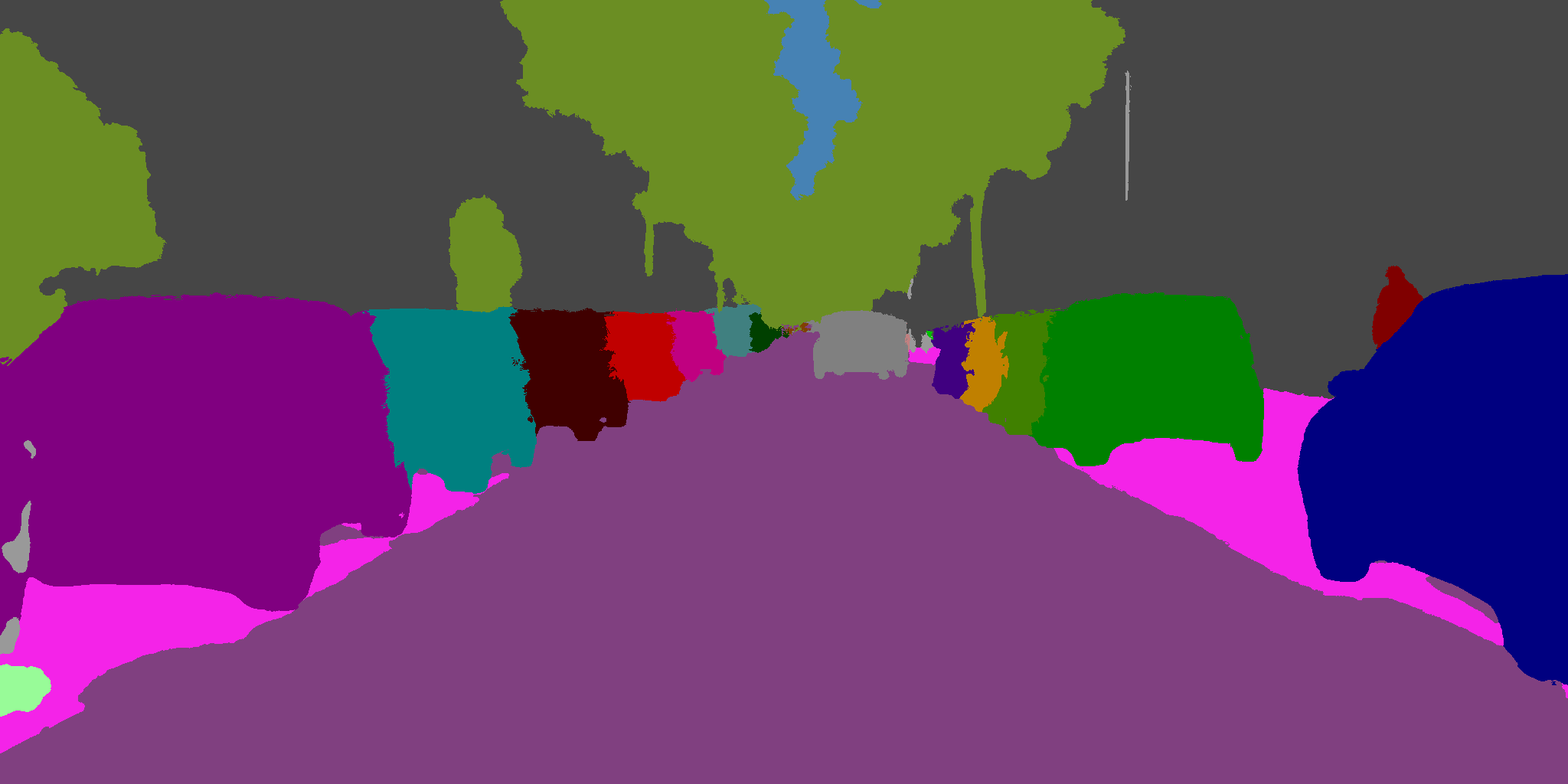} \\
		
		\global \def \im{frankfurt_000001_080091}
		\includegraphics[width=\linewidth]{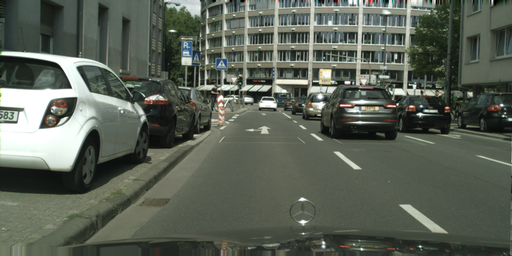} &
		\includegraphics[width=\linewidth]{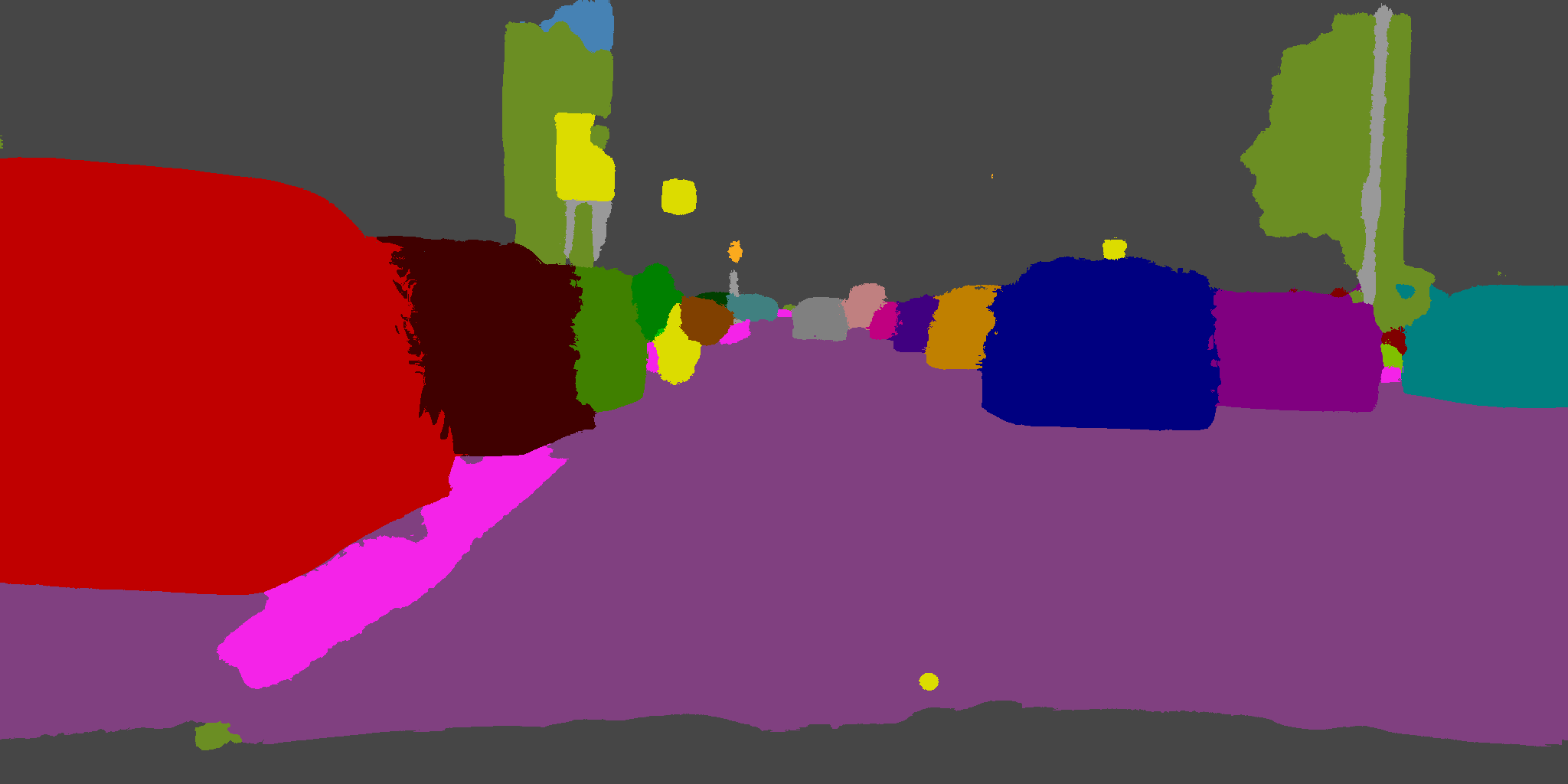} &
		\includegraphics[width=\linewidth]{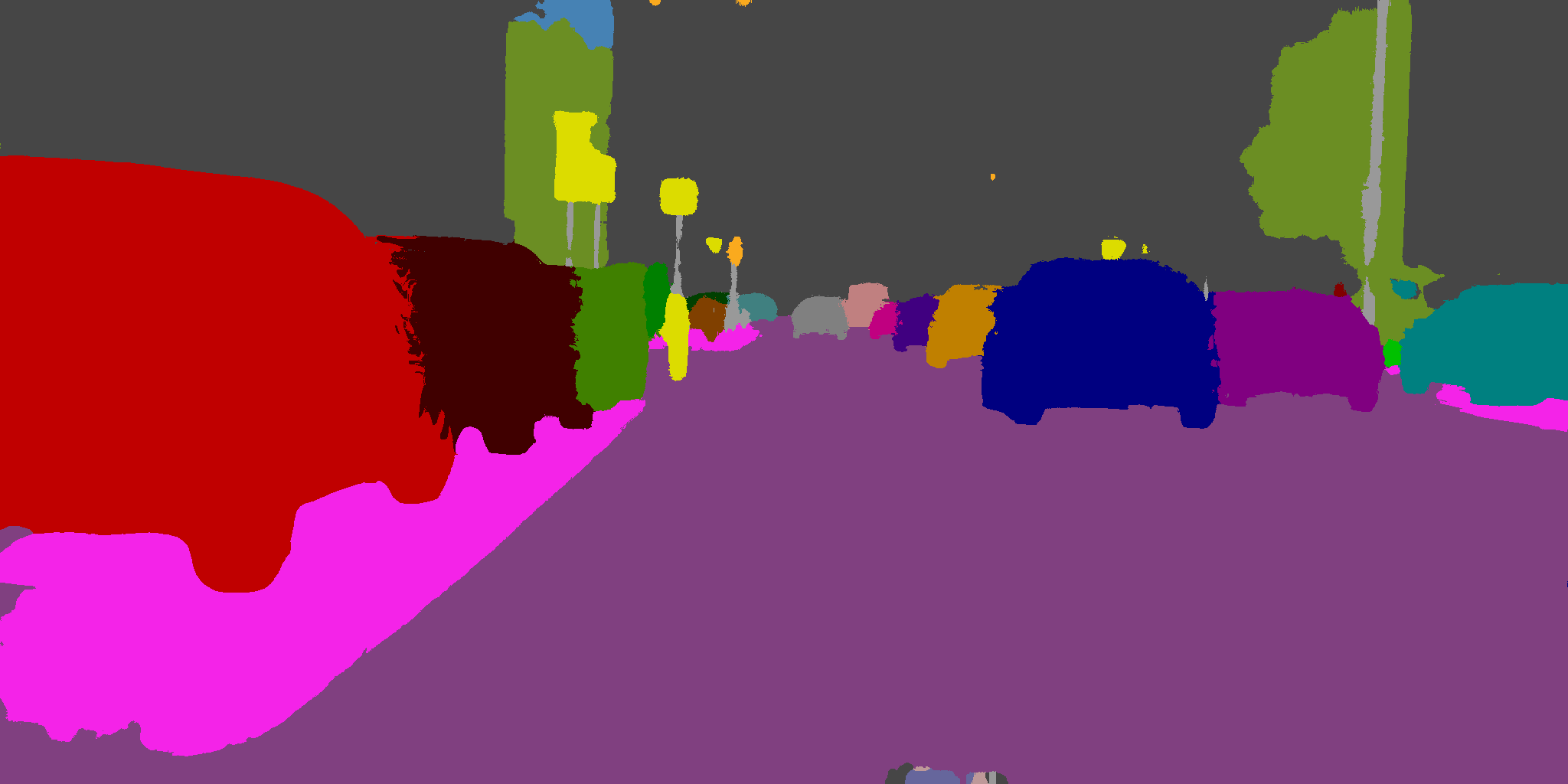} \\
		
		\global \def \im{munster_000087_000019}
		\includegraphics[width=\linewidth]{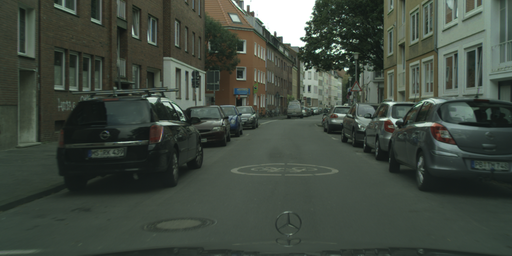} &
		\includegraphics[width=\linewidth]{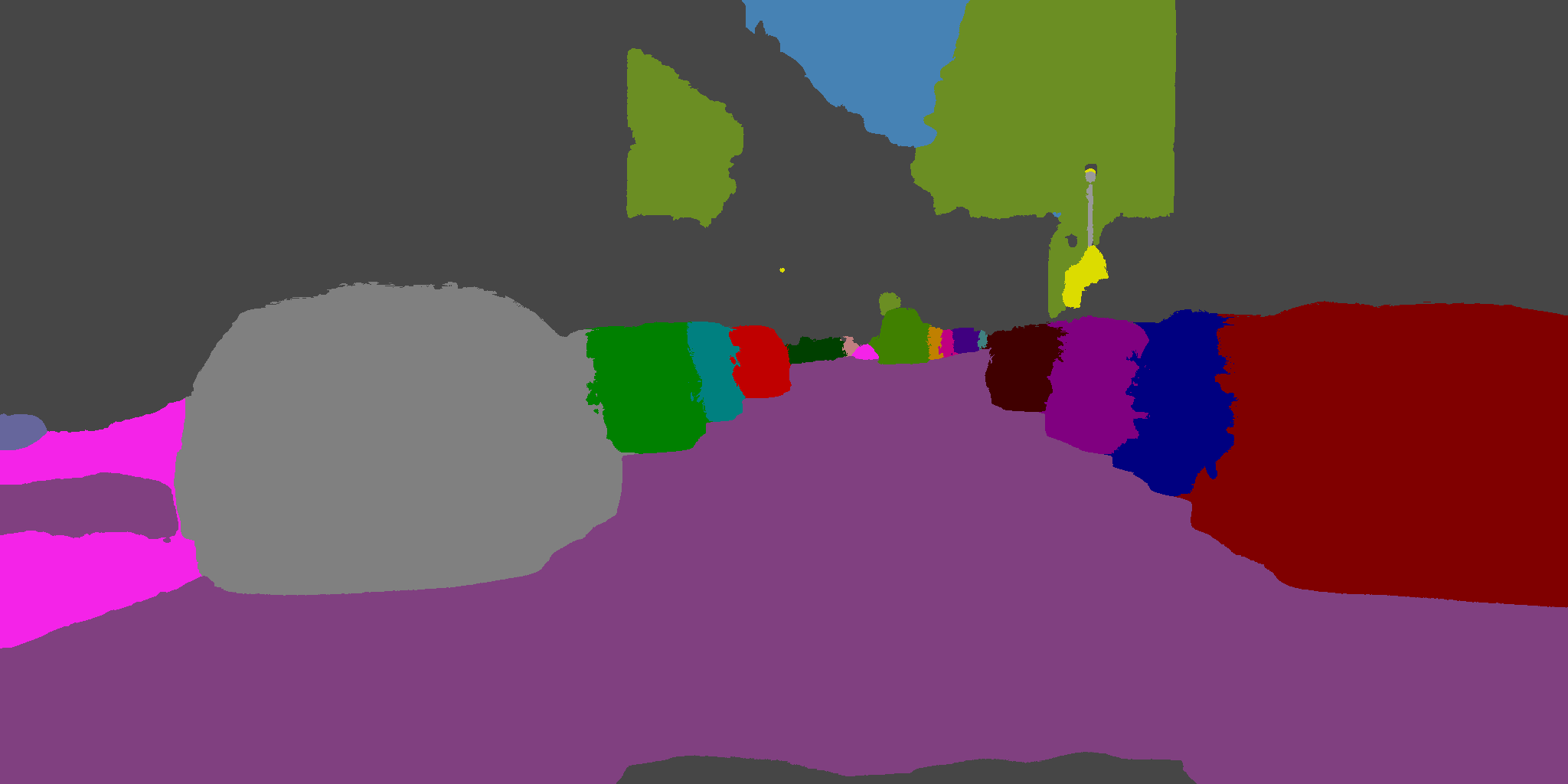} &
		\includegraphics[width=\linewidth]{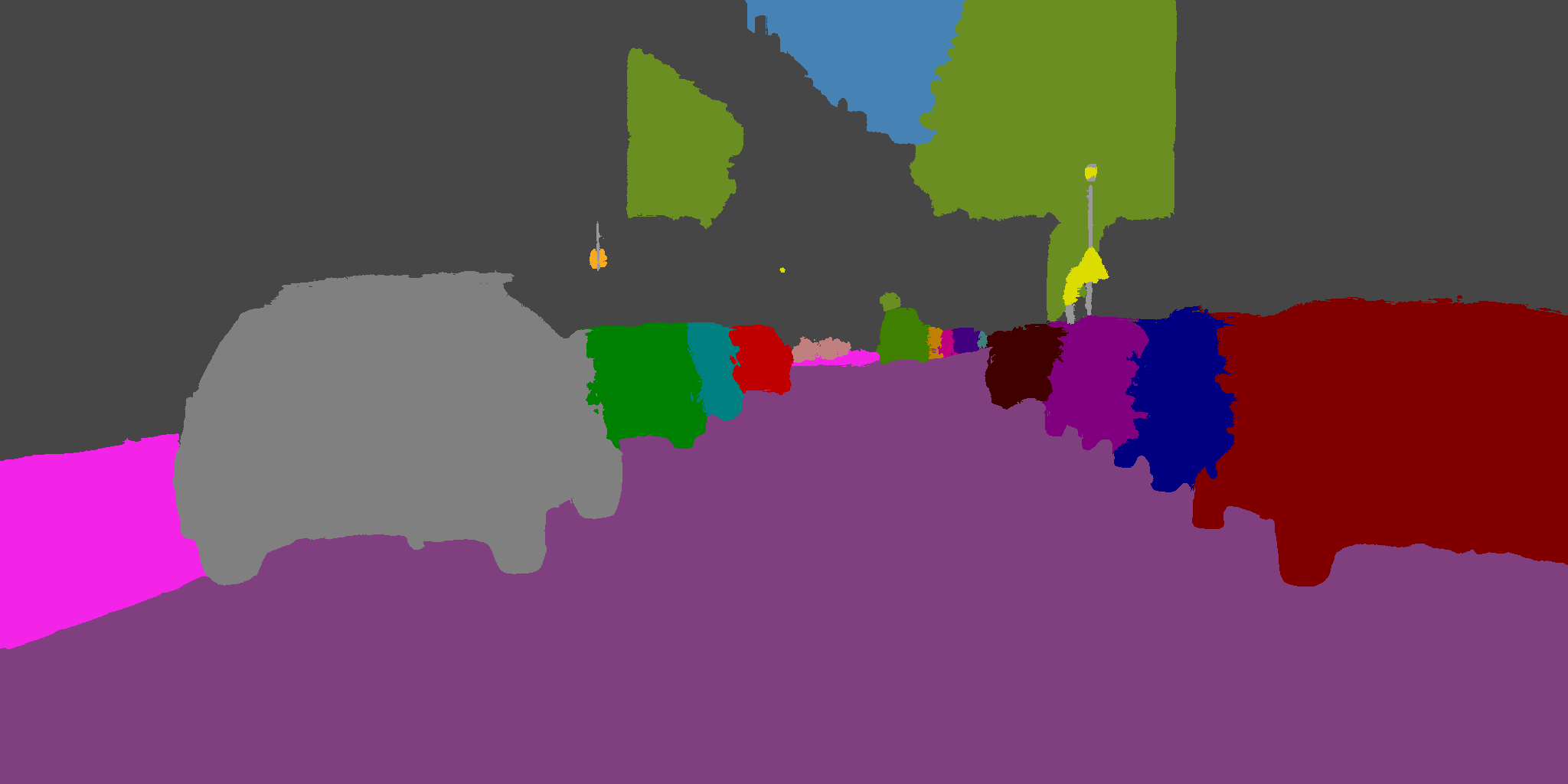} \\
		
		\textcolor{white}{This inserts some vspace} \\

	\end{tabularx}
	\\
	\textbf{Fig.~\ref{fig:cityscapes_qualitative} cont.} Comparison of our weakly- and fully-supervised instance segmentation models on the Cityscapes dataset.
	The last three rows show how the fully-supervised model is also able to segment ``stuff'' classes such as ``sidewalk'' more accurately.
	This was expected since the weakly-supervised model is only trained with image-level tags for ``stuff'' classes, which provides very little localisation information. 
	\label{fig:cityscapes_qualitative_cont}
\end{figure}
\begin{figure}[t]
	\centering

	\begin{tabularx}{\linewidth}{ Y Y Y}
	Input image & Weakly-supervised model & Fully-supervised model \\		
		
	\global \def \im{2007_001311}
	\includegraphics[width=\linewidth]{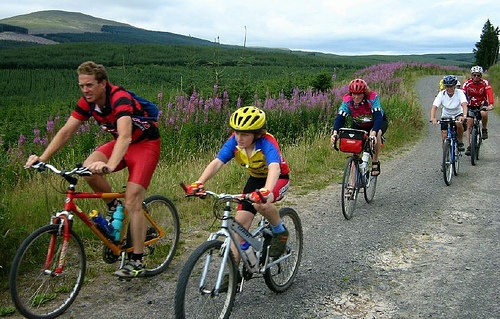} &
	\includegraphics[width=\linewidth]{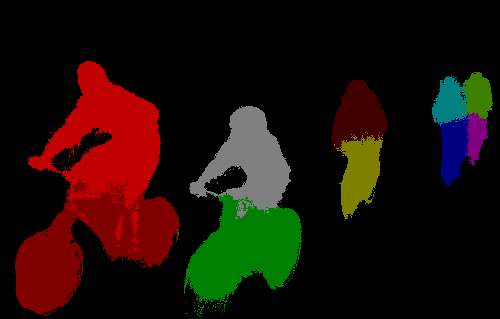} &
	\includegraphics[width=\linewidth]{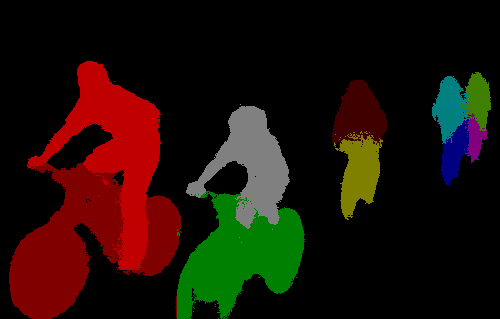} \\
		
	\global \def \im{2009_004987}
	\includegraphics[width=\linewidth]{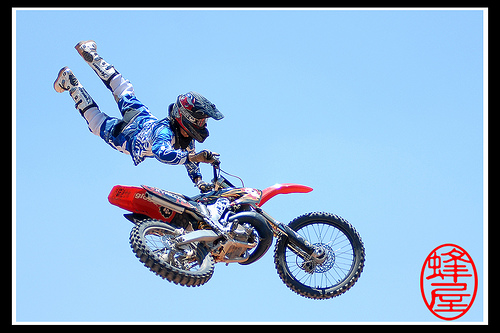} &
	\includegraphics[width=\linewidth]{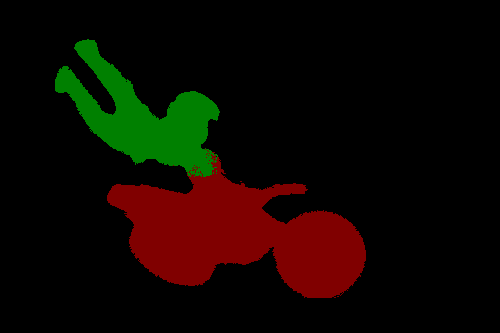} &
	\includegraphics[width=\linewidth]{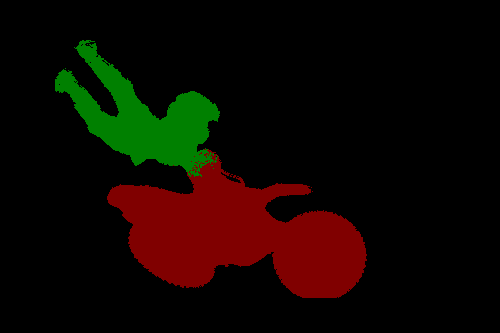} \\

	\global \def \im{2011_002322}
	\includegraphics[width=\linewidth]{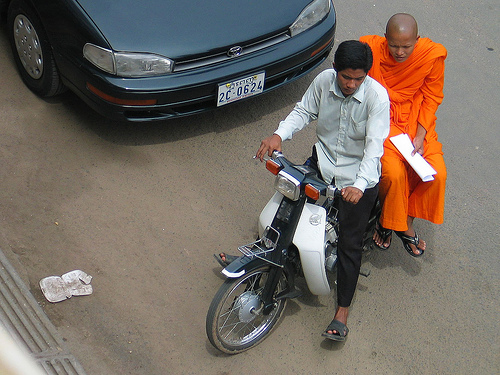} &
	\includegraphics[width=\linewidth]{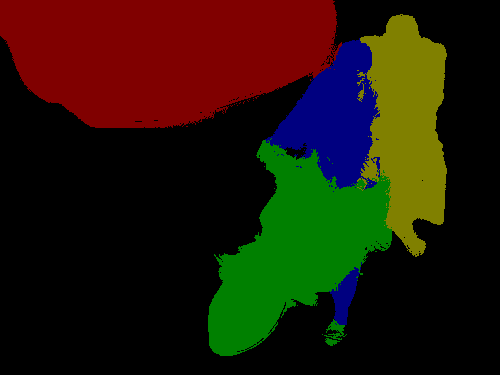} &
	\includegraphics[width=\linewidth]{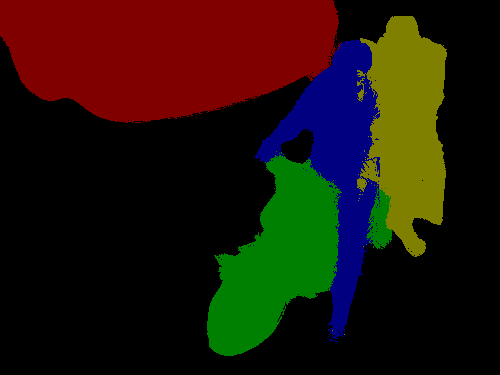} \\
	
	\global \def \im{2010_005626}
	\includegraphics[width=\linewidth]{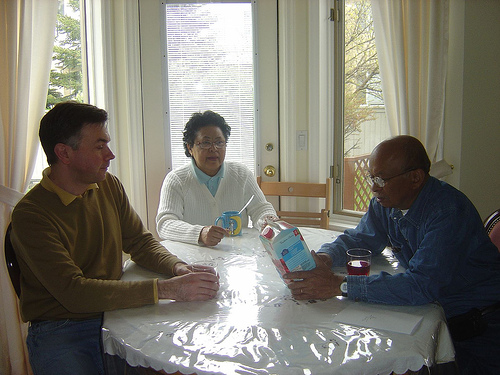} &
	\includegraphics[width=\linewidth]{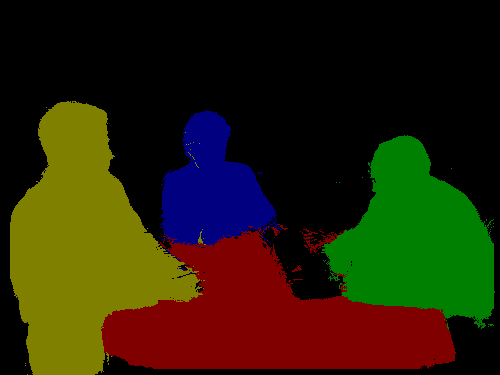} &
	\includegraphics[width=\linewidth]{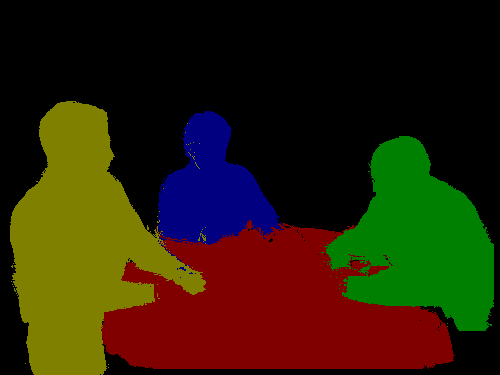} \\
	
	\global \def \im{2010_000174}
	\includegraphics[width=\linewidth]{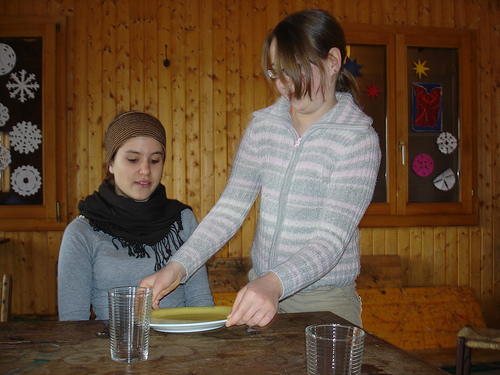} &
	\includegraphics[width=\linewidth]{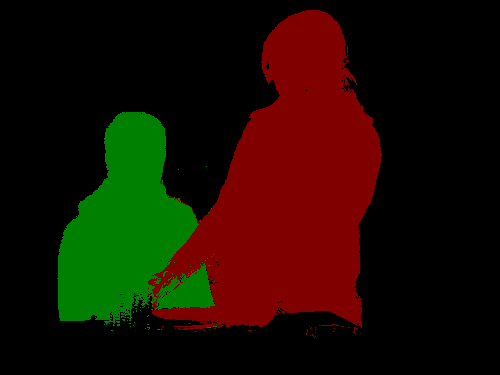} &
	\includegraphics[width=\linewidth]{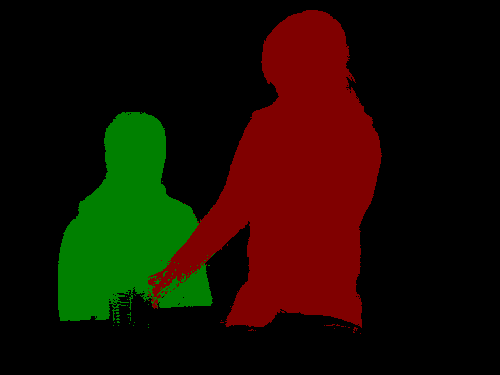} \\
	
	\end{tabularx}
	\caption{Comparison of our weakly- and fully-supervised instance segmentation models on the Pascal VOC validation set.
	The weakly-supervised model typically obtains results similar to its state-of-the-art, fully-supervised counterpart.
	However, the fully-supervised model produces more accurate and precise segmentations, as seen in the last two rows.
	}
	\label{fig:voc_qualitative}
\end{figure}

\begin{figure}[t]

	\begin{tabularx}{\linewidth}{ Y Y Y}
	Input image & Weakly-supervised model & Fully-supervised model \\		
	
	\global \def \im{2007_003020}
	\includegraphics[width=\linewidth]{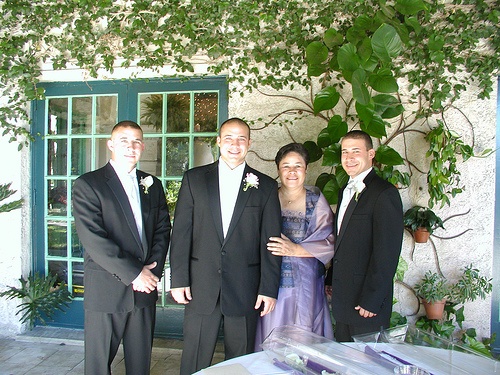} &
	\includegraphics[width=\linewidth]{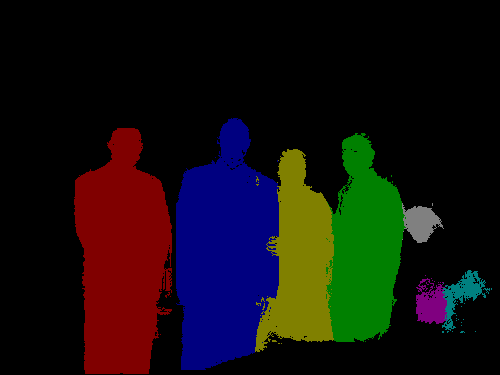} &
	\includegraphics[width=\linewidth]{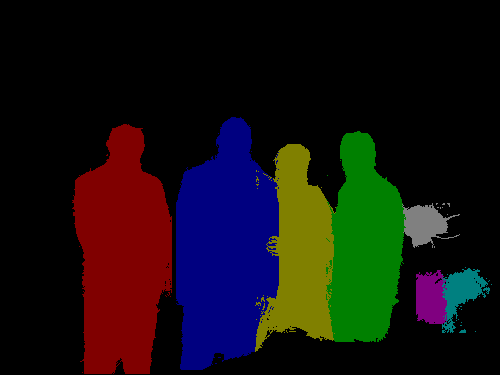} \\
				
	\global \def \im{2010_002921}
	\includegraphics[width=\linewidth]{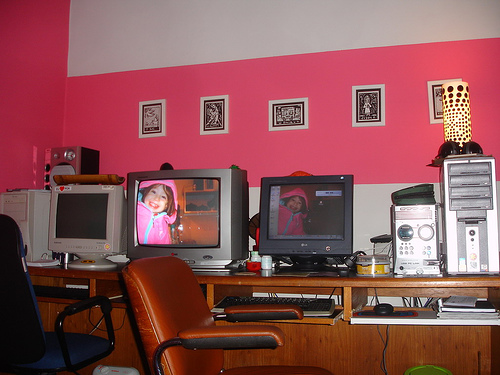} &
	\includegraphics[width=\linewidth]{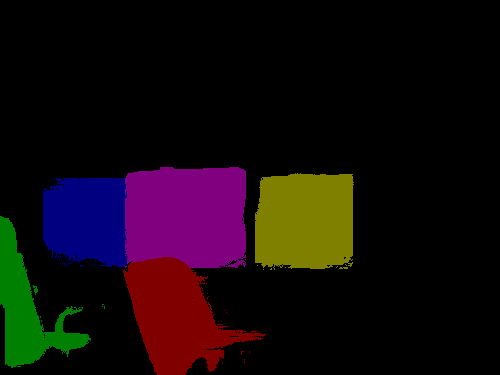} &
	\includegraphics[width=\linewidth]{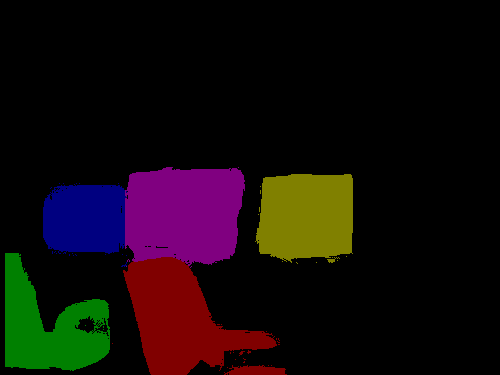} \\

	\global \def \im{2007_001526}
	\includegraphics[width=\linewidth]{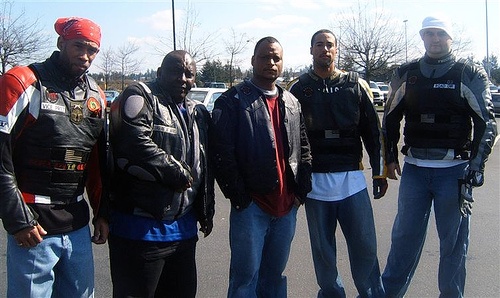} &
	\includegraphics[width=\linewidth]{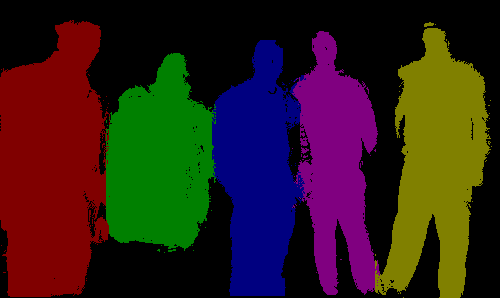} &
	\includegraphics[width=\linewidth]{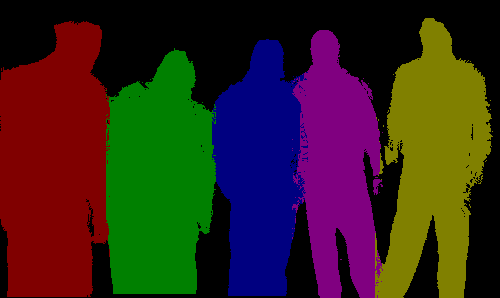} \\

	\global \def \im{2007_000129}
	\includegraphics[width=\linewidth]{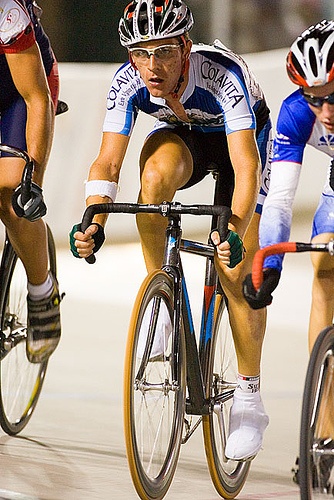} &
	\includegraphics[width=\linewidth]{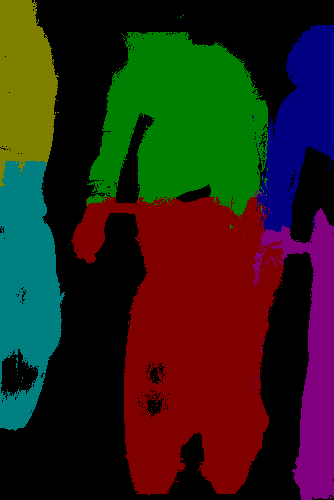} &
	\includegraphics[width=\linewidth]{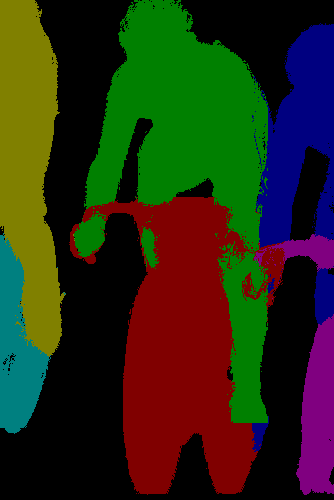} \\

	\global \def \im{2009_005302}
	\includegraphics[width=\linewidth]{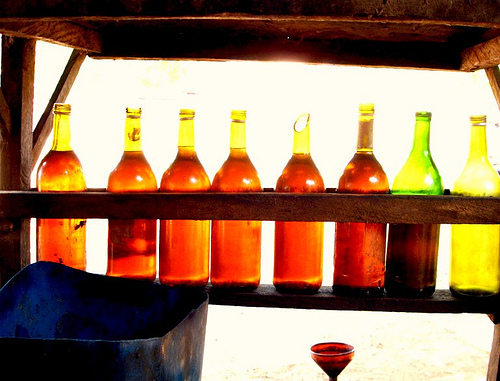} &
	\includegraphics[width=\linewidth]{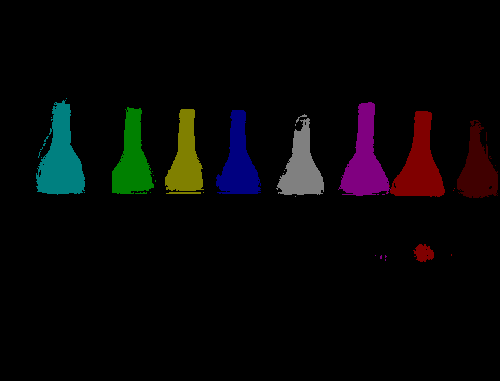} &
	\includegraphics[width=\linewidth]{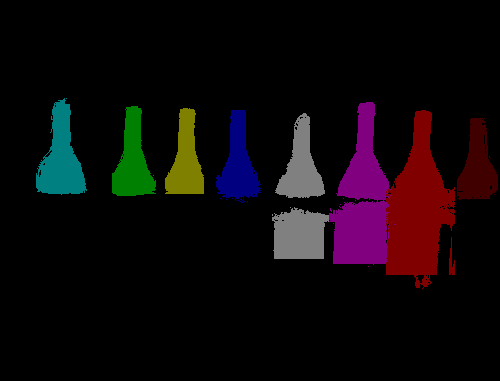} \\
		
	\end{tabularx}
	\textbf{Fig.~\ref{fig:voc_qualitative} cont.} 
	The first and second rows show examples where the results of the two models are similar.
	In the third and fourth rows, the weakly-supervised model does not segment the ``green person'' as well as the fully-supervised model.
	In the last row, both weakly- and fully-supervised models have made an error in not completely segmenting each of the bottles.
	\label{fig:voc_qualitative_cont}
\end{figure}

\begin{landscape}

\begin{table}[t]
\centering
\caption{Per-class results of our weakly- and fully-supervised models for both semantic and instance segmentation on the Cityscapes validation set. The IoU measures semantic segmentation performance, whilst the $AP^{r}_{vol}$ and PQ measure instance segmentation performance.}
\label{tab:cityscapes_per_class}
	\scalebox{0.9}{
		\begin{tabularx}{1.1\linewidth}{l Y Y Y Y Y Y Y Y Y Y Y Y Y Y Y Y Y Y Y Y}
		\toprule
		Metric & Mean & road & side-walk & build-ing & wall & fence & pole & traffic light & traffic sign & vege-tation & terrain & sky & person & rider & car & truck & bus & train & motor-cycle & bi-cycle \\
		\cmidrule(r){1-1} \cmidrule(lr){2-2} \cmidrule(lr){3-21}
		\multicolumn{18}{l}{\textit{Weakly supervised model}} \\
		IoU &   63.6    &   93.3    &   59.3    &   86.6    &   38.7    &   29.6    &   32.0    &   44.0    &   59.2    &   88.7    &   39.1    &   91.7    &   69.4    &   48.4   &   87.4    &   68.0    &   80.7    &   68.0    &   56.0    &   67.5 \\
		$AP^r_{vol}$ & 26.3 & 82.7 &  27.6 &  68.1 &  5.9 &   5.2 & 0.6 &  3.0 &  16.6 &  74.1 &  4.7 &  76.1 & 11.7 &  5.0 &  27.7 &  17.4 &    36.3 &  23.0 &  9.0 &  5.9
		  \\
		PQ            & 40.5 &  91.2 & 47.0 & 79.6 &  14.8 & 12.7 & 5.5 & 13.2 & 37.3 & 83.3 & 16.2 & 82.3 & 30.6 & 25.7 & 46.9 & 33.7 & 55.5 & 37.0 & 31.8 & 24.9 \\
		\cmidrule(r){1-1} \cmidrule(lr){2-2} \cmidrule(lr){3-21}
		\multicolumn{18}{l}{\textit{Fully supervised model}} \\
		IoU &   71.6    &   97.6    &   81.9    &   90.4    &   42.2    &   52.3    &   54.5    &   61.1    &   71.8    &   90.5    &   61.1    &   93.5    &   76.6    &   53.2   &   93.4    &   68.3    &   77.8    &   70.6    &   50.7    &   72.3 \\
		$AP^r_{vol}$ & 34.9 & 94.8 & 56.2 & 73.6 & 10.5 & 7.4 & 11.9 & 10.7 & 31.9 & 77.3 & 16.2 & 78.2 & 21.2 & 15.0 & 32.6 & 25.5 & 41.4 & 30.5 & 15.3 & 12.6 \\
		PQ &  47.3 & 95.5 & 67.9 & 83.4 & 17.2 & 15.5 & 38.0 & 22.2 & 54.7 & 84.7 & 21.7 & 80.4 & 40.4 & 37.1 & 49.8 & 31.8 & 54.1 & 36.4 & 34.3 & 32.5 \\
		\bottomrule
		\end{tabularx}
	}
\end{table}

\begin{table}[t]
\centering
\caption{Per-class results of our weakly- and fully-supervised models for both semantic and instance segmentation on the Pascal VOC validation set. The IoU measures semantic segmentation performance, whilst the $AP^{r}_{vol}$ and PQ measure instance segmentation performance.}
\label{tab:voc_per_class}
	\scalebox{0.9}{
		\begin{tabularx}{1.1\linewidth}{l Y Y Y Y Y Y Y Y Y Y Y Y Y Y Y Y Y Y Y Y Y}
		\toprule
			Metric & Mean & aero-plane & bike & bird & boat & bottle & bus & car & cat & chair & cow & table & dog & horse & motor-bike & per-son & plant & sheep & sofa & train & tv \\
		\cmidrule(r){1-1} \cmidrule(lr){2-2} \cmidrule(lr){3-22}
		\multicolumn{18}{l}{\textit{Weakly supervised model}} \\
		IoU &   75.7 &  85.0    &   35.9    &   88.6    &   70.3    &   77.9    &   91.9    &  83.6    &   90.5    &   39.2    &   84.5    &   59.4    &   86.5    &   82.4    &  81.5    &   84.3    &   57.0    &   85.9    &   55.8    &   85.8    &   70.4\\
		$AP^r_{vol}$ & 55.5 & 68.8    & 26.4    &   74.4    &    50.4   &    37.9   &    70.0   &   49.4  &    78.6   &    22.0   &    57.1   &    37.4   &  78.7 &    61.6   &    61.7    &    50.8   &    42.2   &    54.6   &    46.9   &    74.9   &    66.5
		\\
		PQ     &    59.5 &    69.7   &    18.0   &    76.8   &    55.1   &    48.2   &    75.4   &    54.9 &    77.8   &    26.4   &    65.8   &    43.6   &    73.8   &    62.9   &    68.9 &    60.8   &    48.7   &    62.9   &    53.7   &    75.9   &    71.4 
		\\
		\cmidrule(r){1-1} \cmidrule(lr){2-2} \cmidrule(lr){3-22}
		\multicolumn{18}{l}{\textit{Fully supervised model}} \\
		IoU & 79.0  & 92.0  &   42.2    &   90.6    &   71.1    &   80.7    &   95.0    &   88.5   &   91.9    &   41.5    &   90.6    &   60.3    &   86.5    &   88.3    &   85.4   &   86.9    &   61.7    &   91.6    &   53.3    &   89.2    &   76.8 \\
		$AP^r_{vol}$ & 59.5 &   77.1    &    31.7   &    78.1   &    50.9   &    40.2   &    72.4  &    52.6   &    82.9   &    27.0   &    60.3   &    35.4   &     83.1  &    65.4  &    72.3   &    57.3   &    45.6   &    56.4   &    49.7   &    80.1   &    71.3  \\
		PQ & 63.1   &   77.8    &    29.1   &    79.0   &    57.2   &    48.9   &    75.5   &    59.8  &    81.7   &    31.8   &    67.3   &    46.2   &    77.3   &    69.0   &    75.3  &    64.8   &    52.2   &    62.0   &    54.6   &    79.8   &    73.7 \\
		\bottomrule
		\end{tabularx}
	}
\end{table}

\end{landscape}

\clearpage
\section{Experimental Details}
\label{sec:experimental_details}

\subsection{Network architecture and training}
The underlying semantic segmentation network is a reimplementation of PSPNet \cite{zhao_cvpr_2017} as described in Sec.~3.5 of the main paper, using a ResNet-101 backbone.
This network has an output stride of 8, meaning that the result of the network has to be upsampled by a factor of 8 to obtain the final prediction at the original resolution.

We used most of the same training hyperparameters for training both our fully- and weakly-supervised networks.
A batch size of a single $521 \times 521$ image crop, momentum of $0.9$, and a weight decay of $5 \times 10^{-4}$ were used in all our experiments.

We trained the semantic segmentation module first, and finetuned the entire instance segmentation network afterwards.
For training the semantic segmentation module, the fully supervised models were trained with an initial learning rate of $1 \times 10^{-4}$, which was then reduced to $1 \times 10^{-5}$ when the training loss converged.
We used the same learning rate schedule for our weakly-supervised model on Pascal VOC where we did not do any iterative training.
In total, about $400\text{k}$ iterations of training were performed.
When training our weakly-supervised model iteratively on Cityscapes, we used an initial learning rate of $1 \times 10^{-4}$ which was then halved for each subsequent stage of iterative training.
Each of these iterative training stages were $150\text{k}$ iterations long.
Both of the weakly- and fully-supervised models were initialised with ImageNet-pretrained weights and batch normalisation statistics.

In the instance training stage, we fixed the learning rate to $1\times 10^{-5}$ for both weakly- and fully-supervised experiments on the VOC and Cityscapes datasets. 
We observed that a total of $400 \text{k}$ iterations were required for the models' training losses to converge.

When training the Faster-RCNN object detector \cite{ren_2015}, we used all the default training hyperparameters in the publicly available code.

\subsection{Multi-label classification network}
\label{sec:multilabel_training}

We obtained weak localisation cues, as described in Sec.~3.3 of the main paper, by first training a network to perform multi-label classification on the Cityscapes dataset.

We adapted the same PSPNet \cite{zhao_cvpr_2017} architecture for segmentation for the classification task:
The output of the last convolutional layer (\texttt{conv5\_4}) is followed by a global average pooling layer to aggregate all the spatial information.
Thereafter, a fully-connected layer with 19 outputs (the number of classes in the Cityscapes dataset) is appended.
This network was then trained with a binary cross entropy loss for each of the 19 labels in the dataset.
The loss for a single image is
 \begin{equation}
 L = \frac{1}{N} \sum_{i=1}^{N}{ -y_i \log(\text{sigmoid}(z_i)) - (1 - y_i)\log(1 - \text{sigmoid}(z_i)) },
 \end{equation}
where $\mathbf{y}$ is the ground truth image-level label vector and $y_i = 1$ if the $i^{th}$ class is present in the image and 0 otherwise. $z_i$ is the logit for the $i^{th}$ class output by the final fully-connected layer in the network.

It is not possible to fit an entire $2048 \times 1024$ Cityscapes image in memory to perform multi-label classification.
Using the PSPNet architecture described above (with an output stride of 8), it would take 48.8 GB of memory to train a network with a batch size of 1.
Even the standard ResNet-101 architecture \cite{he_cvpr_2016} (which has a higher output stride of 32, and thus sixteen times less spatial resolution) would take 21.7 GB of memory, which is still almost double the 12GB available in our Titan X GPU.
Consequently, we took 15 fixed crops of size $500\times400$ from the original $2048 \times 1024$ image and trained with these crops instead.
We were careful not to take random crops during training, as this could be a form of extra supervision.
Instead, as we took 15 fixed crops which tile the image and derived image-level labels from them, it effectively means that in a real-world scenario annotators would be asked to annotate image-level labels for fifteen $500 \times 400$ images rather than a single $2048 \times 1024$ image.

This multi-label classification network was trained with a batch size of 1 and a fixed learning rate of $1 \times 10^{-4}$ until the training loss converged.
We found that this occurred after $50 \text{k}$ iterations of training.
At this point, the mean Average Precision (mAP) on the validation set was 78.8.
The mAP is also used by the Pascal VOC dataset to benchmark multi-label classification \cite{everingham_2010}.

\section{Comparison of Pascal VOC and Microsoft COCO annotation quality}
\label{sec:pascal_vs_coco_gt}

\begin{figure}[!t]
\centering

\begin{tabularx}{\linewidth}{ Y Y Y Y }
COCO Image & COCO Label & Pascal VOC Image & Pascal VOC Label \\

\includegraphics[width=\linewidth]{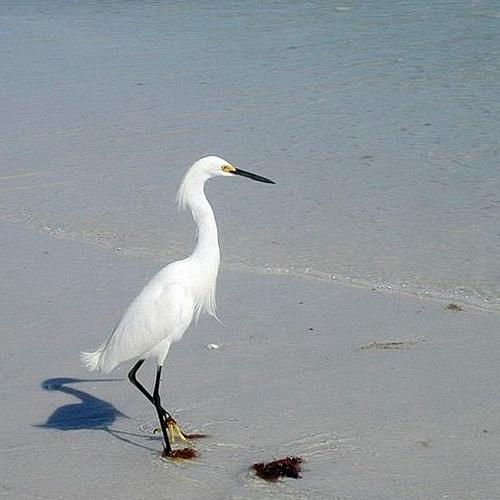} &
\includegraphics[width=\linewidth]{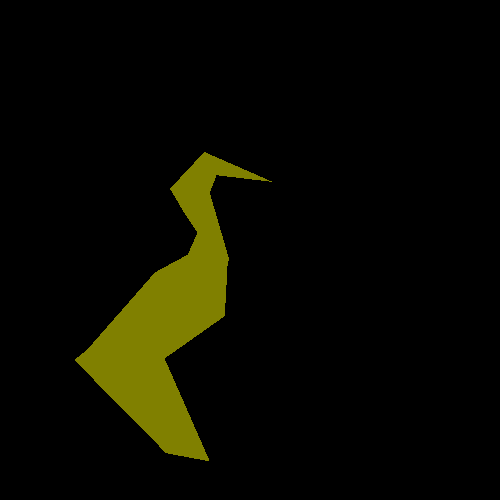} &
\includegraphics[width=\linewidth]{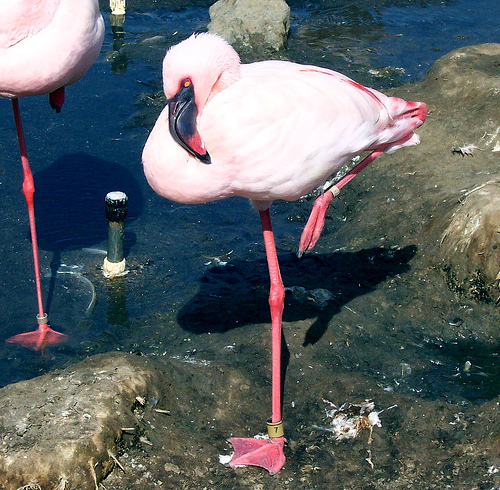} &
\includegraphics[width=\linewidth]{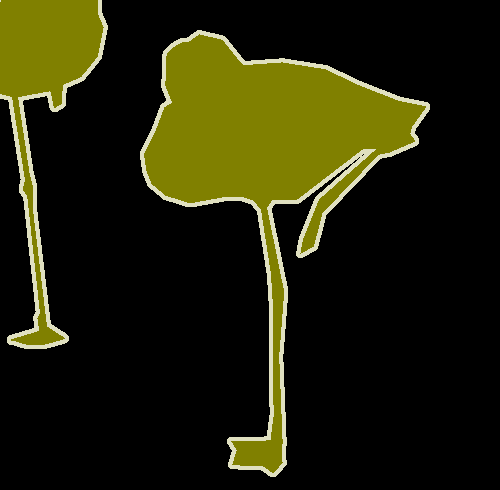}
\\

\includegraphics[width=\linewidth]{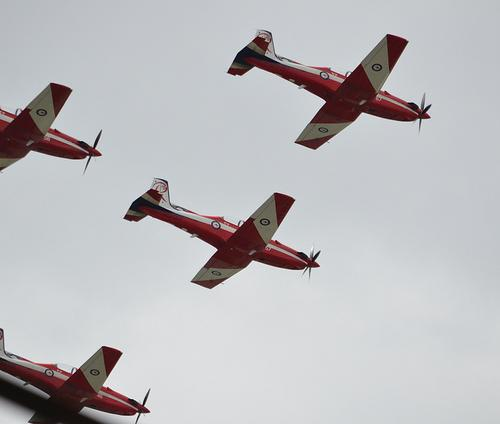} &
\includegraphics[width=\linewidth]{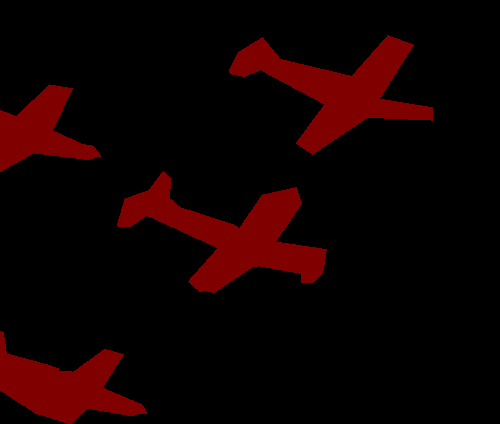} &
\includegraphics[width=\linewidth]{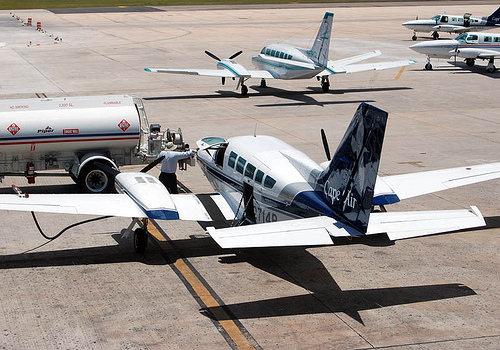} &
\includegraphics[width=\linewidth]{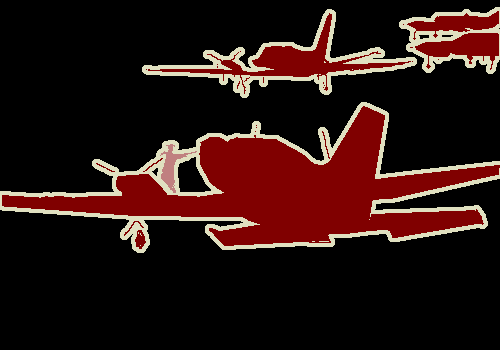}
\\

\includegraphics[width=\linewidth]{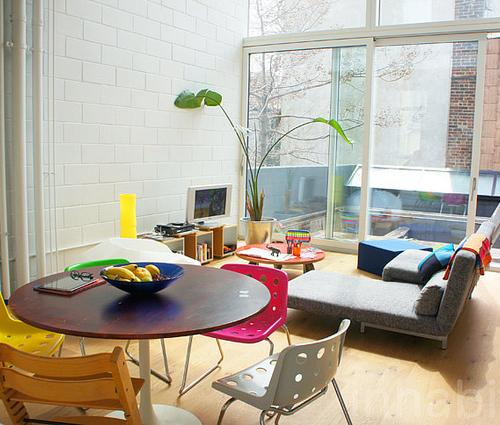} &
\includegraphics[width=\linewidth]{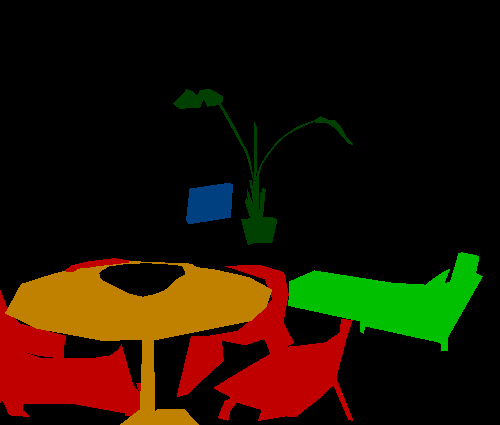} &
\includegraphics[width=\linewidth]{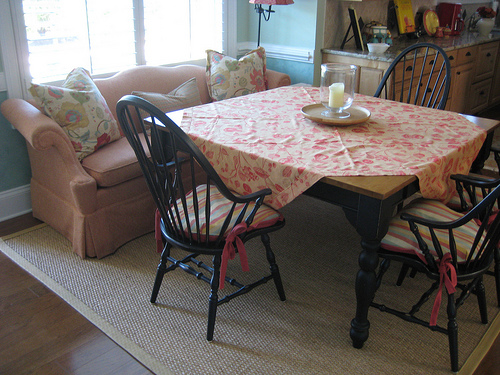} &
\includegraphics[width=\linewidth]{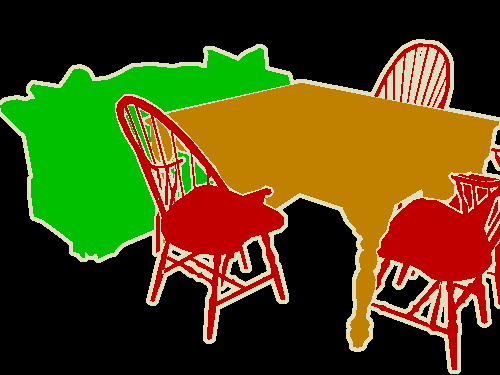}
\\

\includegraphics[width=\linewidth]{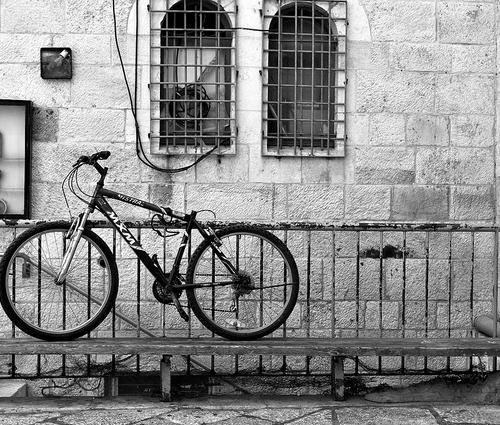} &
\includegraphics[width=\linewidth]{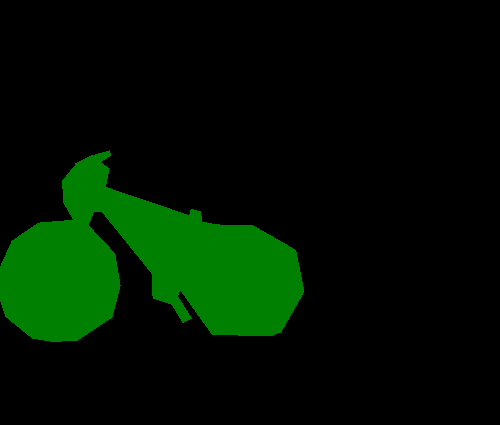} &
\includegraphics[width=\linewidth]{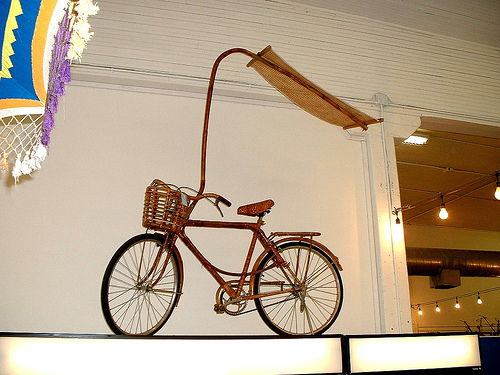} &
\includegraphics[width=\linewidth]{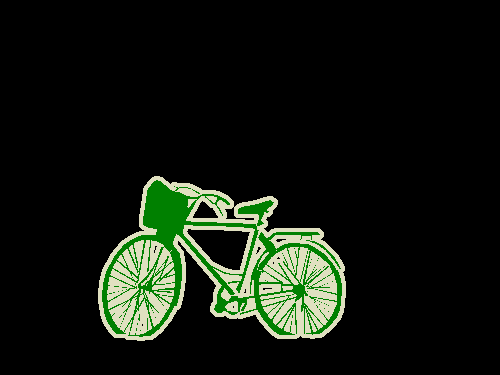}
\\

\includegraphics[width=\linewidth]{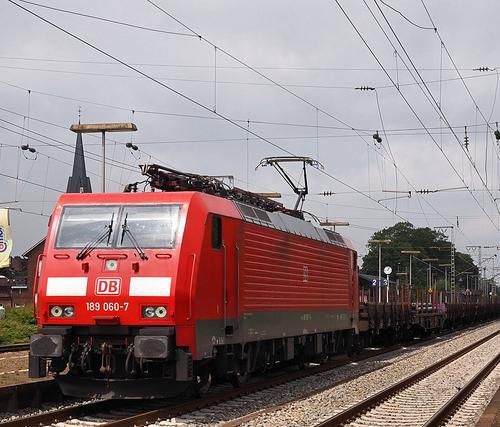} &
\includegraphics[width=\linewidth]{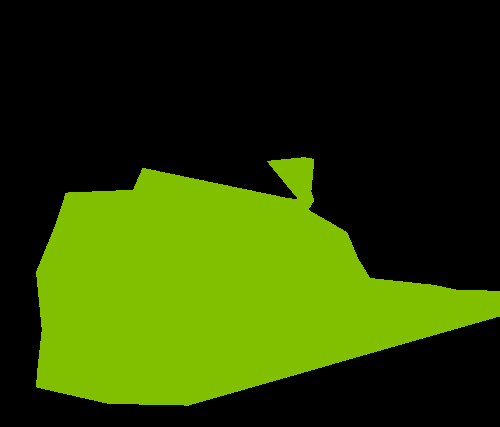} &
\includegraphics[width=\linewidth]{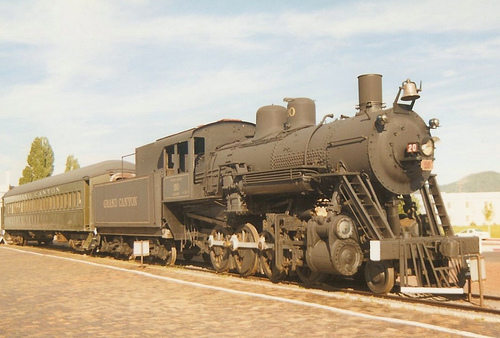} &
\includegraphics[width=\linewidth]{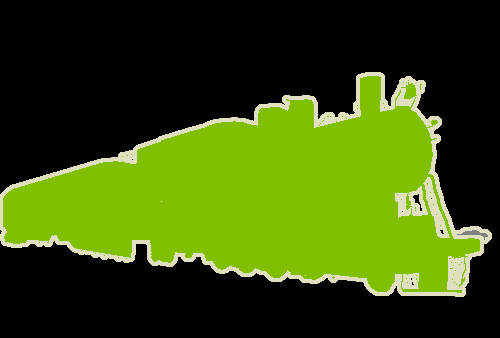}
\\

\includegraphics[width=\linewidth]{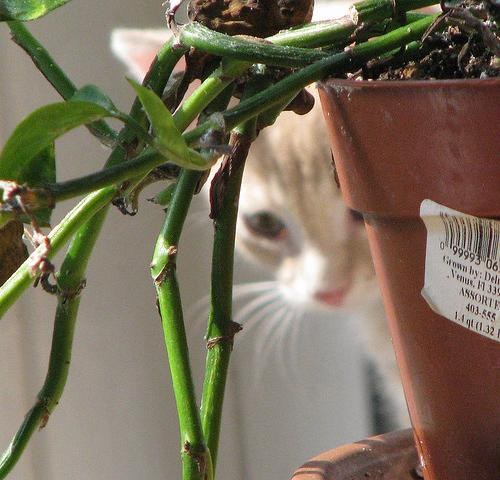} &
\includegraphics[width=\linewidth]{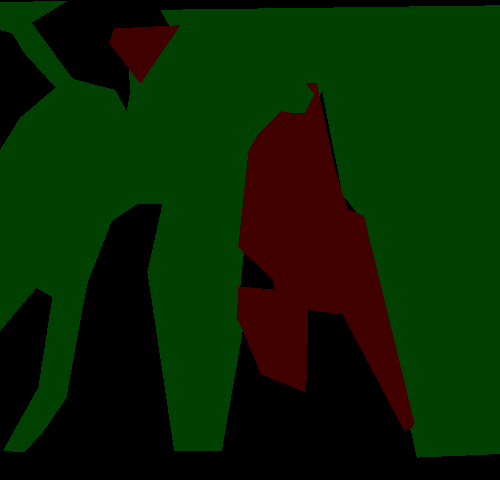} &
\includegraphics[width=\linewidth]{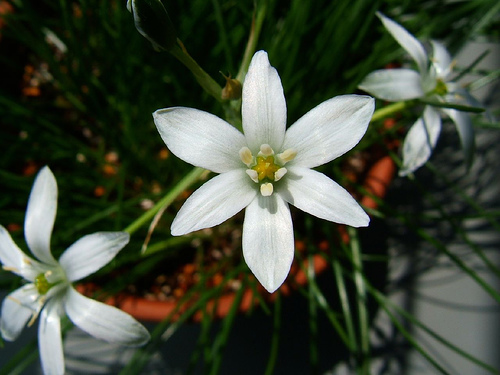} &
\includegraphics[width=\linewidth]{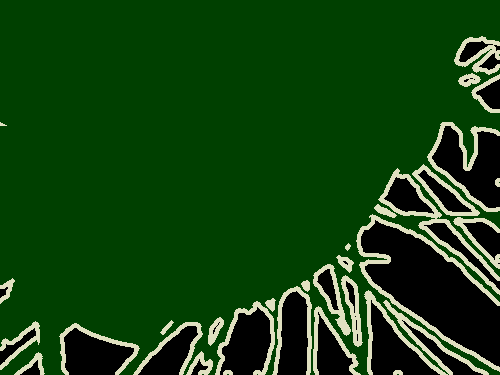}

\end{tabularx}

\caption{Comparison of the annotation quality of images in the Microsoft COCO and Pascal VOC datasets. An image was randomly drawn from COCO, and an image from Pascal VOC with similar content is shown alongside it. The polygons used to annotate the objects in COCO are evident, and the annotations at the boundaries of objects are often incorrect.
Grey regions in the Pascal images indicate ``void'' regions where the annotator was unsure of the correct label.}
\label{fig:voc_vs_coco1}
\end{figure}

Section 4.2 of the main paper mentioned that images in Pascal VOC \cite{everingham_2010} are annotated at a higher quality than those in Microsoft COCO \cite{lin_2014}.
Figure \ref{fig:voc_vs_coco1} illustrates this observation.
Images were randomly drawn from Microsoft COCO, and then images from Pascal VOC with the same semantic classes present are shown alongside for comparison.
The polygons used to annotate the objects in COCO are evident, and the annotations at the boundaries of objects are often incorrect.

\begin{figure}[!t]

\begin{tabularx}{\linewidth}{ Y Y Y Y }
COCO Image & COCO Label & Pascal VOC Image & Pascal VOC Label \\

\includegraphics[width=\linewidth]{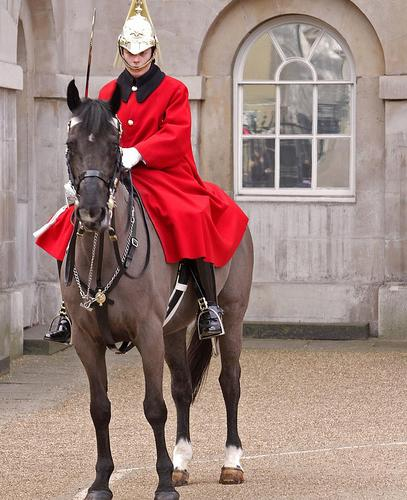} &
\includegraphics[width=\linewidth]{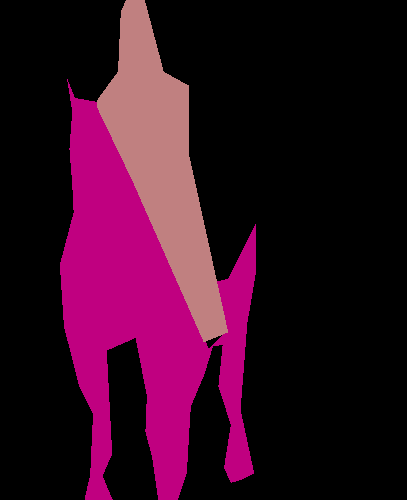} &
\includegraphics[width=0.8\linewidth]{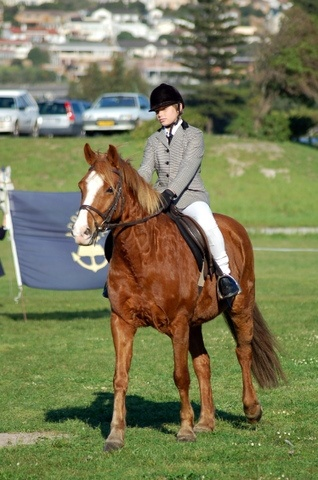} &
\includegraphics[width=0.8\linewidth]{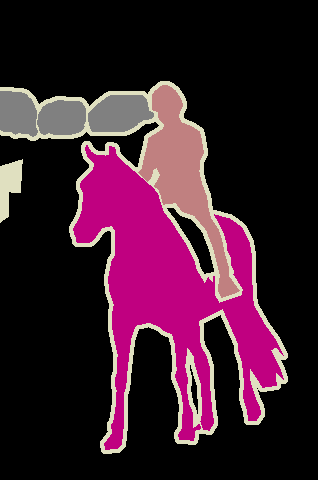}
\\

\includegraphics[width=\linewidth]{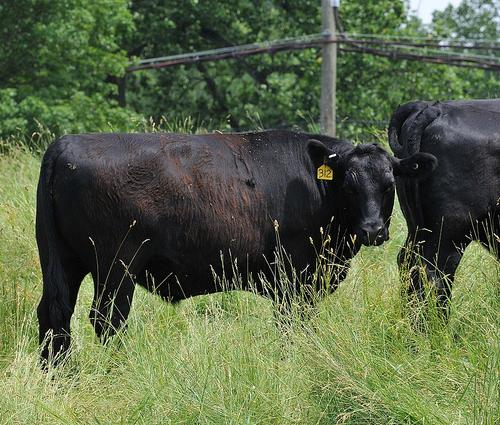} &
\includegraphics[width=\linewidth]{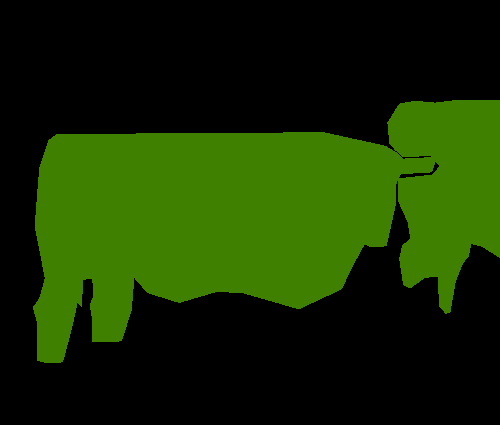} &
\includegraphics[width=\linewidth]{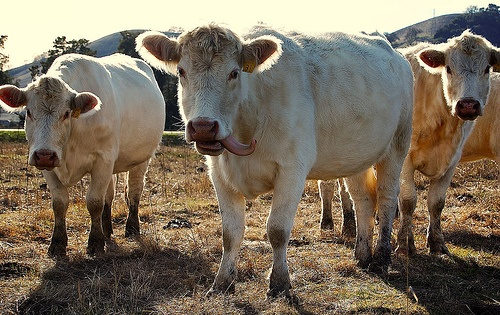} &
\includegraphics[width=\linewidth]{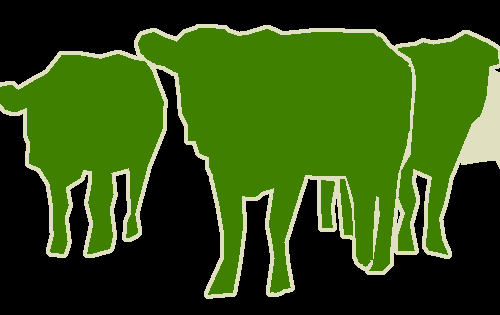}
\\

\includegraphics[width=\linewidth]{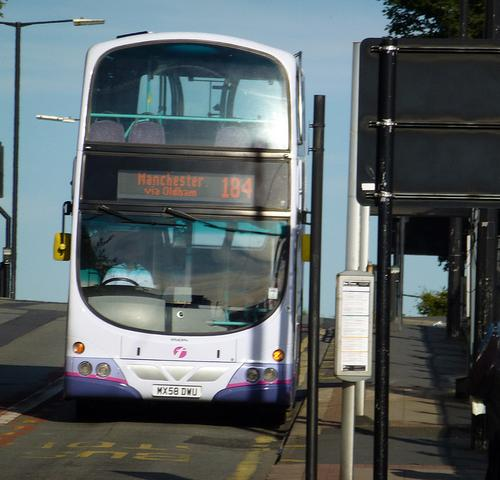} &
\includegraphics[width=\linewidth]{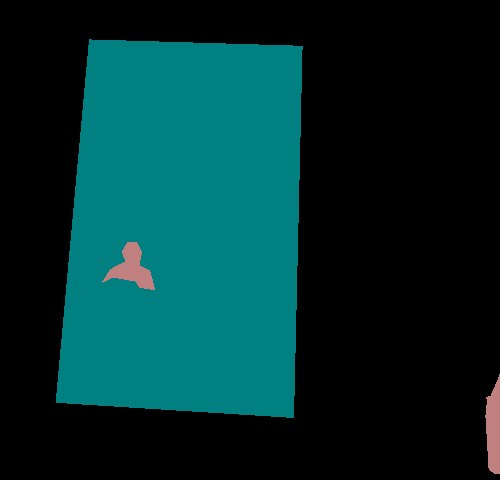} &
\includegraphics[width=\linewidth]{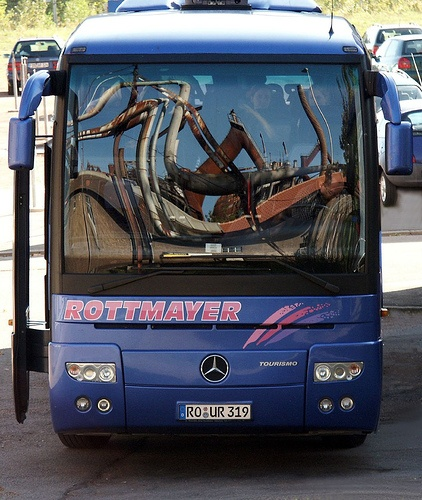} &
\includegraphics[width=\linewidth]{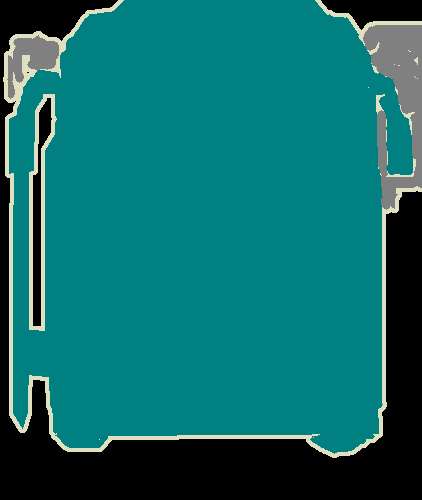}
\\

\includegraphics[width=\linewidth]{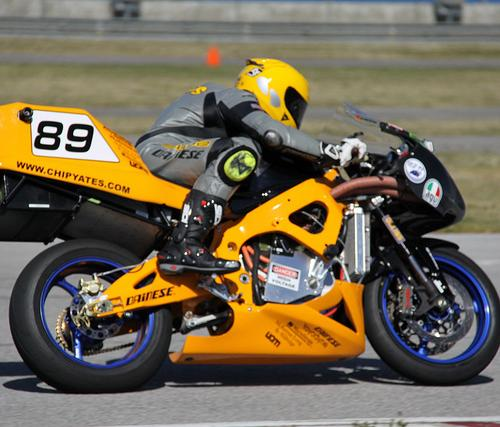} &
\includegraphics[width=\linewidth]{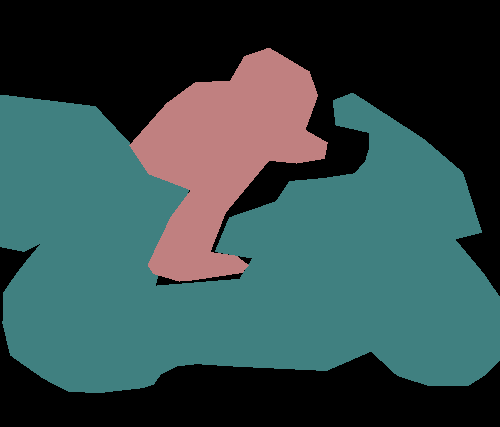} &
\includegraphics[width=0.7\linewidth]{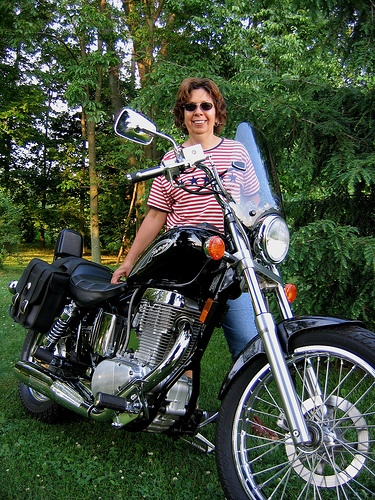} &
\includegraphics[width=0.7\linewidth]{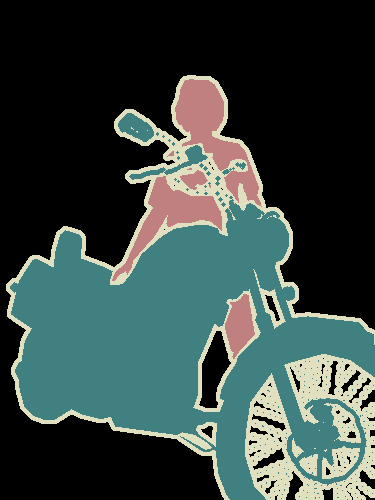}
\\

\includegraphics[width=\linewidth]{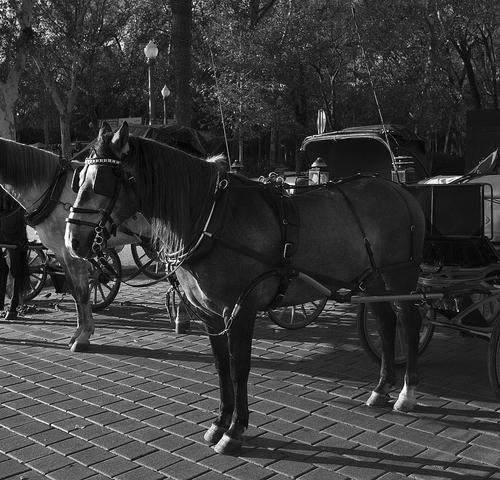} &
\includegraphics[width=\linewidth]{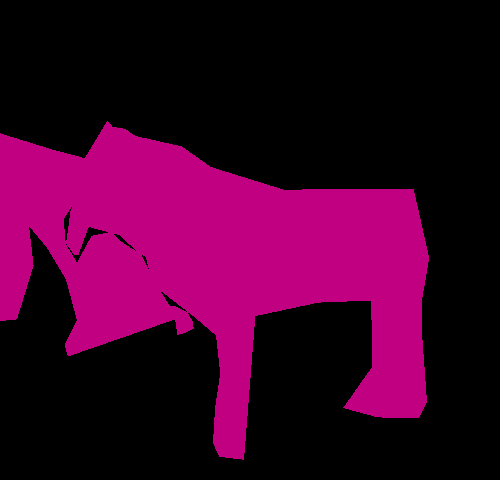} &
\includegraphics[width=\linewidth]{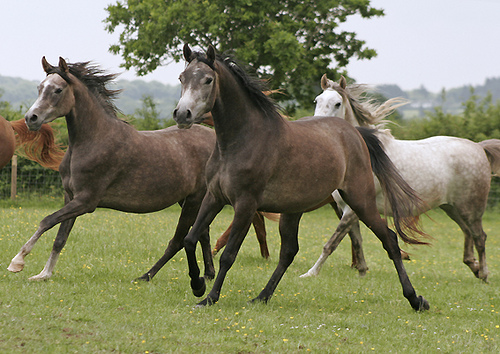} &
\includegraphics[width=\linewidth]{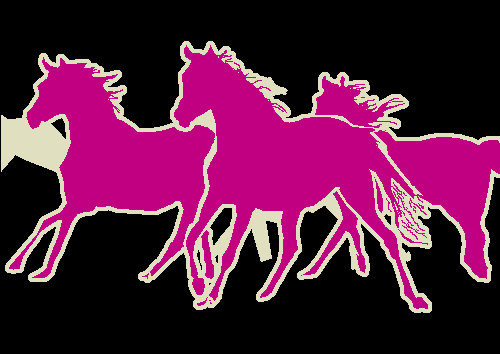}
\\

\textcolor{white}{} \\

\end{tabularx}
\textbf{Fig.~9 cont.} Comparison of the annotation quality of images in the Microsoft COCO and Pascal VOC datasets. An image was randomly drawn from COCO, and an image from Pascal VOC with similar content is shown alongside it.
The polygons used to annotate the objects in COCO are evident, and the annotations at the boundaries of objects are often incorrect.
Grey regions in the Pascal images indicate ``void'' regions where the annotator was unsure of the correct label.
\label{fig:voc_vs_coco2}
\end{figure}

\section{Calculation of reduction factor in annotation time if only weak labels are used}
\label{sec:annotation_time}

The Cityscapes dataset has 11 ``stuff'' classes, and 8 ``thing'' classes annotated.
Over the training and validation sets, there are an average of $17.9$ instances of ``thing'' classes per full-resolution, $2048 \times 1024$ image.

For the calculation in Sec. 1 of the paper, we assumed that each instance of a ``thing'' class is labelled with a bounding box, and that image-level tags are annotated for all present ``stuff'' classes.
We assumed that a bounding box takes 7 seconds per instance to draw \cite{papadopoulos_iccv_2017} and that an image-level tag takes 1 second to label \cite{papadopoulos_eccv_2014}.

Therefore the average time to annotate ``thing'' classes with a bounding-box is $17.9 \times 7 = 125.3$ seconds.
As we took 15 fixed crops per image (as described in Sec.~\ref{sec:multilabel_training}) and there are an average of 3.8 ``stuff'' tags per crop, the average time to annotate stuff classes is $15 \times 3.8 = 57$ seconds.
This totals $182.3 $ seconds $= 3.0 $ minutes per image.
Thus the annotation time is reduced by a factor of $29.6$ (since the images originally required 90 minutes to label at a pixel-level by hand \cite{cordts_cvpr_2016}) if weak annotations in the form of bounding boxes and image-level tags are used.

\end{document}